\documentclass{article}
\usepackage{iclr2021_conference,times}
\usepackage{amsmath,amssymb,bm,amsthm}
\usepackage{booktabs,tabularx,float}
\usepackage{graphicx}
\usepackage{xcolor}
\usepackage{hyperref}
\hypersetup{hidelinks}
\iclrfinalcopy
\newtheorem{theorem}{Theorem}
\newcommand{\latentexteriornote}{Outside the black physical boundary, the $D$-component physical field and displayed reconstruction are deliberately masked; the $D-1$ latent is deliberately left visible because the Fourier/Householder encoder is the nonlocal spatial convolution in Appendix~\ref{app:exterior_latent}.}

\title{Learning in the Transverse Subspace: A Minimal Representation for Divergence-Free Operator Learning}
\author{Yifei Sun\thanks{\texttt{yifeisun@umich.edu}}}

\begin{document}
\maketitle
\lhead{}

\begin{abstract}
Divergence-free vector fields are fundamental state variables in incompressible flows and many PDE systems.  Redundant methods, including Neural Conservation Law (NCL) potentials, map multiple auxiliary fields to the same physical divergence-free field.  Our experiments show that this redundancy can lower static representation-fitting error.  However, their many-to-one, non-invertible mapping does not provide the unique function state required by operator learning, where an evolution operator maps each state function to its future state.

We introduce a minimal representation that encodes a real $D$-component divergence-free vector field on a $D$-dimensional domain as a real $(D-1)$-component vector field on the same domain.  For periodic fields and closed impermeable fields, the direct transform is invertible, isometric, and angle-preserving.  For open nonperiodic flows, we use Fourier extension to construct a compatible periodic field, and a minimum-energy rule selects a reduced representation.

These coordinates provide a direct state space for divergence-free operator learning.  Training data are first encoded, and the neural model learns the evolution directly in the $(D-1)$ reduced components.  At inference, the predicted reduced field is decoded back to a physical $D$-component divergence-free field.  The model never predicts an ambient $D$-component velocity and does not need a projection to remove its longitudinal or divergent component.  Experiments show that this end-to-end divergence-free formulation achieves lower error and greater robustness.  In projection-based and hard-constrained alternatives, a predicted redundant component can be discarded by projection or decoding; because the physical loss does not constrain that direction, it can introduce instability.  Our model never represents this longitudinal/null direction.
\end{abstract}

\section{Introduction}
Divergence-free constraints are fundamental geometric structures associated with conservation in physics. In incompressible fluid dynamics, the velocity field satisfies $\nabla\!\cdot u=0$, expressing local volume preservation; in electromagnetism and magnetohydrodynamics, the magnetic field satisfies $\nabla\!\cdot B=0$; and more generally local conservation is expressed by a continuity equation, $\partial_t\rho+\nabla\!\cdot j=0$, or by a divergence-free current in space--time. Thus, divergence-free structure is not merely a numerical convenience: it is a condition of physical admissibility in fluid dynamics, electromagnetism, magnetohydrodynamics, and other conservation-law systems \citep{chorin1968numerical,toth2000constraint,richterpowell2022ncl}.

This structure poses a basic question for machine learning. Neural models are increasingly used both to reconstruct physical vector fields and to learn their dynamics. In the latter setting, operator learning seeks a function-to-function evolution map
\begin{equation}
\mathcal{G}:u(\cdot,t)\longmapsto u(\cdot,t+\Delta t).
\end{equation}
An unconstrained neural output does not generally remain divergence-free. A substantial literature therefore enforces the constraint exactly rather than only penalizing its violation. The dominant approaches either parameterize the physical field by an auxiliary potential, or predict in the ambient space and subsequently remove the inadmissible component \citep{kim2019deep,richterpowell2022ncl,tompson2017fluidnet}.

Potential-based approaches exploit identities such as $\nabla\!\cdot(\nabla\!\times A)=0$, or their differential-form generalizations, so that a network predicts an auxiliary object whose image under a fixed differential operator is automatically divergence-free. This principle underlies stream-function and vector-potential constructions as well as Neural Conservation Laws (NCL) \citep{kim2019deep,richterpowell2022ncl}. Such constructions guarantee physical admissibility, but their auxiliary coordinates are generally not minimal or identifiable. In three dimensions, a divergence-free vector field has two independent transverse degrees of freedom at each nonzero Fourier mode, whereas a vector potential has three components and is unchanged by the gauge transformation $A\mapsto A+\nabla\phi$. More generally, antisymmetric-potential constructions may use $D(D-1)/2$ components to represent a transverse space of dimension only $D-1$. Consequently, multiple auxiliary fields can encode the same physical state. As we show experimentally, this redundancy can make static representation fitting easier, but it does not provide a unique coordinate for a physical state.

Projection-based approaches address the constraint differently. A network first predicts an ambient $D$-component vector field $\widetilde u$, after which a Leray--Helmholtz projection or pressure-correction procedure removes its longitudinal component \citep{chorin1968numerical,tompson2017fluidnet}. On a periodic Fourier domain, the projection at every nonzero wavevector is
\begin{equation}
P_k=I-\frac{kk^\top}{|k|^2},
\qquad
P_k\bigl(\widehat{\widetilde u}(k)+\alpha k\bigr)=P_k\widehat{\widetilde u}(k)
\quad\text{for any $\alpha$}.
\end{equation}
The projected physical loss is therefore insensitive to the longitudinal component generated before projection. The network may spend capacity predicting a component that is subsequently discarded, while that null direction may drift without changing the projected state. This does not invalidate projection methods, but it introduces unnecessary degrees of freedom that can degrade conditioning or numerical robustness during repeated temporal rollout.

These observations motivate a simple question: \emph{if a divergence-free state intrinsically has only $D-1$ degrees of freedom, why learn $D$ or more quantities and remove their redundancy afterward?} For a $D$-dimensional divergence-free vector field, the Fourier constraint $k^\top\widehat u(k)=0$ implies that every nonzero Fourier coefficient lies in the $(D-1)$-dimensional transverse subspace
\begin{equation}
k^\perp=\{v\in\mathbb{C}^D:k^\top v=0\}.
\end{equation}
This reduction is classical in spectral descriptions of incompressible turbulence, where transverse and helical bases represent a three-component velocity mode using two independent amplitudes \citep{waleffe1992triads,kimura2012spectra}. Our key observation is to use these intrinsic transverse coordinates not merely as an analytical decomposition, but as the state variables learned by a neural operator.

We construct a fixed analytic Fourier--Householder transformation that maps a real $D$-component divergence-free field to a real $(D-1)$-component field on the same spatial domain. At each nonzero Fourier mode, a deterministic orthonormal basis $E(k)\in\mathbb{R}^{D\times(D-1)}$ spans $k^\perp$, yielding $\widehat u(k)=E(k)\widehat a(k)$ and $\widehat a(k)=E(k)^\top\widehat u(k)$. The unconstrained zero mode is handled separately. For periodic fields and closed impermeable fields, the direct transform is invertible, isometric, and angle-preserving. For open nonperiodic flows, we construct a compatible periodic continuation through Fourier extension and use a minimum-energy rule to select a unique reduced representation. This extends the coordinate construction beyond directly periodic data, although the extension/restriction stage is generally not isometric \citep{huybrechs2010fourier}.

The representation changes the operator-learning problem itself. We encode each training trajectory once and learn the reduced evolution $\mathcal{G}_\perp:a(\cdot,t)\mapsto a(\cdot,t+\Delta t)$ directly on the $(D-1)$ transverse components. At inference time, the predicted reduced state is deterministically decoded to a physical $D$-component divergence-free field. The neural model never represents the longitudinal degree of freedom, and every decoded prediction is divergence-free by construction. Thus, rather than learn in redundant ambient coordinates and enforce incompressibility afterward, we learn the evolution directly in a $(D-1)$-component coordinate system for the same physical state space.

\paragraph{Contributions.} First, we introduce a minimal, deterministic $D\!\to\!D-1$ coordinate representation for real divergence-free vector fields. Under the stated zero-mode convention, it is invertible, isometric, and angle-preserving for periodic and closed impermeable fields; Fourier extension together with a minimum-energy selection rule extends it to open nonperiodic domains. Second, we use the reduced coordinates as the state space of operator learning: training and prediction take place entirely in the transverse components, and a fixed decoder maps every predicted state directly to a physical divergence-free field without post-hoc projection. Third, we distinguish static representation fitting from dynamical operator learning. Redundant, non-identifiable parameterizations can facilitate static fitting, whereas temporal evolution requires an identifiable state coordinate; our experiments show that removing the longitudinal/null direction leads to more accurate and robust divergence-free operator learning.

\section{Related Work}
Classical spectral treatments of 3D incompressible turbulence already exploit the $3\to2$ transverse degree-of-freedom reduction. Craya--Herring (equivalently toroidal--poloidal in this setting) and helical decompositions express each divergence-free Fourier velocity mode through two transverse amplitudes; they have been used in high-resolution DNS and to study mode/triad interactions and energy/helicity transfer \citep{kimura2012spectra,waleffe1992triads}. These constructions are formulated around three-dimensional turbulence and do not provide the arbitrary-$D$, reversible $D\to D-1$ coordinate layer considered here.

Leon \& Scheinker (2024) are closer to the learning setting: their Fourier--Helmholtz--Maxwell neural operator (FoHM-NO) predicts a transverse Fourier vector potential and thereby bypasses gauge ambiguity while reducing the modeled field content to physical transverse degrees of freedom \citep{leon2024fohm}. Its demonstrated task is same-time, frame-wise electromagnetic field reconstruction from charge/current data rather than learning a temporal evolution operator: for relativistic beams the network maps $(\Re\widehat{J}_{\perp,z},\Im\widehat{J}_{\perp,z})$ to $(\Re\widehat{A}_{\perp,z},\Im\widehat{A}_{\perp,z})$, from which the electromagnetic fields are reconstructed. It therefore does not instantiate a general reversible $D\to D-1$ coordinate transform for divergence-free fluid operator learning.

Hard divergence-free constructions in learning systems largely fall into a small number of mathematical families. Potential-based methods predict an auxiliary scalar, vector, or antisymmetric field and apply a fixed differential map whose range is divergence-free. Deep Fluids \citep{kim2019deep} decodes a latent state to a stream function or vector potential before applying curl; the hard-constrained vector-potential coarse-graining construction \citep{mohan2023embedding} uses fixed finite-difference derivatives; and the $a$-Net \citep{wandel2021learning} predicts a next-step potential and reconstructs velocity by curl. Neural Conservation Laws \citep{richterpowell2022ncl} generalizes this family through differential forms and antisymmetric potentials, while clawNO \citep{liu2024clawno} adapts the same idea to neural operators with a fixed numerical differentiation layer. The same hard-potential principle also appears in physics-constrained GANs for turbulence \citep{tretiak2022pcgan}, Decoupled-DFNN \citep{cheng2026decoupled}, and DAF-FlowNet \citep{bisbal2026dafflownet}. These methods guarantee admissibility through the output map, but the auxiliary representation is generally redundant: in 3D a vector potential has a longitudinal gauge direction, and in general an antisymmetric $D\times D$ potential uses $D(D-1)/2$ components to represent only $D-1$ transverse degrees of freedom.

Exact projection methods instead predict a velocity-like field in the ambient space and then apply an analytic Leray/Helmholtz projection. The physics-constrained latent Neural ODE \citep{shankar2022validation} decodes a latent state and applies a spectral divergence-free projection; the hard-constraint operator framework \citep{duruisseaux2024hard} formulates transformed-space projection as an architecture-agnostic constraint; and the deterministic \emph{Project} branch \citep{li2026project} applies a differentiable Leray projector. In all of these methods, the network may still produce a longitudinal component that is subsequently discarded.

A distinct third route follows the classical CFD pressure-correction viewpoint and learns the \emph{quantity to be removed}, rather than directly applying the exact projector. Starting from a provisional velocity $u^*$, the exact pressure projection solves $\Delta p=(\rho/\Delta t)\nabla\!\cdot u^*$ and then performs the fixed correction $u=u^*-(\Delta t/\rho)\nabla p$. The Data-Driven Projection Method \citep{yang2016projection} and FluidNet \citep{tompson2017fluidnet} replace the expensive pressure/Poisson solve by a learned pressure estimate, after which a fixed numerical gradient subtraction produces the corrected velocity. Thus the learned object is the pressure/longitudinal correction that must be subtracted from $u^*$, rather than a minimal transverse coordinate or an exact Fourier projector. On a periodic Fourier domain the same correction is available analytically through $P_k=I-kk^\top/|k|^2$.

Another line builds divergence-free function spaces directly through kernels or probabilistic priors. The DFK construction \citep{ni2025dfk} uses matrix-valued divergence-free kernels for continuous flow reconstruction, while Fluids You Can Trust \citep{sharma2026fluids} uses property-preserving divergence-free kernel bases inside an operator-learning framework. Linearly Constrained Gaussian Processes \citep{jidling2017linearly} provides a general Gaussian-process construction for linear differential constraints, and Gaussian Process Hydrodynamics \citep{owhadi2023gph} specializes divergence-free Gaussian-process priors to Euler/Navier--Stokes hydrodynamics. These works differ substantially in task---reconstruction, inference, PDE solving, operator learning, generation, or dynamics---but their divergence-free mechanisms repeatedly reduce to potentials, exact projection, learned pressure correction, or analytically constrained bases.

Our approach targets a different representation question: rather than producing a redundant potential or an ambient $D$-vector that is later projected, we parameterize each nonzero Fourier mode directly by its intrinsic $D-1$ coordinates in $k^\perp$. The resulting map is reversible after the stated zero-mode convention and contains no longitudinal null direction. This gives a minimal coordinate representation of the same admissible divergence-free state space and separates the geometry of incompressibility from the subsequent learning architecture and dynamics.

\section{Methodology}
We first show that a real $D$-component divergence-free vector field occupies only an intrinsic $(D-1)$-dimensional transverse subspace at every nonzero Fourier mode.  Consequently, the field admits an equivalent $(D-1)$-component representation on the same spatial domain.  Learning the evolution or operator directly in these intrinsic coordinates is therefore sufficient: decoding any reduced state recovers a $D$-component divergence-free field by construction.
For $u:\Omega\subset\mathbb{R}^D\to\mathbb{R}^D$, define $\widehat u(k)=\int_\Omega u(x)e^{-ik\cdot x}\,dx$. Starting from the product rule,
\begin{align}
\nabla\!\cdot\!\left(u e^{-ik\cdot x}\right)
&=(\nabla\!\cdot u)e^{-ik\cdot x}+u\cdot\nabla e^{-ik\cdot x}
=(\nabla\!\cdot u)e^{-ik\cdot x}-i(k^\top u)e^{-ik\cdot x},\\
\int_\Omega \nabla\!\cdot\!\left(u e^{-ik\cdot x}\right)dx
&=\int_{\partial\Omega}(u\cdot n)e^{-ik\cdot x}\,dS
=\int_\Omega(\nabla\!\cdot u)e^{-ik\cdot x}\,dx-i k^\top\widehat u(k),\\
\therefore\qquad
k^\top\widehat u(k)
&=i\int_{\partial\Omega}(u\cdot n)e^{-ik\cdot x}\,dS
-i\int_\Omega(\nabla\!\cdot u)e^{-ik\cdot x}\,dx.
\end{align}
Thus, under the two conditions $\nabla\cdot u=0$ in $\Omega$ and $\int_{\partial\Omega}(u\cdot n)e^{-ik\cdot x}\,dS=0$, we obtain exactly $k^\top\widehat u(k)=0$. The second condition holds for impermeable boundaries because $u\cdot n=0$ pointwise, and for periodic boundaries because opposite-face contributions cancel for the periodic Fourier modes. Hence, for every $k\neq0$, $\widehat u(k)\in k^\perp$, so each $D$-component Fourier mode contains exactly $D-1$ independent degrees of freedom, in the setting where the above conditions are satisfied. For general non-periodic boundary conditions, the same result can be obtained through Fourier extension to a periodic computational domain, after which the same transverse construction applies.

This observation suggests learning the dynamics directly in the transverse subspace: every nonzero velocity mode lies and evolves in $k^\perp$, a $(D-1)$-dimensional space.  Projection-and-generate methods predict a $D$-component field before projecting it into $k^\perp$, while potential-based constructions learn redundant vector or matrix coordinates before decoding them to the same $D-1$ degrees of freedom.  We instead learn the $D-1\to D-1$ operator directly in $k^\perp$.

\begin{figure}[t]
\centering
\includegraphics[width=\linewidth]{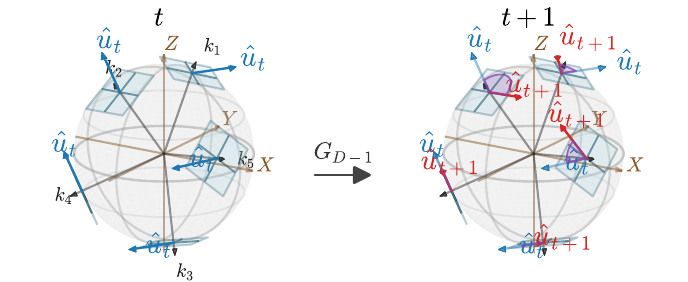}
\caption{Mode-wise function evolution in the minimal transverse representation. For every nonzero wavenumber $k_i$, both the input mode $\widehat u_t(k_i)$ and the decoded next mode $\widehat u_{t+1}(k_i)$ lie in $k_i^\perp$. The learned map $\mathcal G_{D-1}$ evolves only the $D-1$ intrinsic transverse coordinates.}
\label{fig:transverse_evolution}
\end{figure}

Accordingly, for each nonzero pair $\{k,-k\}$, we choose a canonical representative $\kappa(k)=\kappa(-k)$ and construct an orthogonal matrix $Q(k)$ whose last column is $\kappa/|\kappa|$; the remaining columns $E(k)\in\mathbb{R}^{D\times(D-1)}$ form an orthonormal basis of $k^\perp$. A convenient choice is the Householder matrix
\begin{equation}
Q(k)=I-2\frac{v_kv_k^\top}{v_k^\top v_k},
\qquad
v_k=e_D-\frac{\kappa}{|\kappa|}.
\end{equation}
The degenerate aligned case means $\kappa/|\kappa|=e_D$, for which $v_k=0$ and the Householder formula has a zero denominator; in this case we set $Q(k)=I$. This gives the unique coordinates $\widehat u(k)=E(k)\widehat a(k)$ and $\widehat a(k)=E(k)^\top\widehat u(k)\in\mathbb{C}^{D-1}$. The Householder frame is inexpensive, depends only on $k$, and lets us learn only the evolution of these $D-1$ transverse components. Since a standard Fourier neural operator takes real physical-space fields as input and output while transforming internally to Fourier space, we inverse-transform $\widehat a$ and learn on a real $(D-1)$-channel field $a(x)$ over the same $D$-dimensional domain. Enforcing the paired convention $Q(-k)=Q(k)$, hence $E(-k)=E(k)$, preserves conjugate symmetry: for real $u$, $\widehat u(-k)=\overline{\widehat u(k)}$ implies $\widehat a(-k)=\overline{\widehat a(k)}$, so $a(x)$ is real; decoding reverses the map by FFT, multiplication by $E(k)$, and inverse FFT. The zero mode is handled separately.

\paragraph{Three realizations.}
Let $N$ denote the number of points in the physical domain $\Omega$, $\widetilde N\ge N$ the number of points in an extended periodic box $\widetilde\Omega$, $M_D=M_x\otimes I_D$ the restriction from $\widetilde\Omega$ to $\Omega$, and $E=\operatorname{blkdiag}_{k\ne0}E(k)$ the transverse basis operator.  In the direct case,
\begin{equation}
A_0=F_D^{-1}EF_{D-1}.
\end{equation}
For nonperiodic data, Fourier extension enlarges the domain before the transverse transform and then restricts the reconstructed field back to $\Omega$,
\begin{equation}
A_E=M_D\widetilde F_D^{-1}\widetilde E\widetilde F_{D-1}.
\end{equation}
This second realization extends the construction beyond directly periodic or impermeable boundaries: when opposite boundaries do not match, a Fourier continuation supplies additional degrees of freedom outside $\Omega$ so that the extended field closes periodically and the transverse Fourier representation remains applicable.  Because restriction back to $\Omega$ makes the full extension nonunique, $A_E$ is generally many-to-one; the constrained minimum-energy rule selects a unique canonical extension from all representations satisfying $A_Ev=u$.

\begin{figure}[t]
\centering
\includegraphics[width=\linewidth]{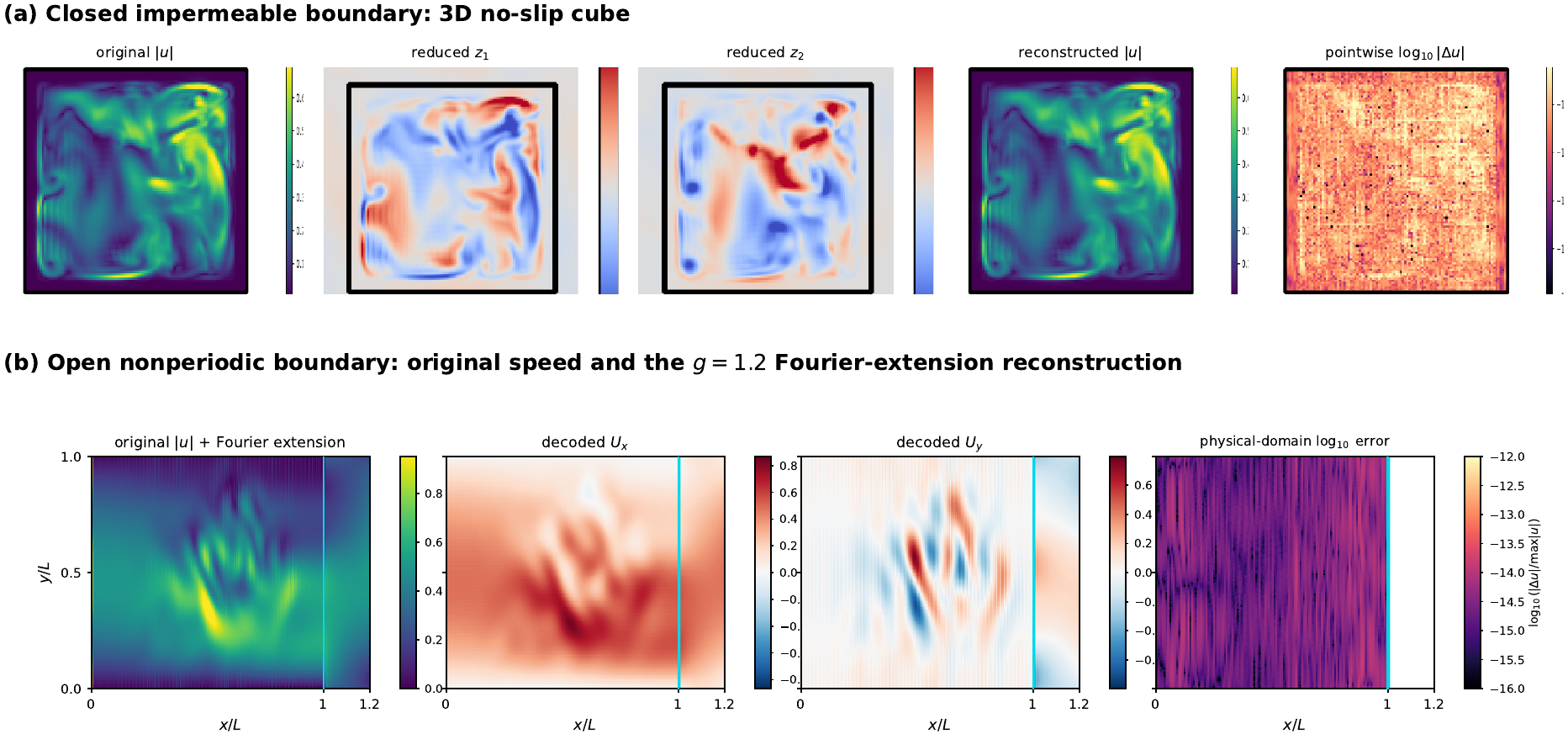}
\caption{Machine-precision reconstruction with the proposed $D\!\to\!D-1\!\to\!D$ encoder and decoder under two boundary conditions. \textbf{(a)} A closed impermeable three-dimensional no-slip cube: the physical magnitude, two real reduced channels, reconstruction, and pointwise error are shown for a representative slice. The two reduced panels retain the nonzero values outside the black physical boundary. \textbf{(b)} An open nonperiodic two-dimensional channel with a $20\%$ Fourier extension ($g=1.2$) and density $r=2$: the original speed augmented by its Fourier continuation, decoded $U_x$, decoded $U_y$, and physical-domain pointwise error are shown. The cyan line marks the physical/virtual boundary. In both settings, encoding to $D-1$ real channels and decoding back to $D$ components reconstructs the physical field to machine precision.}
\label{fig:boundary_machine_precision}
\end{figure}

To match the reduced spectral dimension to the original $N$ spatial degrees of freedom, we further choose a conjugate-symmetric mode set $K$ with $|K|=N$.  With $J_K$ denoting injection of these retained coefficients into the $\widetilde N$-mode extended spectrum and $P_K^F=J_KJ_K^*$ the associated spectral truncation,
\begin{equation}
A_K=M_D\widetilde F_D^{-1}\widetilde E J_KF_{D-1,N}.
\end{equation}
Thus Fourier extension handles boundary compatibility, whereas $K$ controls the retained bandwidth and latent dimension.  Cases I and III are used as direct one-to-one transforms on their admissible spaces, while Case II retains the full extended representation and therefore requires a canonical minimum-energy selection.

\begin{table}[H]
\centering
\small
\setlength{\tabcolsep}{7pt}
\renewcommand{\arraystretch}{1.55}
\begin{tabular}{@{}c c c c@{}}
\toprule
$\mathsf{I}$ & $\varnothing$ & $\displaystyle A_0v=u$ & $\displaystyle v_0=A_0^{-1}u=A_0^*u=F_{D-1}^{-1}E^*F_Du$ \\
\midrule
$\mathsf{II}$ & $\displaystyle \min_v\;\frac12\|v\|_2^2$ & $\displaystyle A_Ev=u$ & $\displaystyle v_E^\star=A_E^\dagger u=A_E^*(A_EA_E^*)^{-1}u$ \\
\midrule
$\mathsf{III}$ & $\varnothing$ & $\displaystyle A_Kv=u$ & $\displaystyle v_K=A_K^{-1}u$ \\
\bottomrule
\end{tabular}
\end{table}

Only Case II has a many-to-one extension ambiguity: if $z\in\ker A_E$, then $A_E(v+z)=A_Ev$.  The minimum-energy constraint selects $v_E^\star\in(\ker A_E)^\perp=\operatorname{Range}(A_E^*)$, restoring a unique canonical representation.  In Cases I and III, $A_0^{-1}$ and $A_K^{-1}$ denote the inverse maps on the corresponding admissible physical spaces.
In numerical implementations of Case II, we compute $A_E^\dagger$ by SVD, $A_E=U\Sigma V^*$, and use a thresholded inverse $\sigma_i^\dagger=1/\sigma_i$ for $\sigma_i>\tau\sigma_{\max}$ and $\sigma_i^\dagger=0$ otherwise; discarding these near-null directions avoids amplifying tiny singular values and improves the numerical stability of the forward--inverse representation.

For the open, nonperiodic Fourier-extension experiments, we found that retaining additional Fourier coefficients was also important for numerical stability: with too few coefficients, the learned solution was either inaccurate in the physical region or developed numerical blow-up in the extension region, whereas increasing the coefficient count improved physical-domain accuracy and prevented this extension-region instability.

\begin{theorem}[Reversible $D\!\to\!D\!-\!1$ representation]\label{thm:minimal_representation}
For Case II, let
\begin{equation}
A_E=M_D\widetilde F_D^{-1}\widetilde E\widetilde F_{D-1}.
\end{equation}
The forward transform (encoder) and inverse transform (decoder) are
\begin{equation}
v=A_E^\dagger u,
\qquad
u=A_Ev.
\end{equation}
For $u\in\operatorname{Range}(A_E)$ and $v\in(\ker A_E)^\perp$,
\begin{equation}
A_EA_E^\dagger u=u,
\qquad
A_E^\dagger A_Ev=v.
\end{equation}
Hence the encoder and decoder are mutually inverse and define a bijection between admissible real $D$-component divergence-free fields and canonical real $(D-1)$-component fields.  The zero Fourier mode $\widehat u(0)$ is stored separately; if $\widehat u(0)=0$, no additional side channel is required.
\end{theorem}

Unlike the redundant raw parameterizations used by NCL, Theorem~\ref{thm:minimal_representation} leaves no unresolved gauge/null direction after the stated restriction.  The direct transform $A_0$ additionally preserves inner products, distances, angles, $L^2$ and $H^s$ norms, spatial-derivative norms, and time-derivative norms, because the Fourier transforms are unitary and $E(k)$ has orthonormal columns inherited from the orthogonal Householder matrix; it commutes with fixed spatial and temporal differentiation, carries discrete and continuous dynamics by an exact coordinate conjugacy, and preserves Jacobian spectra, singular values, local Euclidean conditioning, and phase-space divergence.  Cases II and III retain exact invertibility under their stated canonical/full-rank conditions, but restriction and truncation generally remove the standard Euclidean isometry of Case I.  Detailed operator dimensions, kernel structure, and proofs are given in Appendix~\ref{app:operator_structure}.

The zero mode deserves separate treatment because incompressibility imposes no constraint at $k=0$ and the transverse basis is defined only for $k\neq0$.  Thus a nonzero mean velocity is stored as the $D$-vector $\bar u=\widehat u(0)$ alongside $v$; in the common zero-mean setting used by many periodic benchmark flows, this side channel vanishes identically.

The NCL parameterizations \citep{richterpowell2022ncl} make this redundancy explicit. For its antisymmetric form, $M^\top=-M$ and $\widehat u(k)=i\widehat M(k)k$; hence the linear map $T_k:M\mapsto iMk$ sends the $D(D-1)/2$ independent entries of $M$ into the $(D-1)$-dimensional space $k^\perp$. Therefore
\begin{equation}
\dim\ker T_k=\frac{D(D-1)}{2}-(D-1)=\frac{(D-1)(D-2)}{2},
\end{equation}
so $M$ and $M+C$ produce the same field for any antisymmetric $C$ satisfying $Ck=0$. Equivalently, in coordinates with $k=|k|e_D$, $M=\bigl(\begin{smallmatrix}B&a\\-a^\top&0\end{smallmatrix}\bigr)$ gives $Mk=|k|(a,0)^\top$, and the entire block $B\in\mathfrak{so}(D-1)$ is invisible. NCL's vector form is also non-injective: $M=J_b-J_b^\top$ yields $\widehat u(k)=-|k|^2P_k\widehat b(k)$, so $\widehat b(k)$ and $\widehat b(k)+\alpha(k)k$ are identical after decoding. Thus the matrix form uses $D(D-1)/2$ quantities and the vector form uses $D$, although the physical mode contains only $D-1$ degrees of freedom.  This non-injectivity makes the construction non-invertible and complicates learning one-to-one encoded dynamics in operator learning.

\section{Experiments}
\subsection{Representation Learning}
We initially expected the minimal, invertible transverse coordinates to be advantageous for representation learning as well.  The experiments do not support this expectation: the NCL potential representations, especially the matrix form, can reach lower loss than the minimal coordinates.  A likely reason is that a many-to-one decoder gives each physical divergence-free field many equivalent representations, so finite-step optimization may reach any of many good solutions rather than one unique coordinate.  The 10 random-initialization runs directly support this interpretation: our $D-1$-component reduced vector field converges to the same field in every run, whereas NCL's $D$-component vector potential $B$ and antisymmetric matrix potential $A$ (with $D(D-1)/2$ independent entries) vary across runs even when they decode to the same divergence-free field.  Our controlled sweep further confirms that more redundant dimensions can lower the representation error.  This advantage does not carry over to operator learning, where a non-invertible code cannot uniquely identify the state to evolve and decode; there an invertible $D-1$ representation defines an unambiguous reduced dynamics.  Details are given in Appendix~\ref{app:redundancy_optimization}.

\subsection{Operator Learning}
\label{subsec:operator_learning}

\paragraph{Benchmarks and evaluation protocol.}
We evaluate the minimal transverse representation on three representative divergence-free dynamical systems: two-dimensional Navier--Stokes (2D NS), three-dimensional Navier--Stokes (3D NS), and two-dimensional magnetohydrodynamics (2D MHD). These problems exercise different reductions of divergence-free state dimension. In 2D NS, a two-component divergence-free velocity has one independent degree of freedom per nonzero Fourier mode, giving a strict $2\!\to\!1$ representation. In 3D NS, a three-component divergence-free velocity has two independent transverse degrees of freedom, giving $3\!\to\!2$. In 2D MHD, both the velocity $u$ and magnetic field $B$ are two-component divergence-free fields; applying the $2\!\to\!1$ transform to each reduces the four physical channels $(u_x,u_y,B_x,B_y)$ to two latent channels, i.e. an overall $4\!\to\!2$ state representation.

Across the three benchmarks we use the same comparison framework. \textbf{A---Raw} learns directly in the physical field. \textbf{B---Leray} predicts an ambient vector field and applies a Fourier-space Leray projection to the output. \textbf{C---Potential} enforces divergence-freeness through a potential representation: a streamfunction in 2D NS, a magnetic-potential representation in 2D MHD, and a vector-potential/ClawNO-style representation in 3D NS. \textbf{D---Latent2 (ours)} evolves the proposed minimal $D\!\to\!D-1$ transverse coordinates directly. For 2D NS we additionally include a strong \textbf{FNO-vorticity} baseline that evolves the scalar vorticity field in closed loop.

All models use Fourier-based neural-operator backbones and are trained to convergence on V100/A100 GPUs. Wherever possible, the data split, physical solver, and final evaluation protocol are held fixed. The primary metric is relative $L_2$ error after decoding to the physical field. Because the representations have substantially different numerical scales and spectral weightings, optimizer settings are adapted when needed for stable convergence; training objectives are therefore not compared directly across representations. Final comparisons are made using validation/holdout error in the same physical variables.

A practical advantage observed during these operator-learning experiments is the lower training cost of the minimal representation. Because the input and target states contain only the intrinsic transverse channels, Latent2 requires less GPU memory for the same number of trajectories and spatial resolution, and in our runs it can complete more epochs and parameter-update steps within the same wall-clock budget.

\paragraph{2D MHD.}
In 2D MHD, Latent2 obtains the lowest physical-field error under the common holdout evaluation protocol. The Latent2 model is trained with optimization settings adapted to the scale of its representation, so this comparison reflects the best stably converged model for each representation rather than forcing identical optimizer hyperparameters.

\begin{table}[H]
\centering
\small
\begin{tabular}{lccc}
\toprule
Model & Val. 10-step mean & Holdout 10-step mean & Holdout step-10 \\
\midrule
A---Raw & 0.1343289 & 0.1350950 & 0.1847896 \\
B---Leray & 0.1332429 & 0.1338184 & 0.1917674 \\
C---Potential & 0.1005770 & 0.1012601 & 0.1465121 \\
\textbf{D---Latent2 (ours)} & \textbf{0.0993804} & \textbf{0.0996544} & \textbf{0.1421250} \\
\bottomrule
\end{tabular}
\caption{2D MHD physical relative-$L_2$ error. Lower is better.}
\label{tab:mhd2d_operator}
\end{table}

The gain over the Potential baseline is modest, but the result shows that separately compressing the velocity and magnetic fields to one transverse scalar each does not impair dynamical prediction and can slightly improve the final physical-field error after sufficient training.

\begin{figure}[H]
\centering
\includegraphics[width=\linewidth]{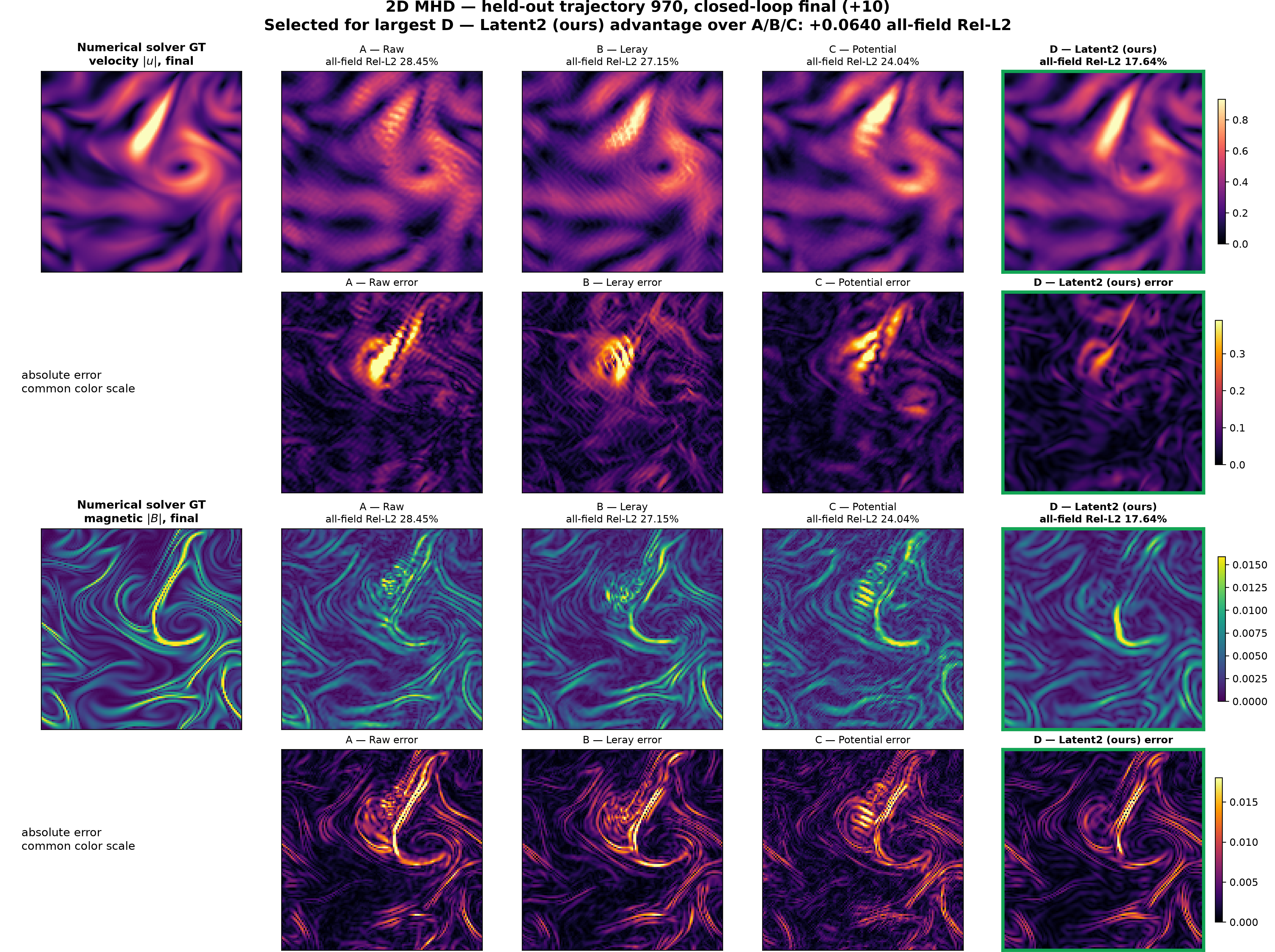}
\caption{2D MHD closed-loop prediction on a selected held-out trajectory at the final $+10$ step.  The top and middle rows compare velocity and magnetic-field magnitudes, and the lower rows show absolute errors on common color scales.  This instance is selected for its large Latent2 advantage over A/B/C; Table~\ref{tab:mhd2d_operator} remains the aggregate evaluation.}
\label{fig:mhd2d_unified}
\end{figure}

\paragraph{3D Navier--Stokes.}
The 3D NS results show a horizon-dependent trade-off. Raw and Leray are slightly better at very short horizons, whereas Latent2 becomes the lowest-error representation as the closed-loop rollout becomes longer.

\begin{table}[H]
\centering
\small
\begin{tabular}{lcccc}
\toprule
Model & $h=1$ & $h=5$ & $h=10$ & $h=20$ \\
\midrule
\textbf{A---Raw} & \textbf{0.074044} & 0.297128 & 0.643126 & 1.060287 \\
\textbf{B---Leray} & 0.075585 & \textbf{0.297055} & 0.617995 & 1.010129 \\
C---Potential & 0.078994 & 0.308620 & 0.626087 & 1.079068 \\
\textbf{D---Latent2 (ours)} & 0.082842 & 0.308565 & \textbf{0.592485} & \textbf{1.000849} \\
\bottomrule
\end{tabular}
\caption{3D NS physical relative-$L_2$ error at different rollout horizons. Lower is better.}
\label{tab:ns3d_operator}
\end{table}

Latent2 is not the best one-step predictor, and the models remain close at $h=5$. At $h=10$ and $h=20$, however, Latent2 has the lowest error. This suggests that the principal benefit of a minimal transverse state need not appear in local one-step fitting; it can instead emerge in the stability of repeated autoregressive evolution.

\begin{figure}[H]
\centering
\includegraphics[width=\linewidth]{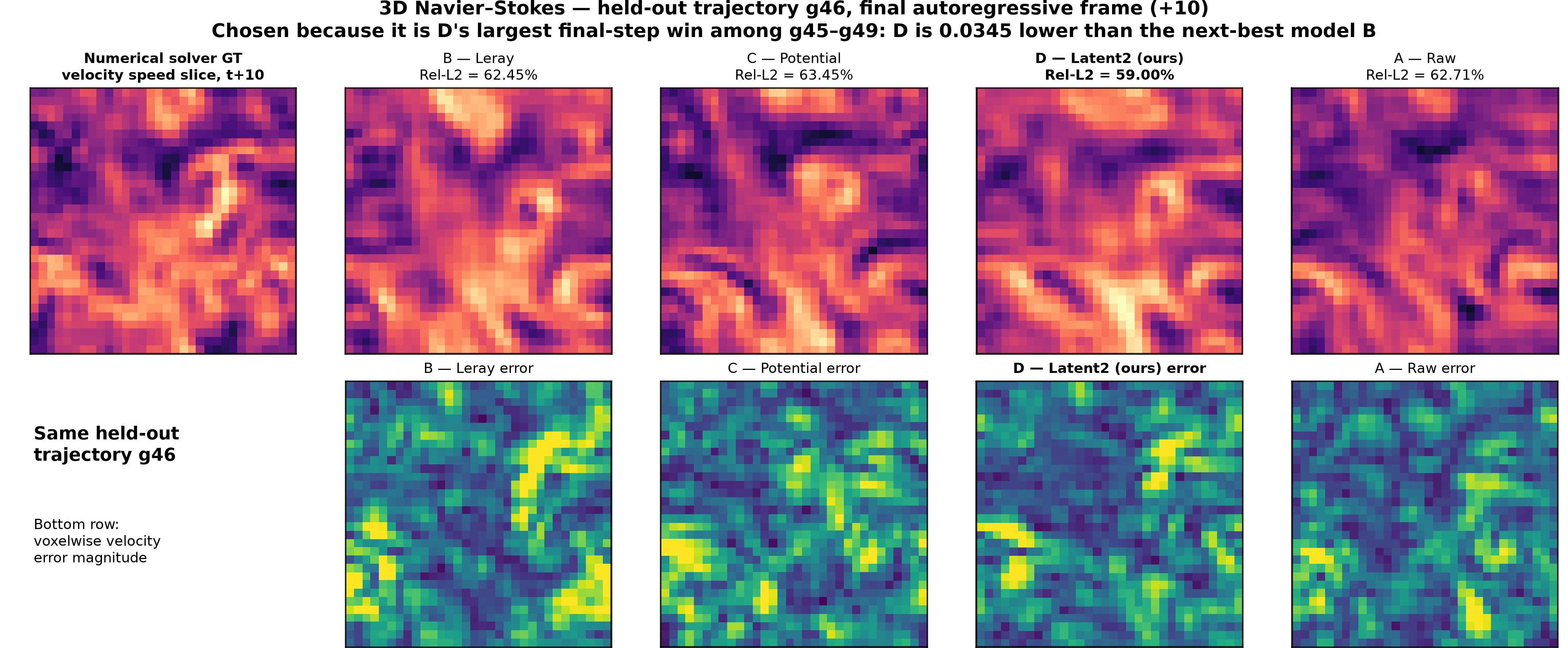}
\caption{3D Navier--Stokes final autoregressive $+10$ frame for held-out trajectory g46.  The panels show the solver velocity-speed slice, predictions from the four representations, and their voxelwise velocity-error magnitudes.  This selected trajectory illustrates the long-horizon advantage; Table~\ref{tab:ns3d_operator} reports the aggregate horizon-wise result.}
\label{fig:ns3d_unified}
\end{figure}

\paragraph{2D Navier--Stokes.}
For 2D NS, we additionally compare against FNO-vorticity. All five models are reevaluated on the same held-out initial conditions with the same input history, numerical-solver target, and closed-loop rollout, and are compared after reconstruction to physical velocity.

\begin{table}[H]
\centering
\small
\begin{tabular}{lc}
\toprule
Model & Clean step-10 velocity Rel-$L_2$ \\
\midrule
\textbf{FNO-vorticity} & \textbf{0.039315} \\
\textbf{D---Latent2 (ours)} & \textbf{0.075430} \\
B---Leray & 0.436152 \\
A---Raw & 0.596742 \\
C---Potential & 1.003959 \\
\bottomrule
\end{tabular}
\caption{2D NS clean step-10 physical velocity error under the common reevaluation protocol.}
\label{tab:ns2d_operator}
\end{table}

FNO-vorticity remains the strongest clean baseline in 2D NS. Latent2 does not surpass it, but it substantially outperforms the Raw, Leray, and Potential models in this evaluation. On an independent 50-trajectory holdout, the 10-step mean physical relative-$L_2$ error is $0.0516114$ for Latent2 and $0.0227320$ for FNO-vorticity, confirming the strength of the vorticity coordinate in two dimensions. The two scalar representations are closely related, but they induce systematically different spectral weightings in the training objective. In particular, their losses differ by wavenumber-dependent factors involving $|k|$, so the same physical prediction error can receive different emphasis across Fourier bands depending on whether the model predicts vorticity, potential variables, or the transverse latent. FNO-vorticity is therefore an especially informative baseline because it also reduces the two physical velocity components to a single scalar state while using a different spectral weighting. The exact relations and the corresponding loss formulas are given in Appendix~\ref{app:ns2d_loss_comparison}.

\begin{figure}[H]
\centering
\includegraphics[width=\linewidth]{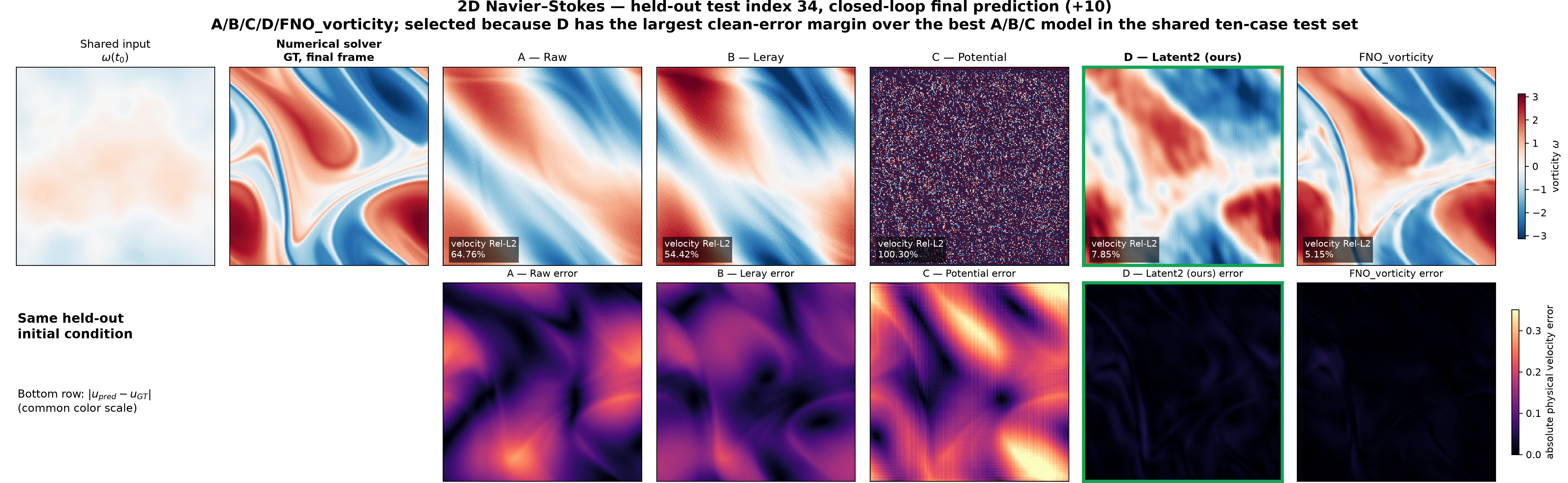}
\caption{The same 2D Navier--Stokes held-out case with the FNO-vorticity prediction included.  At this selected instance, FNO-vorticity has the lowest physical velocity error, followed by Latent2; the aggregate clean-error comparison is reported in Table~\ref{tab:ns2d_operator}.}
\label{fig:ns2d_fno_unified}
\end{figure}

\paragraph{Robustness.}
We also evaluate solver-integrated adversarial perturbations. In 3D NS, under the same perturbation optimized against the Leray model, Latent2 exhibits a much smaller absolute increase in $h=10$ error:

\begin{table}[H]
\centering
\small
\begin{tabular}{lc}
\toprule
Model & Absolute error increase at $h=10$ \\
\midrule
B---Leray & 0.001056259 \\
\textbf{D---Latent2 (ours)} & \textbf{0.000028114} \\
\bottomrule
\end{tabular}
\caption{3D NS response to the same B-targeted observation perturbation.}
\label{tab:ns3d_robustness}
\end{table}

A related transfer effect appears in 2D NS. For the same perturbation optimized against FNO-vorticity, the full-trajectory attack-loss ratio is $3.7538\times$ on FNO-vorticity but only $1.5182\times$ when the identical perturbation is transferred to Latent2. These controlled cases show that the transverse representation can substantially reduce perturbation amplification for adversarial directions optimized against another representation.

The Leray model also exposes a structural source of possible numerical sensitivity. Its raw output contains both transverse and longitudinal components, but the latter is removed by projection and is therefore only weakly constrained by the physical loss; in our current diagnostic, approximately $7.2$--$8.0\%$ of the raw response magnitude is discarded by this projection. This null-direction leakage provides a plausible mechanism for extra sensitivity during rollout, whereas Latent2 represents only the transverse degrees of freedom and has no such discarded longitudinal component.

\paragraph{Representation scale and optimization stability.}
An important practical observation is that using the same FNO architecture does not imply that identical optimizer hyperparameters are appropriate across representations. Raw velocity, potential variables, vorticity, and the transverse latent describe the same physical state but have different Fourier magnitudes and spectral scalings. Their training losses are therefore related by systematic wavenumber-dependent weighting factors: vorticity places relatively greater weight on higher wavenumbers, while potential variables shift weight toward lower wavenumbers. Consequently, changing the prediction variable changes the target distribution, gradient scale, and the relative contribution of different Fourier bands to the training objective. The exact spectral relations are summarized in Appendix~\ref{app:ns2d_loss_comparison}.

In practice, directly reusing optimization settings tuned for a physical-field FNO can severely degrade low-amplitude latent or potential models. We observed cases in which weight decay continually shrank later Fourier/spectral-layer weights toward zero, effectively collapsing the spectral branch even while the training objective continued to decrease. Reducing or removing overly strong weight decay substantially alleviated this behavior. This motivates treating representation and optimization as coupled design choices: data, solver, model capacity, and final physical-space evaluation should be controlled as tightly as possible, while each representation should be allowed optimization settings that permit stable convergence. Cross-representation conclusions should then be based on converged physical-space validation/holdout error rather than raw training objectives measured in differently scaled coordinates.

\paragraph{Overall result.}
Across the three benchmarks, the minimal transverse representation is competitive while using only the intrinsic divergence-free degrees of freedom. In 2D MHD, Latent2 gives the lowest clean physical error among the four compared representations. In 3D NS, it is not the best one-step predictor but has the lowest error at 10- and 20-step rollout horizons. In 2D NS, it remains behind the strong FNO-vorticity baseline but substantially improves over the evaluated Raw, Leray, and Potential models. Together with the controlled robustness results, these experiments indicate that removing the redundant longitudinal degree of freedom can be useful not only as an exact representation constraint, but also for long-horizon autoregressive dynamics and for sensitivity to some adversarial perturbations.

\section{Conclusion}
We introduced a minimal $D\!\to\!D-1$ transverse representation that parameterizes divergence-free vector fields directly by their intrinsic Fourier degrees of freedom, yielding an invertible and isometric coordinate system for the periodic/closed settings considered here and eliminating the longitudinal degree of freedom from operator learning. Across 2D MHD, 3D Navier--Stokes, and 2D Navier--Stokes, the representation is competitive with raw, projection-based, and potential-based alternatives, with particularly encouraging long-horizon and controlled-robustness results, while also reducing the number of state channels used during training. At the same time, the empirical improvement is not uniformly large: the construction is mathematically clean, but its practical advantage depends on the problem and optimization setting, and in 2D Navier--Stokes the vorticity formulation remains a stronger clean baseline. Future work should investigate architectures and optimization schemes designed specifically for transverse coordinates, reduce or better manage frame/gauge discontinuities in dimensions such as $D=3$, and test whether the reduced state space gives larger benefits in higher-resolution, longer-horizon, and more strongly coupled divergence-free systems.

\subsection*{AI Use Statement}
Generative AI tools were used solely for language editing of the manuscript,
including improving grammar, wording, readability, and sentence-level clarity.
They were not used to generate experimental results or alter the reported
results and conclusions. All AI-assisted edits were reviewed by the author, who
takes full responsibility for the final content of the manuscript.

\clearpage
\appendix
\section{Detailed Derivation of the Transverse Fourier Constraint}
\label{app:transverse_derivation}

Let $u:\Omega\subset\mathbb{R}^D\to\mathbb{R}^D$ and define
\begin{equation}
\widehat u(k)=\int_\Omega u(x)e^{-ik\cdot x}\,dx.
\end{equation}
We derive the constraint on $\widehat u(k)$ step by step. First, applying the product rule to the vector field $u(x)e^{-ik\cdot x}$ gives
\begin{align}
\nabla\!\cdot\!\left(u e^{-ik\cdot x}\right)
&=e^{-ik\cdot x}\nabla\!\cdot u+u\cdot\nabla e^{-ik\cdot x}. \label{eq:app_product}
\end{align}
Since
\begin{equation}
\nabla e^{-ik\cdot x}=-ik\,e^{-ik\cdot x},
\end{equation}
Eq.~\eqref{eq:app_product} becomes
\begin{equation}
\nabla\!\cdot\!\left(u e^{-ik\cdot x}\right)
=(\nabla\!\cdot u)e^{-ik\cdot x}-i(k^\top u)e^{-ik\cdot x}. \label{eq:app_expand}
\end{equation}
Integrating over $\Omega$ yields
\begin{equation}
\int_\Omega\nabla\!\cdot\!\left(u e^{-ik\cdot x}\right)dx
=\int_\Omega(\nabla\!\cdot u)e^{-ik\cdot x}\,dx
-i\int_\Omega(k^\top u)e^{-ik\cdot x}\,dx. \label{eq:app_integrate}
\end{equation}
By the divergence theorem, the left-hand side is
\begin{equation}
\int_\Omega\nabla\!\cdot\!\left(u e^{-ik\cdot x}\right)dx
=\int_{\partial\Omega}(u\cdot n)e^{-ik\cdot x}\,dS. \label{eq:app_boundary}
\end{equation}
Because $k$ is independent of $x$,
\begin{equation}
\int_\Omega(k^\top u)e^{-ik\cdot x}\,dx
=k^\top\int_\Omega u(x)e^{-ik\cdot x}\,dx
=k^\top\widehat u(k). \label{eq:app_fourier}
\end{equation}
Substituting Eqs.~\eqref{eq:app_boundary} and \eqref{eq:app_fourier} into Eq.~\eqref{eq:app_integrate} gives
\begin{equation}
\int_{\partial\Omega}(u\cdot n)e^{-ik\cdot x}\,dS
=\int_\Omega(\nabla\!\cdot u)e^{-ik\cdot x}\,dx-i k^\top\widehat u(k),
\end{equation}
and therefore
\begin{equation}
\boxed{
k^\top\widehat u(k)
=i\int_{\partial\Omega}(u\cdot n)e^{-ik\cdot x}\,dS
-i\int_\Omega(\nabla\!\cdot u)e^{-ik\cdot x}\,dx.}
\label{eq:app_master}
\end{equation}
If the field is divergence-free,
\begin{equation}
\nabla\!\cdot u=0 \quad\text{in }\Omega,
\end{equation}
Eq.~\eqref{eq:app_master} reduces to
\begin{equation}
k^\top\widehat u(k)
=i\int_{\partial\Omega}(u\cdot n)e^{-ik\cdot x}\,dS. \label{eq:app_df}
\end{equation}
Hence the second sufficient condition is the vanishing weighted boundary flux
\begin{equation}
\int_{\partial\Omega}(u\cdot n)e^{-ik\cdot x}\,dS=0. \label{eq:app_flux}
\end{equation}
For an impermeable boundary, $u\cdot n=0$ pointwise on $\partial\Omega$, so Eq.~\eqref{eq:app_flux} holds immediately. For a periodic box, opposite faces have opposite normals, while both $u$ and the periodic Fourier factor $e^{-ik\cdot x}$ match across each paired face; their contributions therefore cancel exactly. Under either boundary condition,
\begin{equation}
\boxed{k^\top\widehat u(k)=0.}
\end{equation}
Thus, for every $k\neq0$,
\begin{equation}
\widehat u(k)\in k^\perp
:=\{v\in\mathbb{C}^D:k^\top v=0\},
\end{equation}
and since $k^\perp$ is a codimension-one subspace,
\begin{equation}
\boxed{\dim k^\perp=D-1.}
\end{equation}
Therefore each nonzero Fourier mode of a divergence-free $D$-component vector field satisfying the above boundary condition has exactly $D-1$ independent degrees of freedom.

\section{Operator Structure, Kernels, Invertibility, and Preserved Properties}
\label{app:operator_structure}

This section makes the three realizations in the main text explicit.  All Fourier-space dimensions below refer to the nonzero modes; the single zero mode is stored separately as $\bar u=\widehat u(0)\in\mathbb R^D$.  Let $N_\times$ be the number of nonzero Fourier indices on the physical grid and $\widetilde N_\times$ the corresponding number on the extended grid.

\subsection{Full statement of Theorem~\ref{thm:minimal_representation}}\label{app:full_theorem}
Let $\Omega\subset\widetilde\Omega$.  The field on $\Omega$ need not be periodic.  Assume that a real divergence-free field $u$ admits a periodic divergence-free continuation on $\widetilde\Omega$, and define the nonzero-mode decoder
\begin{equation}
A_E
:=
M_D\widetilde F_D^{-1}\widetilde E\widetilde F_{D-1}.
\end{equation}
Let
\begin{equation}
\mathcal U_E:=\operatorname{Range}(A_E),
\qquad
\mathcal C_E:=(\ker A_E)^\perp=\operatorname{Range}(A_E^*).
\end{equation}
For every $u\in\mathcal U_E$, define the canonical forward transform (encoder) by
\begin{equation}
v_E^\star
:=
\arg\min_{v}\frac12\|v\|_2^2
\qquad
\mathrm{s.t.}\qquad
A_Ev=u,
\end{equation}
with the closed form
\begin{equation}
v_E^\star
=
\mathcal E_E(u)
=
A_E^\dagger u
\in\mathcal C_E.
\end{equation}
The inverse transform (decoder) is
\begin{equation}
\mathcal D_E(v):=A_Ev,
\qquad
v\in\mathcal C_E.
\end{equation}
These maps satisfy
\begin{equation}
A_EA_E^\dagger u=u,
\qquad
u\in\mathcal U_E,
\end{equation}
\begin{equation}
A_E^\dagger A_Ev=v,
\qquad
v\in\mathcal C_E,
\end{equation}
and therefore
\begin{equation}
\boxed{
\mathcal U_E
\overset{\mathcal E_E}{\underset{\mathcal D_E}{\rightleftarrows}}
\mathcal C_E
\subset
L^2(\widetilde\Omega;\mathbb R^{D-1})
}
\end{equation}
is a one-to-one reversible representation.

The zero Fourier mode is not constrained by incompressibility and is retained separately as
\begin{equation}
\bar u:=\widehat u(0)\in\mathbb R^D.
\end{equation}
Hence the complete transform is
\begin{equation}
\boxed{
u\longleftrightarrow(\bar u,v_E^\star)}.
\end{equation}
If $\widehat u(0)=0$, no additional zero-mode channel is required.

\subsection{Per-mode and global operators}
For every $k\neq0$, let
\begin{equation}
E(k)\in\mathbb R^{D\times(D-1)},
\qquad
E(k)^*E(k)=I_{D-1},
\qquad
k^\top E(k)=0.
\end{equation}
The columns of $E(k)$ form an orthonormal basis of $k^\perp$.  A full Householder reflection $Q_H(k)\in\mathbb R^{D\times D}$ may be written as
\begin{equation}
Q_H(k)=I_D-2\frac{r_kr_k^\top}{r_k^\top r_k},
\qquad
r_k=e_D-\frac{\kappa(k)}{|\kappa(k)|},
\end{equation}
with the paired convention $\kappa(-k)=\kappa(k)$; deleting the column aligned with $\kappa(k)$ gives $E(k)$.  Hence $E(-k)=E(k)$ and Hermitian symmetry is preserved.

Stacking the mode-wise bases gives the rectangular block-diagonal operator
\begin{equation}
E_\times
=\operatorname{blkdiag}_{k\neq0}E(k)
\in\mathbb C^{D N_\times\times (D-1)N_\times}.
\end{equation}
It satisfies
\begin{equation}
E_\times^*E_\times=I_{(D-1)N_\times},
\qquad
E_\times E_\times^*=P_\perp,
\end{equation}
where $P_\perp=\operatorname{blkdiag}_{k\neq0}P_k$ and $P_k=I_D-kk^\top/|k|^2$.  Thus $E_\times E_\times^*$ is the identity on the divergence-free Fourier subspace.

Let $F_{D,N}=F_N\otimes I_D$ and $F_{D-1,N}=F_N\otimes I_{D-1}$ denote the component-wise unitary discrete Fourier transforms; on the nonzero-mode subspaces we use the same symbols with the zero coefficient omitted.  Their dimensions are
\begin{equation}
F_{D,N}:\mathbb C^{DN}\to\mathbb C^{DN},
\qquad
F_{D-1,N}:\mathbb C^{(D-1)N}\to\mathbb C^{(D-1)N}.
\end{equation}
The paired symmetry $E(-k)=E(k)$ ensures that a real input field maps to Hermitian-symmetric reduced coefficients and therefore to a real $(D-1)$-component field after inverse FFT.

\subsection{Case I: direct periodic transform}
On the nonzero-mode subspaces the decoder and encoder are
\begin{equation}
A_0=F_{D,N}^{-1}E_\times F_{D-1,N},
\qquad
T_0=A_0^{-1}=A_0^*=F_{D-1,N}^{-1}E_\times^*F_{D,N}.
\end{equation}
Because the Fourier transforms are unitary and $E_\times^*E_\times=I$,
\begin{align}
A_0^*A_0
&=F_{D-1,N}^*E_\times^*E_\times F_{D-1,N}=I,\\
A_0A_0^*
&=F_{D,N}^{-1}P_\perp F_{D,N}.
\end{align}
The second identity equals the identity when restricted to the physical divergence-free subspace.  Therefore
\begin{equation}
\ker A_0=\{0\},
\qquad
A_0:\mathcal V_{D-1}\overset{1:1}{\longrightarrow}\mathcal U_{D,\mathrm{div}}
\end{equation}
is a bijection.  There is no optimization problem because $A_0v=u$ has exactly one solution.

For any $v_1,v_2$,
\begin{equation}
\langle A_0v_1,A_0v_2\rangle
=\langle v_1,v_2\rangle,
\end{equation}
so norms, distances, and angles are preserved.  This makes Case I particularly suitable as a representation or embedding for learning: with unitary normalization, the kinetic-energy norm is unchanged and, for any two samples,
\begin{equation}
\|A_0v_1-A_0v_2\|_2^2=\|v_1-v_2\|_2^2,
\end{equation}
so a squared-error discrepancy is identical before and after the transform.  The encoder--decoder pair $(T_0,A_0)$ is therefore a fixed analytic bijection from $D$ divergence-free components to $D-1$ intrinsic components, rather than a latent embedding learned by a neural network; it introduces neither reconstruction error nor representation ambiguity on the admissible subspaces.  Since spatial differentiation is the scalar Fourier multiplier $ik_j$ on each mode,
\begin{equation}
(ik_jI_D)E(k)=E(k)(ik_jI_{D-1}),
\end{equation}
and hence
\begin{equation}
\partial_{x_j}A_0=A_0\partial_{x_j},
\qquad
\partial^\alpha A_0=A_0\partial^\alpha,
\qquad
\Delta A_0=A_0\Delta.
\end{equation}
Consequently, for every Sobolev order $s$ for which the norms are defined,
\begin{equation}
\|A_0v\|_{H^s}=\|v\|_{H^s},
\qquad
\|\partial^\alpha A_0v\|_{L^2}=\|\partial^\alpha v\|_{L^2}.
\end{equation}
Because $A_0$ is time independent,
\begin{equation}
\partial_t(A_0v)=A_0\partial_tv,
\qquad
\|\partial_t(A_0v)\|_{L^2}=\|\partial_tv\|_{L^2}.
\end{equation}

The same transform gives an exact coordinate change for dynamics.  We first distinguish the two directions of the Case-I representation explicitly.  The forward reduction (encoder) maps the physical divergence-free field to intrinsic coordinates,
\begin{equation}
\boxed{v=T_0u=A_0^{-1}u=A_0^*u},
\end{equation}
whereas the inverse transform (decoder) reconstructs the physical field,
\begin{equation}
\boxed{u=A_0v=T_0^{-1}v}.
\end{equation}
Suppose the original $D$-component dynamics is
\begin{equation}
\boxed{\dot u=G_D(u)},
\qquad
G_D:\mathcal U_{\rm div}\rightarrow\mathcal U_{\rm div}.
\end{equation}
Differentiating the forward transform and substituting the inverse transform gives
\begin{equation}
\dot v=T_0\dot u
       =T_0G_D(u)
       =T_0G_D(T_0^{-1}v),
\end{equation}
so the exact reduced dynamics is
\begin{equation}
\boxed{\dot v=G_{D-1}(v)},
\qquad
\boxed{G_{D-1}=T_0\circ G_D\circ T_0^{-1}
      =A_0^{-1}\circ G_D\circ A_0},
\end{equation}
or, using $A_0^{-1}=A_0^*$,
\begin{equation}
\boxed{G_{D-1}(v)=A_0^*G_D(A_0v)}.
\end{equation}
Thus learning in the original representation means learning $G_D:u\mapsto\dot u$ on the $D$-component divergence-free state space, whereas our reduced formulation learns the conjugate operator $G_{D-1}:v\mapsto\dot v$ directly on the $(D-1)$-component intrinsic state space.
The same conjugacy holds in discrete time: if $u_{n+1}=\Phi(u_n)$, then
\begin{equation}
v_{n+1}=T_0\Phi(A_0v_n).
\end{equation}
Writing $G_{D-1}(v)=T_0G_D(A_0v)$, the continuous-time Jacobian obeys
\begin{equation}
J_{G_{D-1}}(v)=T_0J_{G_D}(u)A_0.
\end{equation}
Since $T_0=A_0^*$ is unitary on the admissible spaces, this is unitary equivalence.  Therefore
\begin{equation}
\operatorname{eig}J_{G_{D-1}}=\operatorname{eig}J_{G_D},
\qquad
\sigma_i(J_{G_{D-1}})=\sigma_i(J_{G_D}),
\qquad
\kappa(T_0)=1.
\end{equation}
Moreover,
\begin{equation}
\operatorname{tr}J_g=\operatorname{tr}J_f,
\end{equation}
so phase-space divergence, and hence a Liouville-type divergence-free identity when present in the original coordinates, is preserved.

\subsection{Case II: Fourier extension and canonicalization}
Let $\widetilde\Omega$ contain $\widetilde N>N$ spatial points.  The component-wise physical restriction is
\begin{equation}
M_D=M_x\otimes I_D
\in\mathbb R^{DN\times D\widetilde N},
\end{equation}
where $M_x$ selects the original $N$ physical locations from the extended grid.  Let
\begin{equation}
\widetilde E_\times
\in\mathbb C^{D\widetilde N_\times\times (D-1)\widetilde N_\times}
\end{equation}
be the extended block-diagonal transverse basis.  The decoder is
\begin{equation}
A_E
=M_D\widetilde F_{D,\widetilde N}^{-1}
\widetilde E_\times
\widetilde F_{D-1,\widetilde N}.
\end{equation}
The Fourier/Householder part preceding $M_D$ is injective, but $M_D$ discards the extension region.  Therefore nonzero reduced perturbations can decode to extended fields that vanish on $\Omega$, giving
\begin{equation}
\ker A_E
=\{z:M_D\widetilde F_D^{-1}\widetilde E_\times\widetilde F_{D-1}z=0\}.
\end{equation}
By rank-nullity,
\begin{equation}
\dim\ker A_E
=(D-1)\widetilde N_\times-\operatorname{rank}(A_E).
\end{equation}
If $A_E$ is surjective onto an admissible physical state space of intrinsic dimension $m$, then $\operatorname{rank}(A_E)=m$ and the nullity is $(D-1)\widetilde N_\times-m$.  Thus a single physical field has the complete inverse family
\begin{equation}
A_E^{-1}(u)=v_0+\ker A_E,
\end{equation}
which is one-to-many as an inverse relation, while the forward map is many-to-one.

The minimum-energy condition removes precisely this ambiguity:
\begin{equation}
v_E^\star
=\arg\min_v\frac12\|v\|_2^2
\quad\mathrm{s.t.}\quad A_Ev=u.
\end{equation}
Its Moore--Penrose solution is
\begin{equation}
v_E^\star=A_E^\dagger u
=A_E^*(A_EA_E^*)^{-1}u
\end{equation}
when the indicated inverse exists on the admissible range, and
\begin{equation}
v_E^\star\in\operatorname{Range}(A_E^*)=(\ker A_E)^\perp.
\end{equation}
Hence
\begin{equation}
A_E:(\ker A_E)^\perp\overset{1:1}{\longrightarrow}\operatorname{Range}(A_E)
\end{equation}
is bijective.  Numerically, if $A_E=U\Sigma V^*$, the thresholded pseudoinverse uses $1/\sigma_i$ only for $\sigma_i>\tau\sigma_{\max}$ and sets the remaining inverse singular values to zero.

Because restriction is not unitary,
\begin{equation}
A_E^*A_E
=\widetilde F_{D-1}^*\widetilde E_\times^*
\widetilde F_D M_D^*M_D\widetilde F_D^{-1}
\widetilde E_\times\widetilde F_{D-1}
\neq I
\end{equation}
in general.  Thus canonical bijectivity does not imply standard $L^2$ isometry or angle preservation.  The forward decoder still commutes with spatial derivatives in the interior of $\Omega$, but the minimum-norm encoder $A_E^\dagger$ need not commute with spatial differentiation.  Since $A_E^\dagger$ is fixed in time, however,
\begin{equation}
\partial_t(A_E^\dagger u)=A_E^\dagger\partial_tu.
\end{equation}
On the canonical subspace the reduced dynamics remains an exact coordinate representation,
\begin{equation}
\dot v=A_E^\dagger f(A_Ev),
\end{equation}
so Jacobian eigenvalues are preserved by similarity, while Euclidean singular values and conditioning generally are not.

\subsection{Case III: extension with matched spectral dimension}
Let $K$ be a conjugate-symmetric retained set with $|K|=N_\times$ nonzero modes.  The coefficient injection
\begin{equation}
J_K\in\mathbb C^{(D-1)\widetilde N_\times\times (D-1)N_\times}
\end{equation}
places the retained reduced coefficients at their extended-grid Fourier indices; $P_K^F=J_KJ_K^*$ is the corresponding spectral projector.  The decoder is
\begin{equation}
A_K
=M_D\widetilde F_D^{-1}\widetilde E_\times J_KF_{D-1,N}.
\end{equation}
Its reduced domain has the same intrinsic nonzero-mode dimension $(D-1)N_\times$ as the target divergence-free state space.  Dimension matching alone does not prove invertibility; Case III therefore chooses/accepts a mode set $K$ only when
\begin{equation}
\operatorname{rank}(A_K)=(D-1)N_\times,
\qquad
\operatorname{Range}(A_K)=\mathcal U_{D,\mathrm{adm}}.
\end{equation}
Under this full-rank condition,
\begin{equation}
\ker A_K=\{0\},
\qquad
v_K=A_K^{-1}u,
\end{equation}
so no minimum-norm optimization is needed.  The spectral selection $J_K$ commutes with spatial Fourier multipliers on the retained modes, and the fixed transform commutes with time differentiation.  However, because $M_D$ and spectral truncation remain present, $A_K$ is not generally unitary and does not automatically preserve the full-space $L^2/H^s$ geometry.

\subsection{Summary of the three realizations}
The three cases differ only in how the admissible reduced coordinates are related to the physical field:
\begin{equation}
\begin{array}{lll}
\text{Case I:} & \ker A_0=\{0\}, & v=A_0^{-1}u,\\
\text{Case II:} & \ker A_E\neq\{0\}\ \text{in general}, & v=A_E^\dagger u\in(\ker A_E)^\perp,\\
\text{Case III:} & \ker A_K=\{0\}\ \text{under the full-rank condition}, & v=A_K^{-1}u.
\end{array}
\end{equation}
Thus Case I is a unitary coordinate change, Case II is a redundant extension followed by canonicalization, and Case III removes the extension overparameterization by spectral dimension matching plus a full-rank requirement.  In all three cases, once restricted to the stated invertible spaces, the physical dynamics can be transported without information loss to $(D-1)$-component coordinates.

\section{Global Smoothness of the Transverse Frame and Empirical Effect of Gauge Seams}
\label{app:frame_smoothness}

The minimal representation requires more than identifying the transverse subspace $k^\perp$: it also requires choosing concrete coordinates inside that subspace.  For every nonzero Fourier mode, let
\begin{equation}
 n(k)=\frac{k}{|k|}\in S^{D-1},
 \qquad
 E(k)=\big[e_1(k),\ldots,e_{D-1}(k)\big],
\end{equation}
where the columns of $E(k)$ are an orthonormal basis of $k^\perp$.  Thus the construction attaches a $(D-1)$-dimensional coordinate frame to every direction on the wavevector sphere $S^{D-1}$.  Ideally, nearby wavevector directions would receive nearby frames: if $n(k_i)$ and $n(k_j)$ are close on the sphere, then $E(k_i)$ and $E(k_j)$ would also be close.

\subsection{Parallelizability and the intrinsic global obstruction}
A smooth choice of a complete tangent frame over the whole sphere is exactly the classical problem of whether the sphere is \emph{parallelizable}.  Apart from the trivial zero-dimensional case, the only parallelizable spheres are
\begin{equation}
 \boxed{S^1,\qquad S^3,\qquad S^7,}
\end{equation}
as established by the classical sphere-parallelizability results \citep{bottmilnor1958parallel,kervaire1958nonparallel}.  Since our wavevector directions lie on $S^{D-1}$, a globally smooth complete real transverse frame can therefore exist only for
\begin{equation}
 \boxed{D=2,\qquad D=4,\qquad D=8}
\end{equation}
for the nontrivial dimensions considered here.  In the common three-dimensional case, $D=3$, the direction sphere is $S^2$.  The hairy-ball obstruction is already sufficient to rule out even one globally nonvanishing continuous tangent vector field on $S^2$, and hence a globally smooth two-vector transverse frame is impossible.  Consequently, in three dimensions the loss of global smoothness is intrinsic to any single global real frame construction; replacing Householder by another single global frame can move or reshape the problematic set, but cannot remove the obstruction.

The paired reality convention used in this paper imposes the additional requirement $E(-k)=E(k)$ through a canonical representative of each pair $\{k,-k\}$.  This convention guarantees real physical-space reduced channels, but the canonical half-space selection can introduce an additional coordinate seam.  Therefore the smoothness of a particular implementation can be more restrictive than sphere parallelizability alone.

\subsection{How the obstruction appears in the Householder construction}
The Householder frame used in Appendix~\ref{app:operator_structure} fixes a reference direction $e_D$ and sets, for the canonical unit direction $n=\kappa/|\kappa|$,
\begin{equation}
 Q_H(n)=I_D-2\frac{(e_D-n)(e_D-n)^\top}{\|e_D-n\|_2^2},
\end{equation}
with a separately defined value at the aligned direction $n=e_D$.  The transverse frame $E(n)$ is obtained from the first $D-1$ columns of $Q_H(n)$.  The aligned direction is a concrete location where the coordinate choice becomes singular.  To see why assigning a special value at that single point does not make the surrounding frame continuous, approach it as
\begin{equation}
 n(\theta,\omega)
 =\cos\theta\,e_D+\sin\theta\,\omega,
 \qquad
 \omega\perp e_D,
 \qquad
 \|\omega\|_2=1.
\end{equation}
As $\theta\to0$, the normalized Householder vector approaches $-\omega$, so the limiting reflection contains the factor
\begin{equation}
 I_D-2\omega\omega^\top,
\end{equation}
which depends on the tangential approach direction $\omega$.  Thus two wavevector directions can become arbitrarily close to one another while their selected transverse coordinate frames remain separated by a finite rotation or sign change.

We quantify this purely geometric effect by the angular frame roughness
\begin{equation}
 \theta_{ij}
 =\arccos\!\left(\operatorname{clip}\!\left(n_i^\top n_j,-1,1\right)\right),
\end{equation}
\begin{equation}
 \boxed{
 R_E(i,j)
 =\frac{\|E(k_i)-E(k_j)\|_F}{\theta_{ij}}.}
 \label{eq:frame_roughness}
\end{equation}
Equation~\eqref{eq:frame_roughness} measures how much the chosen transverse coordinate axes change per unit angular displacement in Fourier space.  In our direct geometric diagnostic, the squared neighbor-to-neighbor frame difference near the reference-axis singular region was approximately $1.98$, compared with approximately $0.023$ in ordinary regions, a ratio of about $85$.  This confirms that the selected frame can vary far more rapidly near the seam even though the physical transverse subspace $k^\perp$ itself changes smoothly.

\subsection{Why frame roughness can create rough reduced Fourier coordinates}
The previous subsection concerns the coordinate frame $E(k)$ itself.  The object learned by the reduced model is instead
\begin{equation}
 \widehat a(k)=E(k)^\top\widehat u(k).
 \label{eq:rough_latent_map}
\end{equation}
These are distinct questions.  A rapidly varying frame does not automatically imply that $\widehat a(k)$ is rough for every physical field, but it makes such roughness possible.  In particular, even when two neighboring physical Fourier coefficients satisfy
\begin{equation}
 \widehat u(k_i)\approx\widehat u(k_j),
\end{equation}
a large change in $E(k)$ can still produce a large change between $\widehat a(k_i)$ and $\widehat a(k_j)$.  In elementary terms, the physical vector can change only slightly while the coordinate axes used to measure it rotate rapidly.

To isolate this mechanism, we first use a controlled Fourier field whose directional dependence is smooth by construction.  For three deterministic coordinate probes $c_m=e_m$, $m=1,2,3$, define
\begin{equation}
 P_\perp(k)=I-\frac{kk^\top}{|k|^2},
 \qquad
 h(|k|)=\exp\!\left[-\frac{(|k|-6)^2}{8}\right],
\end{equation}
\begin{equation}
 \boxed{
 \widehat u_m(k)=h(|k|)P_\perp(k)c_m,
 \qquad
 \widehat u_m(0)=0.}
 \label{eq:smooth_probe_field}
\end{equation}
Because $k^\top P_\perp(k)=0$, every probe is exactly divergence-free.  Away from $k=0$, both the radial envelope and the projector vary smoothly with the wavevector direction, so Eq.~\eqref{eq:smooth_probe_field} provides a clean control in which any additional sharp variation after encoding can be attributed to the coordinate frame rather than to native turbulent fluctuations.

For the same neighboring-mode pairs used in the geometric diagnostic, define the normalized physical and reduced differences
\begin{equation}
 D_u(i,j)
 =\frac{\|\widehat u(k_i)-\widehat u(k_j)\|_2}
 {\sqrt{\left(\|\widehat u(k_i)\|_2^2+\|\widehat u(k_j)\|_2^2\right)/2}+10^{-12}},
 \label{eq:Du_definition}
\end{equation}
\begin{equation}
 D_a(i,j)
 =\frac{\|\widehat a(k_i)-\widehat a(k_j)\|_2}
 {\sqrt{\left(\|\widehat a(k_i)\|_2^2+\|\widehat a(k_j)\|_2^2\right)/2}+10^{-12}}.
 \label{eq:Da_definition}
\end{equation}
The seam set is fixed \emph{before} looking at $D_u$ or $D_a$: we take the top $1\%$ of neighboring pairs ranked by $R_E$ as seam pairs and the lowest $50\%$ as normal pairs.  We then summarize the excess variation by
\begin{equation}
 S_u
 =\frac{\operatorname{median}(D_u\mid\mathrm{seam})}
 {\operatorname{median}(D_u\mid\mathrm{normal})},
 \qquad
 S_a
 =\frac{\operatorname{median}(D_a\mid\mathrm{seam})}
 {\operatorname{median}(D_a\mid\mathrm{normal})},
 \label{eq:Su_Sa_definition}
\end{equation}
and report
\begin{equation}
 \boxed{\Gamma=\frac{S_a}{S_u}.}
 \label{eq:gamma_roughness}
\end{equation}
Here $S_u$ asks how much rougher the original Fourier field is near the geometric seam than in ordinary regions, while $S_a$ asks the same question after the transverse coordinate transform.  Thus $\Gamma>1$ is direct evidence that the coordinate choice adds seam-localized variation beyond that already present in the physical Fourier field.  The three smooth probes expose this effect clearly, confirming that the mechanism is real when the underlying Fourier field is locally smooth.

\subsection{Real forced-HIT data: native spectral variability dominates the added seam effect}
We next test whether the same mechanism is appreciable on actual turbulent data.  We use the publicly released coarsened filtered-DNS data accompanying the Implicit Adaptive Fourier Neural Operator (IAFNO) study \citep{jiang2026iafno}.  The benchmark is \emph{forced homogeneous isotropic turbulence} (forced HIT) in a periodic three-dimensional box.  The public forced-HIT file used in our experiments, \path{FGR1_nofilter_vel_50g_600p_gap200_LES_32.npy}, contains $50$ trajectories, $600$ saved snapshots per trajectory, a $32^3$ spatial grid, and three velocity components.

For the frame-smoothness diagnostic we use five held-out trajectories and sample twenty temporally separated snapshots from each trajectory, giving
\begin{equation}
 \boxed{5\times20=100\ \mathrm{snapshots}.}
\end{equation}
For every snapshot we compute $\widehat u(k)$, encode $\widehat a(k)=E(k)^\top\widehat u(k)$, and evaluate Eqs.~\eqref{eq:Du_definition}--\eqref{eq:gamma_roughness} on exactly the same preselected seam and normal pairs.  The seam set is never re-selected from the turbulent data.  Sampling across five held-out trajectories prevents the conclusion from depending on a single turbulent realization, and the twenty separated times per trajectory prevent it from depending on a single instant.  We additionally use $1000$ bootstrap resamples of the snapshot-level statistic to estimate a $95\%$ confidence interval.

The resulting real-data amplification ratio is
\begin{equation}
 \boxed{
 \Gamma_{\mathrm{HIT}}
 =\frac{S_a^{\mathrm{HIT}}}{S_u^{\mathrm{HIT}}}
 \approx0.9957,}
\end{equation}
with a $95\%$ bootstrap interval of approximately
\begin{equation}
 \boxed{[0.9903,\,1.0051].}
\end{equation}
The interval contains $1$ and the point estimate is itself essentially unity.  Therefore, for these forced-HIT snapshots, we do not observe a systematic increase in seam-localized Fourier roughness after encoding.

The reason is visible in the denominator of this comparison.  In the synthetic probes, neighboring physical Fourier coefficients are deliberately made locally smooth, so a rapid change of coordinate frame stands out strongly.  In an individual turbulent snapshot, neighboring Fourier coefficients already exhibit substantial mode-to-mode variability in amplitude, phase, and transverse direction.  The additional variation introduced by the Householder seam is therefore small relative to the native spectral variability.  The empirical conclusion is consequently more specific than the topological statement: the global frame obstruction is genuine and can be exposed on a smooth control field, while its additional roughness is weak on the tested real forced-HIT distribution.

\subsection{Scope of the limitation}
The seam discussed above is a loss of smoothness with respect to the \emph{wavevector direction}; it should not be interpreted as a literal discontinuity of the finite-resolution reduced field in physical space.  At fixed numerical resolution, each reduced channel is a finite Fourier series and is therefore smooth in $x$.  Moreover, the modewise transform remains orthonormal on $k^\perp$, so the direct periodic representation remains exactly invertible and preserves the represented modal energy.  The practical concern is instead that the coordinate description can make neighboring Fourier coefficients vary more rapidly, can complicate the representation of spatial rotations, and can produce a less localized physical-space transform kernel.  Our controlled experiment confirms that this coordinate effect exists, whereas the forced-HIT experiment shows that its additional Fourier roughness is negligible relative to the natural mode-to-mode variability of the tested turbulent data.

\subsection{Globally smooth direct transverse frames in $D=2,4,8$}
The preceding nullspace calculations are algebraic, while a separate geometric issue arises when a minimal representation chooses an explicit orthonormal basis of $k^\perp$ for every nonzero wavevector.  Writing $n=k/|k|\in S^{D-1}$, such a choice is a global orthonormal tangent frame on $S^{D-1}$.  The classical sphere-parallelizability result implies that, among the nontrivial cases, a globally smooth complete frame exists precisely for
\begin{equation}
\boxed{D=2,\qquad D=4,\qquad D=8,}
\end{equation}
because $S^1$, $S^3$, and $S^7$ are parallelizable \citep{bottmilnor1958parallel,kervaire1958nonparallel}.  In contrast, for $D=3$ the direction sphere is $S^2$, which is not parallelizable.  Hence no construction can choose two everywhere continuous, everywhere nonvanishing, linearly independent real vectors spanning $k^\perp$ over all directions.  A Householder frame can place the problematic set at a convenient reference direction, but in three dimensions the existence of at least one singular direction, seam, or discontinuous transition is intrinsic and cannot be removed by changing the reference vector or by replacing Householder with another single global real frame.

For $D=2$, a global frame is immediate.  For $k=(k_1,k_2)^\top\neq0$, define
\begin{equation}
\boxed{
E_2(k)=\frac{1}{|k|}
\begin{bmatrix}
-k_2\\
 k_1
\end{bmatrix}.}
\label{eq:global_frame_D2}
\end{equation}
Then $k^\top E_2(k)=0$ and $E_2(k)^\top E_2(k)=1$.  Equation~\eqref{eq:global_frame_D2} is smooth for every $k\neq0$ and therefore gives a single globally smooth transverse coordinate on $S^1$, with no distinguished pole.

For $D=4$, the same idea has a particularly simple quaternionic realization.  Identify
\begin{equation}
q(k)=k_1+k_2\mathbf{i}+k_3\mathbf{j}+k_4\mathbf{k}\in\mathbb H,
\end{equation}
where $\mathbf{i}^2=\mathbf{j}^2=\mathbf{k}^2=-1$ and $\mathbf{i}\mathbf{j}=\mathbf{k}$, $\mathbf{j}\mathbf{k}=\mathbf{i}$, $\mathbf{k}\mathbf{i}=\mathbf{j}$.  Right multiplication by the three imaginary quaternion units produces the three tangent directions $q\mathbf{i}$, $q\mathbf{j}$, and $q\mathbf{k}$.  In real coordinates this gives
\begin{equation}
\boxed{
E_4(k)=\frac{1}{|k|}
\begin{bmatrix}
-k_2 & -k_3 & -k_4\\
 k_1 & -k_4 &  k_3\\
 k_4 &  k_1 & -k_2\\
-k_3 &  k_2 &  k_1
\end{bmatrix}.}
\label{eq:global_frame_D4}
\end{equation}
A direct calculation gives
\begin{equation}
k^\top E_4(k)=0,
\qquad
E_4(k)^\top E_4(k)=I_3.
\end{equation}
Equivalently, multiplication by a unit quaternion is an orthogonal map, so it carries the orthonormal set $\{\mathbf{i},\mathbf{j},\mathbf{k}\}$ to an orthonormal basis of the tangent space at $q/|q|$.  Every entry of Eq.~\eqref{eq:global_frame_D4} is linear in $k$ divided only by $|k|$; on $S^3$ the denominator is identically one.  Thus this quaternion construction is globally smooth on the entire direction sphere, with no Householder reference-direction singularity, polar singularity, or chart seam.  In four dimensions, Householder is therefore optional rather than necessary: the quaternion frame provides a simpler global construction that does not introduce the discontinuity associated with a fixed Householder reference direction.

For $D=8$, the corresponding global construction is obtained from the octonions.  Write an octonion as a Cayley--Dickson pair of quaternions,
\begin{equation}
\mathbb O=\mathbb H\oplus\mathbb H\ell,
\qquad
(a,b)(c,d)=\bigl(ac-\overline d\,b,\;da+b\overline c\bigr),
\end{equation}
and use the basis
\begin{equation}
\begin{aligned}
1,\quad e_1&=(\mathbf{i},0),\quad e_2=(\mathbf{j},0),\quad e_3=(\mathbf{k},0),\quad e_4=(0,1),\\
e_5&=(0,\mathbf{i}),\quad e_6=(0,\mathbf{j}),\quad e_7=(0,\mathbf{k}).
\end{aligned}
\end{equation}
For
\begin{equation}
q(k)=k_1+k_2e_1+k_3e_2+k_4e_3+k_5e_4+k_6e_5+k_7e_6+k_8e_7,
\end{equation}
take the seven columns to be the real coordinate vectors of $qe_1,\ldots,qe_7$.  With the above multiplication convention this yields
\begin{equation}
\boxed{
E_8(k)=\frac{1}{|k|}
\begin{bmatrix}
-k_2 & -k_3 & -k_4 & -k_5 & -k_6 & -k_7 & -k_8\\
 k_1 & -k_4 &  k_3 & -k_6 &  k_5 &  k_8 & -k_7\\
 k_4 &  k_1 & -k_2 & -k_7 & -k_8 &  k_5 &  k_6\\
-k_3 &  k_2 &  k_1 & -k_8 &  k_7 & -k_6 &  k_5\\
 k_6 &  k_7 &  k_8 &  k_1 & -k_2 & -k_3 & -k_4\\
-k_5 &  k_8 & -k_7 &  k_2 &  k_1 &  k_4 & -k_3\\
-k_8 & -k_5 &  k_6 &  k_3 & -k_4 &  k_1 &  k_2\\
 k_7 & -k_6 & -k_5 &  k_4 &  k_3 & -k_2 &  k_1
\end{bmatrix}.}
\label{eq:global_frame_D8}
\end{equation}
The octonion norm is multiplicative.  Therefore, for unit $q$, multiplication by $q$ preserves Euclidean inner products, and the seven vectors $qe_1,\ldots,qe_7$ are mutually orthonormal and orthogonal to $q$.  Consequently,
\begin{equation}
k^\top E_8(k)=0,
\qquad
E_8(k)^\top E_8(k)=I_7.
\end{equation}
As in the quaternion case, Eq.~\eqref{eq:global_frame_D8} is a single algebraic formula on all of $\mathbb R^8\setminus\{0\}$ and hence a globally smooth frame on $S^7$.  It has no distinguished reference axis, no pole, and no Householder-type singularity or chart transition.  Different standard sign conventions for the octonion multiplication merely permute or sign-flip the seven columns and give equivalent global frames.

Thus the three special dimensions admit explicit direct constructions,
\begin{equation}
D=2:\ \mathbb C,\qquad
D=4:\ \mathbb H,\qquad
D=8:\ \mathbb O,
\end{equation}
that span $k^\perp$ globally and smoothly.  In $D=4$ and $D=8$, there is consequently no need to force the transverse coordinates through a Householder construction: the quaternionic and octonionic frames avoid the artificial singular direction introduced by that particular parameterization.  The impossibility in $D=3$ is qualitatively different because it is topological rather than a defect of Householder.  These smoothness statements concern the frame on the oriented direction sphere itself; if an implementation additionally imposes an antipodal pairing convention such as $E(-k)=E(k)$ to obtain real reduced channels, that extra gauge convention is a separate constraint and can introduce its own seam even though the quaternionic or octonionic frame itself is globally smooth.

\section{Fourier-Space Derivation and Explicit Redundancy of the NCL Parameterizations}
\label{app:ncl_redundancy}

This appendix starts from the two constructions in Neural Conservation Laws \citep{richterpowell2022ncl} and derives their Fourier-space form step by step. NCL uses the dimension symbol $n$; we write $D$ to match the notation of this paper. The NCL formulas below are their constructions, while the Fourier-space nullspace analysis is our derivation.

\subsection{NCL matrix-field construction in physical space}
NCL begins with differential forms. For an arbitrary $(D-2)$-form $\mu$, the identity $d^2=0$ gives
\begin{equation}
0=d^2\mu=d(d\mu),
\end{equation}
and NCL defines
\begin{equation}
v=\star d\mu,
\label{eq:ncl_original_form}
\end{equation}
which is divergence-free. Writing
\begin{equation}
\mu=\frac12\sum_{i,j=1}^D \mu_{ij}\,\star(dx_i\wedge dx_j),
\qquad \mu_{ji}=-\mu_{ij},
\end{equation}
gives the coordinate expression
\begin{equation}
v_i(x)=\sum_{j=1}^D \frac{\partial \mu_{ij}(x)}{\partial x_j}.
\end{equation}
NCL then sets $A_{ij}=\mu_{ij}$, so $A^\top=-A$, and obtains its first practical construction,
\begin{equation}
\boxed{
 v_i(x)=\sum_{j=1}^D \partial_j A_{ij}(x)
 }
\qquad\Longleftrightarrow\qquad
v=\operatorname{div}_{\rm row}A.
\label{eq:ncl_matrix_physical_detailed}
\end{equation}
The divergence vanishes directly because mixed derivatives commute and $A_{ji}=-A_{ij}$:
\begin{align}
\nabla\cdot v
&=\sum_{i,j}\partial_i\partial_j A_{ij}
=\frac12\sum_{i,j}\partial_i\partial_j\left(A_{ij}+A_{ji}\right)=0.
\label{eq:ncl_matrix_divzero}
\end{align}
Thus the matrix field has $D(D-1)/2$ independent scalar components, but its decoded field is constrained to the divergence-free subspace.

\subsection{Fourier transform of the matrix-field construction}
Use the Fourier convention
\begin{equation}
\widehat f(k)=\int f(x)e^{-ik\cdot x}\,dx,
\end{equation}
for which differentiation becomes multiplication by $ik_j$,
\begin{equation}
\widehat{\partial_j f}(k)=ik_j\widehat f(k).
\label{eq:fourier_derivative_rule}
\end{equation}
Applying Eq.~\eqref{eq:fourier_derivative_rule} to Eq.~\eqref{eq:ncl_matrix_physical_detailed} gives, component by component,
\begin{align}
\widehat v_i(k)
&=\sum_{j=1}^D ik_j\widehat A_{ij}(k)
=i\sum_{j=1}^D \widehat A_{ij}(k)k_j.
\end{align}
Stacking all components yields the matrix-vector product
\begin{equation}
\boxed{\widehat v(k)=i\widehat A(k)k.}
\label{eq:ncl_matrix_fourier_detailed}
\end{equation}
Because $\widehat A^\top=-\widehat A$,
\begin{equation}
k^\top\widehat v(k)=i\,k^\top\widehat A(k)k=0,
\end{equation}
so every decoded mode lies in $k^\perp$. For each nonzero $k$, NCL's matrix decoder is therefore the linear map
\begin{equation}
T_k:\mathfrak{so}(D)\rightarrow k^\perp,
\qquad
T_k(\widehat A)=i\widehat A k.
\label{eq:ncl_Tk_detailed}
\end{equation}

\subsection{Where the matrix-field redundancy is located}
The redundancy is exactly the kernel of $T_k$. Let $C(k)$ be any antisymmetric matrix satisfying
\begin{equation}
C(k)^\top=-C(k),
\qquad
C(k)k=0.
\label{eq:ncl_C_condition}
\end{equation}
Then adding $C$ changes the representation but not the physical field:
\begin{align}
T_k(\widehat A+C)
&=i(\widehat A+C)k
=i\widehat A k+iCk
=T_k(\widehat A).
\label{eq:ncl_add_redundancy}
\end{align}
Hence
\begin{equation}
\boxed{
\ker T_k=\{C\in\mathfrak{so}(D):Ck=0\}.
}
\label{eq:ncl_matrix_kernel}
\end{equation}
For a desired transverse mode $\widehat v\in k^\perp$, one explicit representation is
\begin{equation}
\widehat A_0(k)
=-\frac{i}{|k|^2}\left(\widehat v(k)k^\top-k\widehat v(k)^\top\right),
\label{eq:ncl_A0_detailed}
\end{equation}
because $i\widehat A_0k=\widehat v$. Therefore the complete family of matrix representations of the \emph{same} physical mode is
\begin{equation}
\boxed{
\widehat A(k)=\widehat A_0(k)+C(k),
\qquad C(k)^\top=-C(k),\quad C(k)k=0.
}
\label{eq:ncl_A_family_detailed}
\end{equation}
Equation~\eqref{eq:ncl_A_family_detailed} shows explicitly how redundancy can be added: $C(k)$ may vary arbitrarily inside this nullspace while $\widehat v(k)$ remains exactly unchanged.

To make this visible entry by entry, rotate coordinates so that $k=|k|e_D$. Every antisymmetric matrix has the block form
\begin{equation}
\widehat A=
\begin{pmatrix}
B & a\\
-a^\top & 0
\end{pmatrix},
\qquad B^\top=-B,
\qquad a\in\mathbb C^{D-1}.
\label{eq:ncl_block_detailed}
\end{equation}
Multiplication by $k$ gives
\begin{equation}
\widehat A k
=|k|
\begin{pmatrix}a\\0\end{pmatrix},
\qquad
\widehat v
=i|k|
\begin{pmatrix}a\\0\end{pmatrix}.
\label{eq:ncl_block_output_detailed}
\end{equation}
Thus only $a$ affects the physical field; the entire block $B\in\mathfrak{so}(D-1)$ is invisible. Equivalently, one may add
\begin{equation}
C=
\begin{pmatrix}
\Delta B & 0\\
0 & 0
\end{pmatrix},
\qquad \Delta B^\top=-\Delta B,
\label{eq:ncl_block_added}
\end{equation}
with arbitrary $\Delta B$, and the output does not change because $Ck=0$.

The rank is $D-1$ and therefore, by rank-nullity,
\begin{equation}
\boxed{
\dim\ker T_k
=\frac{D(D-1)}2-(D-1)
=\frac{(D-1)(D-2)}2.
}
\label{eq:ncl_matrix_nullity_detailed}
\end{equation}
The matrix representation is minimal only in $D=2$; the invisible dimensions are $1$ in $D=3$, $3$ in $D=4$, and grow quadratically with $D$.

\paragraph{Explicit 3D example.}
For $k=\kappa e_3$, write
\begin{equation}
\widehat A=
\begin{pmatrix}
0 & c & a_1\\
-c & 0 & a_2\\
-a_1 & -a_2 & 0
\end{pmatrix}.
\end{equation}
Then
\begin{equation}
\widehat v=i\widehat A k
=i\kappa
\begin{pmatrix}a_1\\a_2\\0\end{pmatrix},
\end{equation}
which is completely independent of $c$. Thus $c$ can be replaced by $c+\delta c$ for any $\delta c$ without changing the physical Fourier mode. In this coordinate system the redundant addition is explicitly
\begin{equation}
C=
\begin{pmatrix}
0 & \delta c & 0\\
-\delta c & 0 & 0\\
0 & 0 & 0
\end{pmatrix},
\qquad Ck=0.
\end{equation}

\subsection{NCL vector-field construction in physical space}
NCL's second construction starts from an arbitrary vector field $b:\mathbb R^D\to\mathbb R^D$. In differential-form notation NCL sets $\mu=\delta\nu$ and obtains
\begin{equation}
v=\star d\delta\nu.
\end{equation}
Writing $\nu$ in the $(D-1)$-form basis gives, up to the sign convention of the differential-form operators,
\begin{equation}
\delta\nu
=\frac12\sum_{i,j=1}^D
\left(
\frac{\partial \nu_i}{\partial x_j}
-\frac{\partial \nu_j}{\partial x_i}
\right)
\star(dx_i\wedge dx_j).
\end{equation}
NCL therefore gives the practical matrix formula
\begin{equation}
\boxed{A=J_b-J_b^\top,}
\label{eq:ncl_vector_original}
\end{equation}
where $(J_b)_{ij}=\partial_j b_i$. Equation~\eqref{eq:ncl_vector_original} by itself only defines the antisymmetric matrix field $A$; the physical vector field is obtained only after applying the row-wise divergence decoder
\begin{equation}
v_i=\sum_{j=1}^{D}\partial_j A_{ij}.
\label{eq:ncl_vector_row_div}
\end{equation}
Substituting the entries of Eq.~\eqref{eq:ncl_vector_original},
\begin{equation}
A_{ij}=\partial_j b_i-\partial_i b_j,
\end{equation}
into Eq.~\eqref{eq:ncl_vector_row_div} gives, component by component,
\begin{align}
v_i
&=\sum_j\partial_j\left(\partial_j b_i-\partial_i b_j\right)\\
&=\sum_j\partial_j^2 b_i
-\partial_i\sum_j\partial_j b_j\\
&=\Delta b_i-\partial_i(\nabla\cdot b).
\end{align}
Thus Eq.~\eqref{eq:ncl_vector_physical_operator} below is not an independent assumption and is not equivalent to Eq.~\eqref{eq:ncl_vector_original} alone; it is exactly the decoded vector field obtained from Eq.~\eqref{eq:ncl_vector_original} together with the row-wise divergence in Eq.~\eqref{eq:ncl_vector_row_div}:
\begin{equation}
\boxed{v=\Delta b-\nabla(\nabla\cdot b)
=\left(\Delta I-\nabla\nabla^\top\right)b.}
\label{eq:ncl_vector_physical_operator}
\end{equation}
Here $I$ is the $D\times D$ identity matrix and the operator $\nabla\nabla^\top$ acts on a vector field as
\begin{equation}
(\nabla\nabla^\top)b=\nabla(\nabla\cdot b).
\label{eq:grad_div_definition}
\end{equation}

Before transforming the matrix formula, it is useful to Fourier-transform Eq.~\eqref{eq:ncl_vector_physical_operator} directly. With the convention $\widehat{\partial_j f}(k)=i k_j\widehat f(k)$, where the lowercase $i=\sqrt{-1}$ is the imaginary unit,
\begin{equation}
\widehat{\Delta b}(k)=-|k|^2\widehat b(k),
\qquad
\widehat{\nabla(\nabla\cdot b)}(k)=-kk^\top\widehat b(k).
\end{equation}
Therefore
\begin{align}
\widehat v(k)
&=\left(-|k|^2I+kk^\top\right)\widehat b(k)\\
&=-|k|^2\left(I-\frac{kk^\top}{|k|^2}\right)\widehat b(k)\\
&=\boxed{-|k|^2P_k\widehat b(k)},
\qquad
P_k=I-\frac{kk^\top}{|k|^2}.
\label{eq:ncl_vector_fourier_direct}
\end{align}
The matrix $P_k$ is exactly the orthogonal projector onto $k^\perp$: it satisfies
\begin{equation}
P_k^\top=P_k,\qquad P_k^2=P_k,\qquad P_k k=0,
\end{equation}
and for every $z$ with $k^\top z=0$ one has $P_k z=z$. Hence the projector in Eq.~\eqref{eq:ncl_vector_fourier_direct} is not introduced by analogy; it follows algebraically from Fourier-transforming the physical-space operator in Eq.~\eqref{eq:ncl_vector_physical_operator}.

\subsection{Fourier transform of the vector-field construction}
The same Fourier formula can be derived independently from the matrix representation in Eq.~\eqref{eq:ncl_vector_original}. From the derivative rule, the Fourier transform of the Jacobian is
\begin{equation}
\widehat{J_b}(k)=i\widehat b(k)k^\top.
\end{equation}
Hence Eq.~\eqref{eq:ncl_vector_original} becomes
\begin{align}
\widehat A(k)
&=i\widehat b(k)k^\top-i k\widehat b(k)^\top\\
&=i\left(\widehat b(k)k^\top-k\widehat b(k)^\top\right).
\label{eq:ncl_vector_A_fourier_detailed}
\end{align}
Substituting this into the matrix decoder $\widehat v=i\widehat A k$ gives
\begin{align}
\widehat v(k)
&=i\,i\left(\widehat b k^\top-k\widehat b^\top\right)k\\
&=-\left(|k|^2\widehat b-k(k^\top\widehat b)\right)\\
&=-\left(|k|^2I-kk^\top\right)\widehat b\\
&=\boxed{-|k|^2P_k\widehat b(k)},
\qquad
P_k=I-\frac{kk^\top}{|k|^2}.
\label{eq:ncl_vector_fourier_detailed}
\end{align}
Thus the NCL vector construction is exactly a mode-wise Leray projection of the arbitrary vector $\widehat b(k)$, followed by the scalar factor $-|k|^2$.

\subsection{Where the vector-field redundancy is located}
Decompose the arbitrary vector into transverse and longitudinal parts,
\begin{equation}
\widehat b(k)=\widehat b_\perp(k)+\alpha(k)k,
\qquad
k^\top\widehat b_\perp=0.
\end{equation}
Since $P_k k=0$ and $P_k\widehat b_\perp=\widehat b_\perp$,
\begin{equation}
\widehat v(k)=-|k|^2\widehat b_\perp(k).
\end{equation}
The entire longitudinal component $\alpha(k)k$ is therefore invisible. For a prescribed physical mode $\widehat v\in k^\perp$, one particular representation is
\begin{equation}
\widehat b_0(k)=-\frac{\widehat v(k)}{|k|^2},
\end{equation}
and every representation of the same mode is
\begin{equation}
\boxed{
\widehat b(k)=\widehat b_0(k)+\alpha(k)k,
\qquad \alpha(k)\in\mathbb C.
}
\label{eq:ncl_b_family_detailed}
\end{equation}
This is the explicit many-to-one direction: changing $\alpha(k)$ changes the NCL representation but leaves $\widehat v(k)$ exactly unchanged.

For $k=\kappa e_D$ this becomes particularly transparent:
\begin{equation}
\widehat b=(b_1,\ldots,b_{D-1},b_D)^\top,
\qquad
\widehat v=-\kappa^2(b_1,\ldots,b_{D-1},0)^\top.
\end{equation}
The last component $b_D$ can take any value without affecting the output. In three dimensions, for example,
\begin{equation}
(b_1,b_2,b_3)
\quad\text{and}\quad
(b_1,b_2,b_3+\delta)
\end{equation}
produce exactly the same $\widehat v$ for every $\delta$.

\subsection{Interpretation of the two nullspaces}
The two NCL constructions reach the same transverse physical space but carry different redundant variables. The matrix construction maps
\begin{equation}
\mathfrak{so}(D)\ni\widehat A
\longmapsto i\widehat A k\in k^\perp
\end{equation}
with a $(D-1)(D-2)/2$-dimensional nullspace, while the vector construction maps
\begin{equation}
\mathbb C^D\ni\widehat b
\longmapsto -|k|^2P_k\widehat b\in k^\perp
\end{equation}
with the one-dimensional nullspace $\operatorname{span}\{k\}$. The matrix construction is therefore not literally a projection of a vector; it is a surjective linear decoder from antisymmetric matrices with a large kernel. The vector construction is literally a scaled Leray projection. In both cases many distinct internal representations correspond to one identical physical Fourier mode, whereas a direct transverse parameterization uses only the $D-1$ coordinates of $k^\perp$.

Finally, at $k=0$ all derivative factors vanish, so both derivative-based NCL constructions produce zero for that mode; this is the constant-mode exception also reflected in NCL's universality statement.

\section{DFK as a Finite-Basis Realization of the NCL Vector Construction}
\label{app:dfk_ncl_relation}

This appendix makes explicit the mathematical relation between the DFK construction \citep{ni2025dfk} and the vector-field parameterization of Neural Conservation Laws \citep{richterpowell2022ncl} derived in Appendix~\ref{app:ncl_redundancy}. The connection is established algebraically rather than by analogy. The DFK construction represents a free $D$-component vector field with a finite collection of localized radial kernels and then applies the second-order transverse operator $-I\Delta+\nabla\nabla^\top$. The NCL vector-$b$ construction represents a free $D$-component vector field with a neural parameterization and applies the opposite-sign operator $\Delta I-\nabla\nabla^\top$. Thus, after identifying the two free vector fields, the decoded divergence-free fields differ only by an overall sign. In three dimensions, the DFK operator also admits a double-curl rewrite; this is an algebraic consequence of the standard vector identity and is derived below. A single curl is already sufficient for the divergence-free guarantee.

\subsection{DFK: kernel expansion and the double-curl identity}
The DFK construction \citep{ni2025dfk} starts from a scalar radial kernel $\phi_i(x)$ and defines the matrix-valued divergence-free kernel
\begin{equation}
\Psi_i(x)
=
\left(-I\Delta+\nabla\nabla^\top\right)\phi_i(x).
\label{eq:ni_dfk_kernel}
\end{equation}
With a constant vector coefficient $\omega_i\in\mathbb R^D$, one basis contribution is
\begin{equation}
u_i^{\rm DFK}(x)
=
\Psi_i(x)\omega_i
=
\left(-I\Delta+\nabla\nabla^\top\right)
\bigl(\phi_i(x)\omega_i\bigr),
\label{eq:ni_dfk_basis}
\end{equation}
and the reconstructed field is
\begin{equation}
u_{\rm DFK}(x)
=
\sum_{i=1}^{M}u_i^{\rm DFK}(x).
\label{eq:ni_dfk_sum}
\end{equation}
We use two typographically similar but different symbols throughout this appendix. The uppercase $I$ denotes the $D\times D$ identity matrix, whereas the lowercase $i=\sqrt{-1}$ that appears after Fourier transformation is the imaginary unit. Since the Laplacian is a scalar differential operator applied componentwise,
\begin{equation}
I\Delta=\Delta I,
\end{equation}
so the notations $-I\Delta$ and $-\Delta I$ mean exactly the same operator. Likewise,
\begin{equation}
(\nabla\nabla^\top)F=\nabla(\nabla\cdot F).
\label{eq:grad_div_F}
\end{equation}
With these conventions, the divergence of the DFK operator is
\begin{align}
\nabla\cdot\left[\left(-\Delta I+\nabla\nabla^\top\right)F\right]
&=-\Delta(\nabla\cdot F)
+\nabla\cdot\nabla(\nabla\cdot F)\\
&=-\Delta(\nabla\cdot F)+\Delta(\nabla\cdot F)=0.
\label{eq:dfk_divergence_zero_explicit}
\end{align}
Thus every basis contribution and every linear combination are divergence-free.

In three dimensions, the same second-order operator has a double-curl representation. Componentwise,
\begin{align}
\bigl[\nabla\times(\nabla\times F)\bigr]_i
&=\partial_i\left(\sum_j\partial_jF_j\right)
-\sum_j\partial_j^2F_i\\
&=\bigl[\nabla(\nabla\cdot F)-\Delta F\bigr]_i.
\end{align}
Therefore
\begin{equation}
\boxed{
\nabla\times(\nabla\times F)
=
\nabla(\nabla\cdot F)-\Delta F
=
\left(-\Delta I+\nabla\nabla^\top\right)F.}
\label{eq:double_curl_identity}
\end{equation}
Equation~\eqref{eq:dfk_divergence_zero_explicit} and Eq.~\eqref{eq:double_curl_identity} are related but are not the same statement: Eq.~\eqref{eq:double_curl_identity} identifies the full operator in three dimensions, while Eq.~\eqref{eq:dfk_divergence_zero_explicit} states only that its output has zero divergence. Equation~\eqref{eq:double_curl_identity} immediately implies the divergence-free property because the divergence of any curl vanishes.

Setting $F=\phi_i\omega_i$ gives
\begin{equation}
\boxed{
u_i^{\rm DFK}
=
\nabla\times\nabla\times(\phi_i\omega_i)}.
\label{eq:ni_dfk_double_curl}
\end{equation}
Thus the ``double-curl'' description is simply an algebraic rewrite of the second-order DFK operator in three dimensions; the DFK itself is introduced through Eq.~\eqref{eq:ni_dfk_kernel}.

The Fourier-space form follows directly from the same identity. Under $\widehat{\nabla\times F}(k)=ik\times\widehat F(k)$, each spatial derivative contributes one factor of the lowercase imaginary unit $i$. Hence
\begin{align}
\widehat{\nabla\times\nabla\times F}(k)
&=(ik)\times\left[(ik)\times\widehat F(k)\right]\\
&=-k\times\left(k\times\widehat F(k)\right)\\
&=-\left[k\bigl(k^\top\widehat F(k)\bigr)-|k|^2\widehat F(k)\right]\\
&=\left(|k|^2I-kk^\top\right)\widehat F(k)\\
&=\boxed{|k|^2P_k\widehat F(k)},
\qquad
P_k=I-\frac{kk^\top}{|k|^2}.
\label{eq:double_curl_fourier_projector}
\end{align}
The penultimate equality is the vector triple-product identity $k\times(k\times a)=k(k^\top a)-|k|^2a$. The final equality is just factorization of $|k|^2$; $P_k$ is the orthogonal projector onto $k^\perp$ because it annihilates the component parallel to $k$ and leaves every vector orthogonal to $k$ unchanged. Thus the Fourier-space projector is exactly the transform of the physical-space operator $-\Delta I+\nabla\nabla^\top$, not a separate modeling assumption.

This identity also makes clear why two curls are not required merely to enforce incompressibility. The same work separately compares the one-curl Curl Kernel \citep{ni2025dfk} of the form
\begin{equation}
u_{\rm curl}
=
\nabla\times(\phi\omega),
\label{eq:ni_one_curl}
\end{equation}
which already satisfies
\begin{equation}
\boxed{
\nabla\cdot u_{\rm curl}
=
\nabla\cdot\bigl(\nabla\times(\phi\omega)\bigr)
=0.}
\label{eq:one_curl_divfree}
\end{equation}
Therefore the second curl in Eq.~\eqref{eq:ni_dfk_double_curl} is not an additional divergence-free requirement. It changes the basis produced from the same underlying scalar kernel, including its local geometry and spectral weighting.

Now define the free vector field represented by the kernels,
\begin{equation}
b_{\rm DFK}(x)
=
\sum_{i=1}^{M}\phi_i(x)\omega_i.
\label{eq:ni_free_b}
\end{equation}
Linearity of the differential operator gives
\begin{align}
u_{\rm DFK}
&=
\sum_{i=1}^{M}
\left(-I\Delta+\nabla\nabla^\top\right)
\bigl(\phi_i\omega_i\bigr)\\
&=
\left(-I\Delta+\nabla\nabla^\top\right)
\sum_{i=1}^{M}\phi_i\omega_i\\
&=
\boxed{
\left(-I\Delta+\nabla\nabla^\top\right)b_{\rm DFK}.}
\label{eq:ni_b_operator}
\end{align}
In three dimensions, Eqs.~\eqref{eq:double_curl_identity} and \eqref{eq:ni_b_operator} also give
\begin{equation}
u_{\rm DFK}=\nabla\times\nabla\times b_{\rm DFK}.
\label{eq:ni_free_b_double_curl}
\end{equation}
The many kernels therefore form a finite localized basis for an otherwise free $D$-component auxiliary vector field $b(x)$.

\subsection{Step-by-step equivalence with the NCL vector-$b$ construction}
Appendix~\ref{app:ncl_redundancy} derives NCL's vector construction from
\begin{equation}
A=J_b-J_b^\top,
\label{eq:ncl_vector_recalled_A}
\end{equation}
followed by the row-wise divergence
\begin{equation}
(v_{\rm NCL})_i
=
\sum_{j=1}^{D}\partial_j A_{ij}.
\label{eq:ncl_vector_recalled_div}
\end{equation}
Substituting $A_{ij}=\partial_jb_i-\partial_i b_j$ into Eq.~\eqref{eq:ncl_vector_recalled_div} gives
\begin{align}
(v_{\rm NCL})_i
&=
\sum_j\partial_j
\left(\partial_jb_i-\partial_i b_j\right)\\
&=
\sum_j\partial_j^2b_i
-
\partial_i\sum_j\partial_jb_j\\
&=
\Delta b_i-\partial_i(\nabla\cdot b).
\end{align}
Hence
\begin{equation}
\boxed{
v_{\rm NCL}
=
\left(\Delta I-\nabla\nabla^\top\right)b.}
\label{eq:ncl_vector_operator_again}
\end{equation}
Equation~\eqref{eq:ncl_vector_operator_again} is the physical-space NCL operator derived directly from Eqs.~\eqref{eq:ncl_vector_recalled_A}--\eqref{eq:ncl_vector_recalled_div}.

Compare Eq.~\eqref{eq:ncl_vector_operator_again} with the DFK result in Eq.~\eqref{eq:ni_b_operator}:
\begin{equation}
\left(-I\Delta+\nabla\nabla^\top\right)
=
-\left(\Delta I-\nabla\nabla^\top\right).
\label{eq:dfk_ncl_operator_sign}
\end{equation}
Therefore, if the same free vector field is used, $b=b_{\rm DFK}$,
\begin{equation}
\boxed{
u_{\rm DFK}=-v_{\rm NCL}.}
\label{eq:dfk_ncl_equivalence}
\end{equation}
The overall sign can be absorbed into the kernel coefficients $\omega_i$, so the two constructions generate exactly the same function class after the free vector field is identified.

The same equivalence is explicit in Fourier space. Using
\begin{equation}
\widehat{\Delta b}(k)=-|k|^2\widehat b(k),
\qquad
\widehat{\nabla(\nabla\cdot b)}(k)
=-kk^\top\widehat b(k),
\end{equation}
Eq.~\eqref{eq:ncl_vector_operator_again} becomes
\begin{align}
\widehat v_{\rm NCL}(k)
&=
\left(-|k|^2I+kk^\top\right)\widehat b(k)\\
&=
-\left(|k|^2I-kk^\top\right)\widehat b(k)\\
&=
\boxed{-|k|^2P_k\widehat b(k)},
\qquad
P_k=I-\frac{kk^\top}{|k|^2}.
\label{eq:ncl_symbol_again}
\end{align}
For the DFK operator,
\begin{align}
\widehat u_{\rm DFK}(k)
&=
\left(|k|^2I-kk^\top\right)\widehat b_{\rm DFK}(k)\\
&=
\boxed{|k|^2P_k\widehat b_{\rm DFK}(k)}.
\label{eq:dfk_symbol}
\end{align}
Equations~\eqref{eq:ncl_symbol_again} and \eqref{eq:dfk_symbol} are exact negatives of one another. They have the same range $k^\perp$ and the same null direction $\operatorname{span}\{k\}$. This is the precise sense in which the DFK construction is a special kernel-basis realization of the NCL vector-$b$ construction.

\subsection{Relation to the deterministic \emph{Project} branch}
The same mode-wise redundancy appears in the deterministic \emph{Project} branch \citep{li2026project}. That method first produces an unconstrained $D$-component Fourier mode $\widehat q(k)\in\mathbb C^D$ and then applies the Leray projector
\begin{equation}
\widehat u_{\rm P\&G}(k)
=
P_k\widehat q(k),
\qquad
P_k=I-\frac{kk^\top}{|k|^2}.
\label{eq:project_generate_explicit}
\end{equation}
Writing $\widehat q(k)=\widehat q_\perp(k)+\alpha(k)k$ gives
\begin{equation}
P_k\widehat q(k)=\widehat q_\perp(k),
\qquad
P_k\bigl(\alpha(k)k\bigr)=0.
\end{equation}
Thus the deterministic \emph{Project} branch \citep{li2026project}, the NCL vector-$b$ construction \citep{richterpowell2022ncl}, and the DFK construction \citep{ni2025dfk} share the same mode-wise geometric pattern: they introduce a $D$-component ambient representation containing one longitudinal degree of freedom that is invisible to the final divergence-free field. Their operators differ in spectral scaling: the deterministic \emph{Project} branch uses $P_k$, NCL uses $-|k|^2P_k$, and DFK uses $|k|^2P_k$. In all three cases the admissible range is $k^\perp$ and the longitudinal direction $\operatorname{span}\{k\}$ is removed. Our representation differs by never introducing this redundant longitudinal degree of freedom: it starts directly from $D-1$ coordinates in $k^\perp$.

\subsection{What differs: the representation of the free vector field}
NCL uses a coordinate neural network to parameterize the free $D$-component field,
\begin{equation}
b(x)=b_\theta(x)\in\mathbb R^D.
\end{equation}
The DFK construction \citep{ni2025dfk} instead uses the finite kernel expansion in Eq.~\eqref{eq:ni_free_b},
\begin{equation}
b_{\rm DFK}(x)
=
\sum_{i=1}^{M}\phi_i(x)\omega_i,
\end{equation}
with trainable kernel locations, scales, and vector weights. Thus both constructions first specify a free $D$-component vector field and then apply the same second-order transverse differential operator up to sign; the difference is the function class used to represent that free field.

\subsection{Kernel counts and trainable degrees of freedom}
The finite-basis nature of the DFK representation is visible in the reported parameter counts \citep{ni2025dfk}. In the two-dimensional K\'arm\'an vortex-street experiment, DFK-Wen4 uses $5{,}367$ kernels and $26{,}835$ trainable parameters. This corresponds to five trainable quantities per kernel: two center coordinates, one scale, and a two-component vector weight. Their Curl Kernel comparison uses $6{,}794$ kernels and $27{,}176$ trainable parameters.

In the three-dimensional analytic-vortex experiment, the kernel methods use $21{,}117$ kernels and $147{,}819$ trainable parameters. The count is seven parameters per kernel,
\begin{equation}
(x_i,y_i,z_i,h_i,\omega_{ix},\omega_{iy},\omega_{iz}),
\end{equation}
namely three center coordinates, one scale, and a three-component vector weight. These numbers make the interpretation concrete: DFK represents the auxiliary vector field $b$ with a large collection of localized basis functions and then applies the transverse operator.

\subsection{Comparison with the minimal transverse representation}
The NCL vector form and DFK retain a free $D$-component auxiliary vector field before the transverse operator is applied. Mode by mode, the longitudinal part of that auxiliary field is invisible because
\begin{equation}
P_k\left(\widehat b_\perp+\alpha k\right)=\widehat b_\perp.
\end{equation}
The NCL matrix form instead retains an antisymmetric matrix field $A(x)\in\mathfrak{so}(D)$ and applies its row-wise divergence. Mode by mode, this is the surjective linear decoder $\widehat A(k)\mapsto i\widehat A(k)k\in k^\perp$, rather than a vector projection. Our representation instead stores a real $(D-1)$-component reduced field $a(x,t)$. At each time $t$, a spatial Fourier transform produces its mode coefficients $\widehat a(k,t)$, these $D-1$ coefficients are mapped by the transverse basis $E(k)$ to the physical Fourier coefficient $\widehat u(k,t)$, and an inverse spatial Fourier transform returns the real $D$-component field. The comparison is therefore
\begin{equation}
\boxed{
\begin{array}{rcl}
\text{NCL vector form} &:&
\begin{gathered}[t]
b_\theta(x)\in\mathbb R^D
\xrightarrow{\ \Delta I-\nabla\nabla^\top\ } u_{\rm NCL\text{-}vec}(x)\in\mathbb R^D\\[-1mm]
\text{equivalently:}\quad
\widehat b_\theta(k)\in\mathbb C^D
\xrightarrow{\ -|k|^2P_k\ } \widehat u_{\rm NCL\text{-}vec}(k)\in k^\perp
\end{gathered},\\[2mm]
\text{DFK} &:&
\begin{gathered}[t]
b_{\rm DFK}(x)=\sum_i\phi_i(x)\omega_i\in\mathbb R^D
\xrightarrow{\ -\Delta I+\nabla\nabla^\top\ } u_{\rm DFK}(x)\in\mathbb R^D\\[-1mm]
\text{equivalently:}\quad
\widehat b_{\rm DFK}(k)\in\mathbb C^D
\xrightarrow{\ |k|^2P_k\ } \widehat u_{\rm DFK}(k)\in k^\perp
\end{gathered},\\[2mm]
\text{NCL matrix form} &:&
\begin{gathered}[t]
A_\theta(x)\in\mathfrak{so}(D)
\xrightarrow{\ \operatorname{div}_{\rm row}\ } u_{\rm NCL\text{-}mat}(x)\in\mathbb R^D\\[-1mm]
\text{equivalently:}\quad
\widehat A_\theta(k)\in\mathfrak{so}(D,\mathbb C)
\xrightarrow{\ i\widehat A_\theta(k)k\ } \widehat u_{\rm NCL\text{-}mat}(k)\in k^\perp
\end{gathered},\\[2mm]
\text{ours} &:&
\begin{aligned}[t]
a(x,t)\in\mathbb R^{D-1}
&\xrightarrow{\ \mathcal F_x\ } \widehat a(k,t)\in\mathbb C^{D-1}\\
&\xrightarrow{\ E(k)\ } \widehat u(k,t)\in k^\perp
\xrightarrow{\ \mathcal F_x^{-1}\ } u(x,t)\in\mathbb R^D.
\end{aligned}
\end{array}}
\end{equation}
Here $\mathfrak{so}(D)=\{A\in\mathbb R^{D\times D}:A^\top=-A\}$ and $\mathfrak{so}(D,\mathbb C)$ is its complex Fourier-space counterpart; $P_k=I-kk^\top/|k|^2$. Also, $\mathcal F_x$ denotes the spatial Fourier transform only, so time is not Fourier transformed. Thus $a(x,t)$ is the real reduced field seen by the learning model, while $\widehat a(k,t)$ and $\widehat u(k,t)$ are internal Fourier-space representations used by the fixed analytic transform. The paired $k/-k$ convention preserves Hermitian symmetry, so the final inverse transform is real. The NCL vector and DFK rows use the same transverse second-order construction up to an overall sign and differ mainly in how the free $D$-component vector field is represented. The NCL matrix row is instead a redundant antisymmetric-matrix decoder with a large kernel. The final row removes these auxiliary null directions and learns directly with only $D-1$ real channels.

\clearpage
\section{Navier--Stokes and Magnetohydrodynamics in Minimal Transverse Coordinates}
\label{app:ns_mhd_reduced_dynamics}

This appendix makes explicit how the divergence-free state representation developed in this paper changes the governing equations themselves.  The main point is structural: an incompressible evolution is usually written as a vector evolution equation together with a separate divergence constraint.  If the state is parameterized directly in the transverse Fourier subspace, the divergence constraint is satisfied identically and therefore no longer appears as an independent equation.  In addition, pressure gradients lie entirely in the longitudinal Fourier direction and vanish after transformation to the reduced coordinates.  We first show this for the Navier--Stokes equations and then for incompressible magnetohydrodynamics (MHD).

\subsection{Incompressible Navier--Stokes in $D$ dimensions}
For a constant-density incompressible fluid, the velocity field $u(x,t)\in\mathbb R^D$ satisfies
\begin{align}
\partial_t u+(u\!\cdot\!\nabla)u
&=-\nabla p+\nu\Delta u+f,
\label{eq:ns_D_momentum}\\
\nabla\!\cdot u&=0,
\label{eq:ns_D_incompressibility}
\end{align}
where the pressure has been divided by the constant density, $\nu$ is the kinematic viscosity, and $f$ is a body force.  Thus the standard incompressible formulation consists of a $D$-component evolution equation coupled to the scalar constraint in Eq.~\eqref{eq:ns_D_incompressibility}.  The pressure acts as the Lagrange multiplier that keeps the velocity evolution inside the divergence-free subspace \citep{chorin1968numerical}.

\subsubsection{Two dimensions: scalar vorticity and stream function}
In two spatial dimensions, introduce the constant $90^\circ$ rotation matrix
\begin{equation}
R=
\begin{bmatrix}
0&-1\\
1&0
\end{bmatrix},
\qquad
R^\top R=I,
\qquad
R^\top=-R,
\end{equation}
and define a scalar stream function $\psi$ by
\begin{equation}
u=R\nabla\psi\equiv\nabla^\perp\psi.
\label{eq:2d_stream_function}
\end{equation}
Because the Hessian of $\psi$ is symmetric while $R$ is antisymmetric,
\begin{equation}
\nabla\!\cdot u
=\operatorname{tr}\!\left(R\nabla^2\psi\right)=0
\end{equation}
identically.  Hence incompressibility is built into the stream-function representation rather than imposed as a separate equation.

Let the scalar vorticity be $\omega=\operatorname{curl}_{2D}u$.  With the convention in Eq.~\eqref{eq:2d_stream_function},
\begin{equation}
\omega=\Delta\psi.
\end{equation}
Taking the curl of Eq.~\eqref{eq:ns_D_momentum} eliminates the pressure gradient and gives
\begin{equation}
\partial_t\omega+(u\!\cdot\!\nabla)\omega
=\nu\Delta\omega+\operatorname{curl}_{2D}f.
\label{eq:2d_vorticity_ns}
\end{equation}
Substituting $u=R\nabla\psi$ and $\omega=\Delta\psi$ produces a single scalar evolution equation,
\begin{equation}
\boxed{
\partial_t\Delta\psi
+\bigl(R\nabla\psi\!\cdot\!\nabla\bigr)\Delta\psi
=\nu\Delta^2\psi+\operatorname{curl}_{2D}f.}
\label{eq:2d_stream_vorticity_single}
\end{equation}
The pair ``momentum equation $+$ incompressibility constraint'' has therefore been replaced by one scalar stream-function/vorticity equation, with incompressibility satisfied automatically.

For the inviscid unforced limit $\nu=0$, Eq.~\eqref{eq:2d_vorticity_ns} reduces to material transport of scalar vorticity,
\begin{equation}
\partial_t\omega+(u\!\cdot\!\nabla)\omega=0.
\end{equation}
Under periodic boundaries, sufficient decay at infinity, or boundary conditions that remove the corresponding flux terms, this implies conservation of both kinetic energy and enstrophy,
\begin{equation}
\mathcal E
=\frac12\int |u|^2\,dx,
\qquad
\mathcal Z
=\frac12\int \omega^2\,dx.
\label{eq:2d_energy_enstrophy}
\end{equation}
Here the second invariant is \emph{enstrophy}, not entropy.  For viscous Navier--Stokes, these quantities are dissipative rather than exactly conserved; in particular, in the unforced periodic setting,
\begin{equation}
\frac{d\mathcal E}{dt}
=-\nu\int |\nabla u|^2\,dx,
\qquad
\frac{d\mathcal Z}{dt}
=-\nu\int |\nabla\omega|^2\,dx.
\end{equation}

\subsubsection{Three dimensions: vortex stretching}
In three dimensions, the vorticity is the vector
\begin{equation}
\omega=\nabla\times u.
\end{equation}
Taking the curl of the incompressible momentum equation yields
\begin{equation}
\boxed{
\partial_t\omega+(u\!\cdot\!\nabla)\omega
=(\omega\!\cdot\!\nabla)u
+\nu\Delta\omega+\nabla\times f.}
\label{eq:3d_vorticity_ns}
\end{equation}
The additional nonlinear term
\begin{equation}
(\omega\!\cdot\!\nabla)u
\end{equation}
is the vortex-stretching term.  It is absent from the two-dimensional scalar-vorticity equation.  Consequently, even in the inviscid unforced limit, the three-dimensional enstrophy is not generally conserved:
\begin{equation}
\frac{d}{dt}\frac12\int |\omega|^2\,dx
=
\int \omega\cdot(\omega\!\cdot\!\nabla)u\,dx,
\label{eq:3d_enstrophy_stretching}
\end{equation}
whereas kinetic energy remains conserved for smooth inviscid solutions under the same no-flux assumptions.  Three-dimensional Euler also conserves helicity $\int u\cdot\omega\,dx$ under the standard corresponding assumptions.  With viscosity, kinetic energy is dissipated and Eq.~\eqref{eq:3d_enstrophy_stretching} acquires the additional negative viscous contribution $-\nu\int|\nabla\omega|^2dx$.

The dimensional distinction can therefore be summarized without expanding component equations:
\begin{equation}
\boxed{
\begin{array}{c|c|c}
&\text{2D incompressible flow}&\text{3D incompressible flow}\\
\hline
\text{vorticity}&\text{scalar}&\text{vector}\\
\text{vortex stretching}&\text{absent}&(\omega\!\cdot\!\nabla)u\\
\text{inviscid energy}&\text{conserved}&\text{conserved}\\
\text{inviscid enstrophy}&\text{conserved}&\text{not generally conserved}
\end{array}}
\label{eq:2d_3d_ns_summary}
\end{equation}

\subsection{The $D\to D-1$ transverse formulation of Navier--Stokes}
The stream-function formulation above is a special two-dimensional way to build incompressibility into the state variable.  The transverse representation in this paper provides an analogous construction in arbitrary spatial dimension $D$.

For each nonzero Fourier mode $k$, let
\begin{equation}
E(k)\in\mathbb R^{D\times(D-1)},
\qquad
E(k)^*E(k)=I_{D-1},
\qquad
E(k)E(k)^*=P_k
=I-\frac{kk^\top}{|k|^2},
\end{equation}
so the columns of $E(k)$ form an orthonormal basis of $k^\perp$.  Using the notation of the main text, define
\begin{equation}
A_0=\mathcal F_D^{-1}E\mathcal F_{D-1},
\qquad
A_0^*=\mathcal F_{D-1}^{-1}E^*\mathcal F_D.
\label{eq:A0_appendix_f}
\end{equation}
Write the physical velocity as
\begin{equation}
u=A_0a,
\qquad
a=A_0^*u,
\qquad
a(x,t)\in\mathbb R^{D-1}.
\end{equation}
For every nonzero mode,
\begin{equation}
k^\top\widehat u(k)
=k^\top E(k)\widehat a(k)=0,
\end{equation}
so $\nabla\!\cdot u=0$ is an identity of the representation.

Now apply $A_0^*$ to Eq.~\eqref{eq:ns_D_momentum}.  Because $E(k)^*k=0$,
\begin{equation}
A_0^*\nabla p=0.
\label{eq:pressure_annihilation}
\end{equation}
Because $A_0$ and the Laplacian are both Fourier multipliers and $E(k)^*E(k)=I_{D-1}$,
\begin{equation}
A_0^*\Delta A_0a=\Delta a.
\end{equation}
Therefore the incompressible Navier--Stokes system reduces to the single $(D-1)$-component evolution equation
\begin{equation}
\boxed{
\partial_t a
+A_0^*\!\left[((A_0a)\!\cdot\!\nabla)(A_0a)\right]
=\nu\Delta a+A_0^*f.}
\label{eq:reduced_ns_D_minus_1}
\end{equation}
No separate equation $\nabla\!\cdot u=0$ is required, and the pressure variable also disappears from the reduced evolution because its gradient is purely longitudinal.  The physical velocity is recovered analytically as $u=A_0a$.  Thus, modulo the separately handled zero Fourier mode described in the main text, the usual pair
\begin{equation}
\left\{
\begin{array}{l}
\text{$D$-component momentum equation},\\
\nabla\!\cdot u=0
\end{array}
\right.
\end{equation}
is represented directly by one evolution equation for a $(D-1)$-component state.

\subsection{From Maxwell plus fluid dynamics to incompressible MHD}
We next make the same structure explicit for MHD.  The physical derivation is naturally three-dimensional because the cross product is three-dimensional; after eliminating the cross products, the final advective/tension form extends directly to the $D$-dimensional transverse notation.

\subsubsection{Electromagnetic part}
In the nonrelativistic, low-frequency MHD approximation, the Maxwell equations needed for magnetic evolution are
\begin{align}
\nabla\!\cdot B&=0,
\label{eq:maxwell_divB}\\
\partial_t B&=-\nabla\times E,
\label{eq:maxwell_faraday}\\
\nabla\times B&=\mu_0J,
\label{eq:maxwell_ampere_mhd}
\end{align}
where the displacement-current term is neglected in Amp\`ere's law.  A resistive Ohm law for a moving conducting fluid is
\begin{equation}
E+u\times B=\sigma^{-1}J,
\label{eq:mhd_ohm}
\end{equation}
with conductivity $\sigma$.  The ideal-MHD limit is $\sigma\to\infty$, giving $E+u\times B=0$.

Substituting Eq.~\eqref{eq:mhd_ohm} into Faraday's law and then using Eq.~\eqref{eq:maxwell_ampere_mhd} gives
\begin{align}
\partial_tB
&=\nabla\times(u\times B)-\sigma^{-1}\nabla\times J\\
&=\nabla\times(u\times B)+\eta_m\Delta B,
\label{eq:mhd_induction_curl}
\end{align}
where
\begin{equation}
\eta_m=\frac{1}{\mu_0\sigma}
\end{equation}
is the magnetic diffusivity and $\nabla\!\cdot B=0$ has been used.  Taking the divergence of Eq.~\eqref{eq:mhd_induction_curl} shows that an initially divergence-free magnetic field remains divergence-free; nevertheless, numerical schemes often require explicit control of this constraint \citep{toth2000constraint}.

\subsubsection{Fluid part and Lorentz force}
Mass conservation is
\begin{equation}
\partial_t\rho+\nabla\!\cdot(\rho u)=0.
\end{equation}
For constant density, this becomes
\begin{equation}
\nabla\!\cdot u=0.
\label{eq:mhd_divu_origin}
\end{equation}
The fluid momentum equation with the electromagnetic force is
\begin{equation}
\rho\left[\partial_tu+(u\!\cdot\!\nabla)u\right]
=-\nabla p+\rho\nu\Delta u+J\times B+\rho f,
\label{eq:mhd_momentum_lorentz}
\end{equation}
where the electric force density is neglected in the usual quasi-neutral nonrelativistic MHD limit.  Amp\`ere's law and $\nabla\!\cdot B=0$ give the vector identity
\begin{equation}
J\times B
=\frac{1}{\mu_0}(\nabla\times B)\times B
=\frac{1}{\mu_0}\left[(B\!\cdot\!\nabla)B
-\nabla\frac{|B|^2}{2}\right].
\label{eq:mhd_lorentz_tension_pressure}
\end{equation}
Thus the Lorentz force separates into magnetic tension and a magnetic-pressure gradient.

Define the Alfv\'en-scaled magnetic field and total pressure by
\begin{equation}
b=\frac{B}{\sqrt{\mu_0\rho}},
\qquad
\Pi=\frac{p}{\rho}+\frac{|b|^2}{2},
\end{equation}
for constant $\rho$.  Using
\begin{equation}
\nabla\times(u\times b)
=(b\!\cdot\!\nabla)u-(u\!\cdot\!\nabla)b
+u(\nabla\!\cdot b)-b(\nabla\!\cdot u),
\end{equation}
and imposing both divergence constraints, incompressible resistive MHD becomes
\begin{align}
\partial_tu+(u\!\cdot\!\nabla)u
&=-\nabla\Pi+(b\!\cdot\!\nabla)b+\nu\Delta u+f,
\label{eq:mhd_four_u}\\
\partial_tb+(u\!\cdot\!\nabla)b
&=(b\!\cdot\!\nabla)u+\eta_m\Delta b,
\label{eq:mhd_four_b}\\
\nabla\!\cdot u&=0,
\label{eq:mhd_four_divu}\\
\nabla\!\cdot b&=0.
\label{eq:mhd_four_divb}
\end{align}
This is the four-equation structure relevant to the present representation: two vector evolution equations plus two divergence constraints.  The two constraints have different physical origins: $\nabla\!\cdot u=0$ follows from mass conservation at constant density, whereas $\nabla\!\cdot B=0$ is Gauss's law for magnetism.

\subsection{Reducing incompressible MHD from four equations to two}
Because both $u$ and $b$ are divergence-free vector fields on the same domain, they can be encoded independently by the same transverse frame.  Let
\begin{equation}
u=A_0a,
\qquad
b=A_0c,
\qquad
a,c\in\mathbb R^{D-1}.
\label{eq:mhd_reduced_fields}
\end{equation}
Then
\begin{equation}
\nabla\!\cdot(A_0a)=0,
\qquad
\nabla\!\cdot(A_0c)=0
\end{equation}
identically for all nonzero modes.  Applying $A_0^*$ to Eqs.~\eqref{eq:mhd_four_u}--\eqref{eq:mhd_four_b}, and again using $A_0^*\nabla\Pi=0$, gives
\begin{align}
\boxed{
\partial_ta
+A_0^*\!\left[(u\!\cdot\!\nabla)u-(b\!\cdot\!\nabla)b\right]
=\nu\Delta a+A_0^*f,}
\label{eq:mhd_reduced_a}\\
\boxed{
\partial_tc
+A_0^*\!\left[(u\!\cdot\!\nabla)b-(b\!\cdot\!\nabla)u\right]
=\eta_m\Delta c,}
\label{eq:mhd_reduced_c}
\end{align}
with the analytic reconstructions
\begin{equation}
u=A_0a,
\qquad
b=A_0c.
\end{equation}
Therefore the standard four-equation incompressible MHD structure
\begin{equation}
\left\{
\begin{array}{l}
\text{velocity evolution},\\
\text{magnetic-field evolution},\\
\nabla\!\cdot u=0,\\
\nabla\!\cdot b=0
\end{array}
\right.
\end{equation}
is represented by only two reduced vector evolution equations, one for $a$ and one for $c$.  Both divergence constraints are satisfied by construction, and the total-pressure gradient is removed automatically by the transverse coordinate map.  In an operator-learning implementation, a neural model may therefore evolve $(a,c)$ directly in $2(D-1)$ real channels and decode them analytically to the physical $(u,b)$ fields, rather than predicting $2D$ ambient channels and subsequently enforcing two separate solenoidal constraints.

\clearpage
\section{Direct Random Generation of Divergence-Free Fields in Transverse Coordinates}
\label{app:random_transverse_generation}

The same $D\to D-1$ representation can be used to generate random divergence-free fields directly, without first generating an unconstrained $D$-component field and then projecting it.  For every nonzero Fourier mode $k$, let
\begin{equation}
E(k)\in\mathbb R^{D\times(D-1)},
\qquad
E(k)^*E(k)=I_{D-1},
\qquad
k^\top E(k)=0,
\end{equation}
where $E(k)$ is the transverse orthonormal frame obtained from the Householder construction used in the main method.

Choose one representative from each Fourier pair $\{k,-k\}$.  In the intrinsic $(D-1)$-dimensional coordinates, sample independent real and imaginary Gaussian vectors,
\begin{equation}
\xi_R(k),\xi_I(k)\sim\mathcal N(0,I_{D-1}),
\qquad
\widehat a(k)=S(|k|)\bigl(\xi_R(k)+i\xi_I(k)\bigr),
\end{equation}
where $S(|k|)$ controls the spectral decay.  For example,
\begin{equation}
S(|k|)=\left(1+\frac{|k|^2}{k_0^2}\right)^{-\alpha/2}
\qquad\text{or}\qquad
S(|k|)=\exp\!\left[-\left(\frac{|k|}{k_c}\right)^q\right].
\end{equation}
Larger $\alpha$ or $q$ suppresses high frequencies more strongly, while smaller values retain more high-frequency energy.

Decode each sampled reduced coefficient directly into the physical $D$-component Fourier coefficient,
\begin{equation}
\widehat u(k)=E(k)\widehat a(k)\in\mathbb C^D.
\end{equation}
No projection is required, because
\begin{equation}
k^\top\widehat u(k)=k^\top E(k)\widehat a(k)=0.
\end{equation}
To obtain a real physical-space field, impose Hermitian symmetry on the paired mode,
\begin{equation}
\widehat u(-k)=\overline{\widehat u(k)},
\end{equation}
with the zero mode chosen separately and self-conjugate Nyquist modes taken real when present.  The final field is then
\begin{equation}
\boxed{
 u(x)=\mathcal F^{-1}\widehat u,
 \qquad
 \nabla\!\cdot u=0
}
\end{equation}
up to numerical roundoff.  Thus the complete generation pipeline is
\begin{equation}
\boxed{
\begin{gathered}
\mathcal N(0,I_{D-1})
\xrightarrow{\;S(|k|)\;}
\widehat a(k)\in\mathbb C^{D-1}
\xrightarrow{\;E(k)\;}
\widehat u(k)\in k^\perp\subset\mathbb C^D \\
\widehat u(k)
\xrightarrow{\;\text{Hermitian pairing}\;}
\widehat u(-k)=\overline{\widehat u(k)} \\
\widehat u(-k)
\xrightarrow{\;\mathcal F^{-1}\;}
u(x)\in\mathbb R^D,
\quad \nabla\!\cdot u=0.
\end{gathered}
}
\end{equation}
The random field is therefore generated directly in the intrinsic transverse subspace, with exactly $D-1$ random degrees of freedom per nonzero Fourier mode and no post-hoc Leray projection.

\clearpage
\section{Spectral Meaning of the 2D Navier--Stokes Prediction Losses}
\label{app:ns2d_loss_comparison}

This appendix summarizes why the five 2D Navier--Stokes prediction variables used in our comparisons induce different spectral weightings.  The equations below describe the squared-$L^2$ error numerator; common averaging or relative-$L^2$ normalization does not change the corresponding mode weighting within each representation.

\paragraph{Velocity--vorticity relation.}
For every nonzero Fourier mode of a two-dimensional incompressible velocity field,
\begin{equation}
\widehat u(k)=\frac{i k^\perp}{|k|^2}\widehat\omega(k),
\qquad
|\widehat\omega(k)|^2=|k|^2|\widehat u(k)|^2.
\label{eq:ns2d_u_omega_loss_relation}
\end{equation}
Let $e_u(k)=\widehat u_{\rm pred}(k)-\widehat u_{\rm true}(k)$ and $e_\omega(k)=\widehat\omega_{\rm pred}(k)-\widehat\omega_{\rm true}(k)$.  For divergence-free errors, Eq.~\eqref{eq:ns2d_u_omega_loss_relation} gives
\begin{equation}
|e_\omega(k)|^2=|k|^2|e_u(k)|^2.
\end{equation}

\paragraph{Five prediction variables.}
The corresponding losses can therefore be read as follows:
\begin{equation}
\boxed{
\begin{array}{c|c|c}
\text{Model} & \text{Predicted variable} & \text{Equivalent spectral weighting}\\
\hline
\text{Original FNO-}\omega & \omega
& \displaystyle L_\omega=\sum_k |e_\omega|^2
 =\sum_k |k|^2|e_u|^2\\[2mm]
A\ (\text{raw}) & u
& \displaystyle L_A=\sum_k |e_u|^2\\[2mm]
B\ (\text{Leray}) & P_k u_{\rm raw}
& \displaystyle L_B=\sum_k |P_k e_{\rm raw}|^2\\[2mm]
C\ (\text{potential}) & \psi
& \displaystyle L_C=\sum_k |e_\psi|^2
 =\sum_{k\neq0}\frac{|e_u|^2}{|k|^2}\\[3mm]
D2\ (\text{transverse}) & a
& \displaystyle L_{D2}=\sum_k |e_a|^2
 =\sum_k |e_u|^2
\end{array}}
\label{eq:ns2d_five_loss_table}
\end{equation}
Here $P_k=I-kk^\top/|k|^2$.  Model B predicts an ambient velocity and removes its longitudinal component by projection, so only the transverse part of the raw error reaches the projected output loss.  For model C, $u=\nabla^\perp\psi$ implies $\widehat u=i k^\perp\widehat\psi$, hence $|e_\psi|^2=|e_u|^2/|k|^2$.  For D2, the orthonormal transverse coordinate is isometric to velocity, so $|e_a|^2=|e_u|^2$.

Thus the main spectral distinction is immediate: direct vorticity prediction weights velocity error by $|k|^2$ and therefore emphasizes high wavenumbers; the potential representation weights it by $1/|k|^2$ and therefore emphasizes low wavenumbers; raw velocity and D2 use the ordinary velocity $L^2$ weighting.  D2 differs from raw velocity in representation, not in the ideal $L^2$ norm: it stores only the single admissible transverse degree of freedom in 2D.

\paragraph{Radial-versus-phase ablation.}
In two dimensions the D2 coefficient can be written, up to the chosen transverse-basis orientation, as
\begin{equation}
\widehat a(k)=\frac{\gamma(k)}{|k|}\widehat\omega(k),
\qquad |\gamma(k)|=1,
\end{equation}
where $\gamma(k)$ is the mode-dependent phase/sign factor generated by the basis convention.  The four ablations isolate the two effects:
\begin{equation}
\boxed{
F0:\ \widehat\omega,
\qquad
F1:\ \frac{\widehat\omega}{|k|},
\qquad
F2:\ \gamma(k)\widehat\omega,
\qquad
F3:\ \frac{\gamma(k)}{|k|}\widehat\omega=\widehat a.
}
\end{equation}
Because $|\gamma(k)|=1$,
\begin{equation}
\boxed{
L_{F0}=L_{F2}=\sum_k |e_\omega(k)|^2,
\qquad
L_{F1}=L_{F3}=\sum_{k\neq0}\frac{|e_\omega(k)|^2}{|k|^2}=L_u.
}
\end{equation}
Hence $1/|k|$ changes the spectral weighting, whereas the phase/sign factor changes the representation without changing its $L^2$ mode weights.

\clearpage
\section{Geometric Comparison of Three Evolution Representations}
\label{app:three_evolution_representations}
\begin{figure}[H]
\centering
\includegraphics[width=\linewidth,height=0.80\textheight,keepaspectratio]{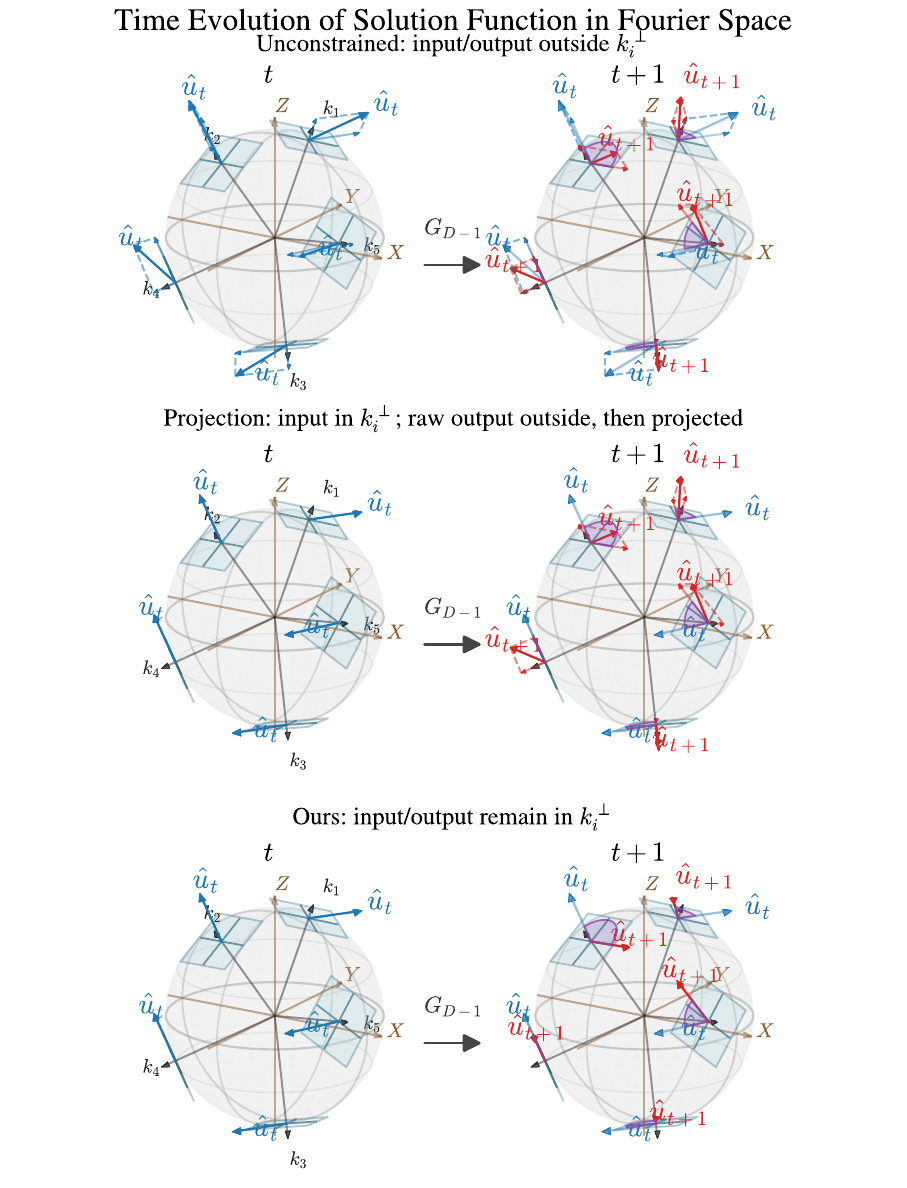}
\caption{Geometric comparison of three temporal-evolution representations for each nonzero wavenumber $k_i$. \textbf{Top (unconstrained):} neither the input nor the output is required to be divergence-free or to lie in $k_i^\perp$, so the learned evolution uses ambient $D$-component modes. \textbf{Middle (Project-and-Generate):} the model uses an ambient $D$-component representation with no constraint on the longitudinal direction; its raw output is likewise unconstrained and is returned to $k_i^\perp$ only by a subsequent projection. \textbf{Bottom (ours):} the input and output are represented directly by the $D-1$ transverse coordinates in $k_i^\perp$. The neural operator never predicts the longitudinal component that projection would discard, and no projection is needed after prediction.}
\label{fig:three_evolution_representations}
\end{figure}

\clearpage
\section{Representative Physical-Space Reconstruction Gallery}
\label{app:reconstruction_gallery}
Each figure compares the original field, its real-space $D-1$ latent channel(s), the reconstructed field, and pointwise error.  The 3D figures additionally show $xy$, $xz$, and $yz$ slices at 25\%, 50\%, and 75\%.

\subsection{Why a reduced latent can extend beyond a closed physical boundary}
\label{app:exterior_latent}
The white exterior of a closed-domain figure is not physical velocity data.  To apply an FFT, the physical field is embedded in a periodic computational box $\widetilde\Omega$; write this array as $u_{\rm emb}(x,t)=\chi_\Omega(x)u(x,t)$, where $\chi_\Omega$ is the physical-domain mask.  Here ``zero outside'' is an embedding convention, not a claim that the physical velocity is defined outside the wall.  The physical $D$-component input, reconstruction, and error are consequently masked outside $\Omega$ in the plots.

At each time $t$, the direct encoder on the nonzero modes is $a(\cdot,t)=\mathcal F_{D-1}^{-1}E^*\mathcal F_Du_{\rm emb}(\cdot,t)$.  Let $B_{\ell j}(k)=[E(k)^*]_{\ell j}=E_{j\ell}(k)$ and define its continuous inverse-Fourier kernel by
\begin{equation}
 K_{\ell j}(x)=\mathcal F_x^{-1}\!\left[B_{\ell j}(k)\right](x),
 \qquad
 \mathcal K_{\ell j}(x,\tau)=K_{\ell j}(x)\,\delta(\tau).
 \label{eq:exterior_kernel}
\end{equation}
Consequently the complete transform is a matrix-valued $D\!\to\!D-1$ convolution over space--time,
\begin{equation}
 a_\ell(x,t)
 =\sum_{j=1}^{D}\int_{\mathbb R}\!\int_{\widetilde\Omega}
 \mathcal K_{\ell j}(x-y,t-s)\,u_{{\rm emb},j}(y,s)\,dy\,ds
 =\sum_{j=1}^{D}\int_{\widetilde\Omega}
 K_{\ell j}(x-y)\chi_\Omega(y)u_j(y,t)\,dy.
 \label{eq:exterior_convolution}
\end{equation}
The $\delta(t-s)$ factor records that the fixed Fourier/Householder transform does not mix different times; its nonlocality is spatial.  The zero Fourier mode remains the separately stored $D$-component mean described above.  Thus, for $x\notin\Omega$, the interior values $y\in\Omega$ in Eq.~\eqref{eq:exterior_convolution} can yield $a_\ell(x,t)\ne0$.  These exterior values are valid reduced coordinates required by the exact transform; they are not a physical velocity outside the wall, and masking or deleting them would generally destroy exact decoding.

The degree of globality---the spatial extent and tail length of $K$---is controlled by the regularity of the Fourier multiplier $B(k)$.  A smooth multiplier has rapidly decaying inverse-Fourier coefficients.  Conversely, a sharper transition, larger discontinuity, or gauge seam in $E(k)$ produces more slowly decaying, broader spatial tails in $K$ and therefore a more global convolution.  The canonical Householder choice and the paired $k/-k$ convention can introduce such a directional seam (Appendix~\ref{app:operator_structure}); this explains why the latent exterior can be visibly global even when the physical field is restricted to a closed geometry.  On a finite FFT box the kernel is periodized, so ``local'' means concentration and decay within that box, not compact support.  This is distinct from the open-flow cards: there the virtual region is an intentional Fourier continuation of the physical inlet/outlet domain.

\paragraph{Why the boundary-exterior reduced latent is often more pronounced in three dimensions.}
For a two-dimensional physical field, the wavevector direction lies on $S^1$, whose tangent bundle admits a globally continuous unit tangent vector: for $n=(n_1,n_2)$ one may take $e(n)=(-n_2,n_1)^\top$.  Thus the $2\to1$ transverse multiplier can be chosen globally smooth on the direction circle, and its inverse-Fourier kernel can be comparatively concentrated.  In contrast, for a three-dimensional physical field the direction lies on $S^2$ and the $3\to2$ construction requires two independent tangent directions at every point.  No single continuous global tangent frame exists on $S^2$ (the hairy-ball obstruction).  Consequently, every one-chart $3\to2$ frame must have a directional discontinuity or singular chart transition.  In the Householder construction this unavoidable defect appears near the reference-axis pole; changing the gauge can move the pole or seam, but cannot remove it.

This forced loss of angular regularity makes the three-dimensional multiplier $E(k)$ less smooth in Fourier space than a globally smooth two-dimensional choice.  Its inverse-Fourier matrix kernel therefore has slower-decaying, more global spatial tails.  Applied through Eq.~\eqref{eq:exterior_convolution}, those longer tails carry interior field values farther into $x\notin\Omega$, so the $D-1$ latent outside a closed physical boundary is commonly much more visibly nonzero in three dimensions than in two.  The exact magnitude still depends on the geometry and mask, the field spectrum near the seam, the FFT box, and the gauge.  Moreover, the paired $k/-k$ convention used to make the latent real can add a coordinate seam even in two dimensions; the key distinction is that only the three-dimensional discontinuity is topologically unavoidable.

\raggedbottom
\subsection{Open-boundary Fourier extension: accuracy and stored-coefficient trade-off}
\label{app:open_extension_tradeoff}
For a nonperiodic channel with inflow and outflow profiles that differ locally, a direct zero-padded FFT is not an admissible pure transverse representation: the boundary flux creates a longitudinal component.  We instead construct a periodic, spectrally divergence-free extension $U$ that restricts to the physical field, $RU\approx u$.  The extension length is $g=\widetilde L/L$, and the axial Fourier density is $r=N_x^{\rm Fourier}/N_x$.  At $g=1.2$ and $r=2$, the extension introduces only a $20\%$ virtual interval while doubling the axial Fourier coefficient density.  The resulting full codec payload stores exactly $D-1$ transverse coefficients at every nonzero mode (and the separately handled mean), yet reconstructs the physical field after decoding and cropping with relative error $8.45\times10^{-15}$.

Table~\ref{tab:open_extension_storage} records the subsequent coefficient-sparsity experiment on this stable open-channel representation.  The smallest Fourier coefficients were discarded by magnitude and the retained $D-1$ payload was decoded without an auxiliary longitudinal residual channel.  The storage measure is the real-scalar-equivalent coefficient count divided by the original physical $D$-component array count; it does not charge index or quantization metadata.  Thus the table directly separates the exact intrinsic component reduction from the practical cost of the extra Fourier modes required by an open-boundary extension.

\begin{table}[H]
\centering
\small
\setlength{\tabcolsep}{5pt}
\renewcommand{\arraystretch}{1.15}
\begin{tabularx}{\linewidth}{@{}>{\centering\arraybackslash}p{0.25\linewidth}>{\centering\arraybackslash}p{0.25\linewidth}>{\raggedright\arraybackslash}X@{}}
\toprule
Retained $D-1$ payload / original data & $\|R U_{\rm decoded}-u\|_2/\|u\|_2$ & Interpretation\\
\midrule
$1.20$ (full payload) & $8.45\times10^{-15}$ & machine-precision crop reconstruction\\
$1.18$ & $9.28\times10^{-6}$ & below $10^{-5}$, but exceeds the original storage\\
$1.10$ & $3.62\times10^{-5}$ & $10^{-5}$--$10^{-4}$ accuracy range\\
$1.00$ & $6.57\times10^{-5}$ & $10^{-5}$--$10^{-4}$ accuracy range at equal storage\\
$0.90$ & $9.46\times10^{-5}$ & $10^{-4}$ accuracy with $10\%$ fewer stored values\\
$0.50$ & $3.01\times10^{-4}$ & stronger compression with reduced accuracy\\
\bottomrule
\end{tabularx}
\caption{Open-channel Fourier-extension storage--accuracy frontier.  The extension uses $g=1.2$ and axial density $r=2$ before coefficient pruning.  Every nonzero retained mode uses only the intrinsic $D-1$ transverse coordinates; no longitudinal residual is stored.}
\label{tab:open_extension_storage}
\end{table}

These results establish two positive facts.  First, different inflow and outflow profiles do not prevent high-accuracy $D\to D-1\to D$ reconstruction: after sufficient Fourier extension and spectral oversampling, the strict periodic divergence-free extension is recovered to machine precision.  Second, the same representation can be truncated below the original total data count while retaining a relative physical-domain error of order $10^{-4}$ (and the $10^{-5}$--$10^{-4}$ range at equal storage).  For this single-field, magnitude-pruning experiment, a strict error below $10^{-5}$ required $1.18$ times the original storage.  Hence per-mode $D\to D-1$ reduction is exact, while a guarantee of both sub-$10^{-5}$ error and lower total storage would require additional data-dependent compression structure beyond independent Fourier-coefficient pruning.

\paragraph{Why the early open-boundary reconstructions were inaccurate.}
The initial low-density Fourier extensions produced percent-level physical-domain discrepancies (often around $1\%$--$2\%$) and, in the poorly resolved virtual region, could generate very large amplitudes.  This was not a failure of the $D\to D-1$ codec: encoding and decoding an already admissible periodic solenoidal extension is lossless to roundoff.  The issue was that a coarse Fourier representation could not simultaneously resolve the continuation and enforce the required constraints robustly.  Increasing the axial Fourier-coefficient density supplied the missing degrees of freedom; with the exact solenoidal constraint imposed on this refined representation, the virtual-region amplitude remained controlled and the physical crop converged to machine precision.  Thus Fourier refinement is a numerical-stability requirement of the extension construction, not an extra learned channel or a change to the intrinsic $D-1$ representation.

\paragraph{Spectral versus finite-difference divergence.}
For discretized data, ``divergence-free'' is defined relative to the differentiation operator used to test it.  A field can satisfy $D^{\rm FD}_p\!\cdot u\approx0$ for a finite-difference stencil of order $p$ while failing the FFT condition $\mathcal F^{-1}[ik\cdot\widehat u]=0$ required by the transverse Fourier codec, because the finite-difference multiplier $\widetilde k_p(kh)$ is generally not the spectral multiplier $k$.  Low-order stencils can therefore report a small finite-difference divergence while leaving appreciable longitudinal Fourier content.  In the derivative-order diagnostic, raising the centered stencil order through $8$, $10$, and $12$ made the finite-difference diagnostic progressively closer to the spectral result on resolved interior modes.  It remains an approximation at any finite order, however; strict spectral divergence-freeness is established only by the Fourier-space condition itself (or by a representation built for the solver's own discrete symbol).

\begin{figure}[H]
\centering
\includegraphics[width=\linewidth,height=0.83\textheight,keepaspectratio]{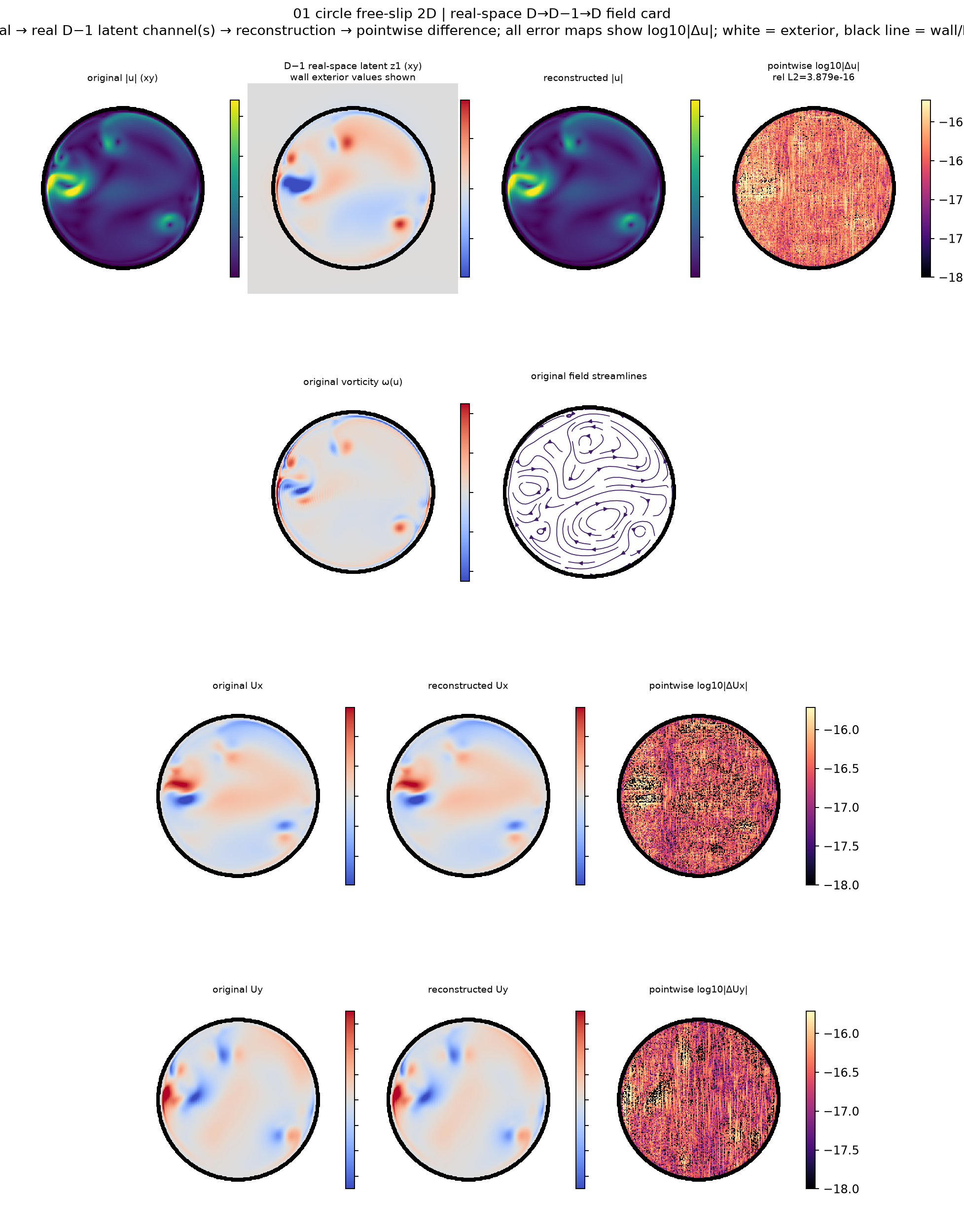}
\par\smallskip
\noindent\textbf{2D circular free-slip cavity.} This is a two-dimensional disk.  Its circumference is a closed free-slip and impermeable boundary ($u\cdot n=0$), so it has neither periodic identification nor inlet/outlet Fourier extension.  \latentexteriornote\par
\end{figure}

\clearpage
\begin{figure}[H]
\centering
\includegraphics[width=\linewidth,height=0.83\textheight,keepaspectratio]{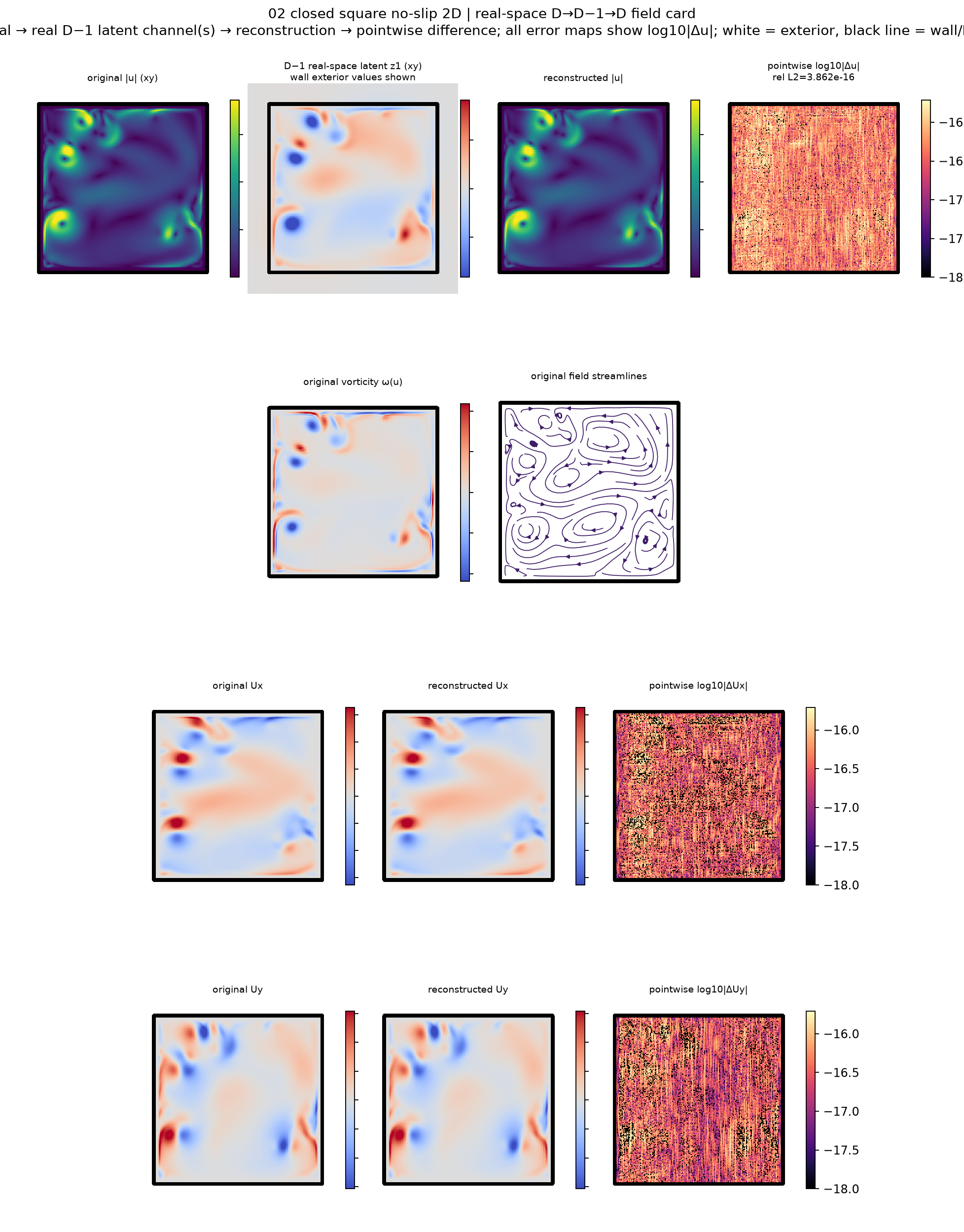}
\par\smallskip
\noindent\textbf{2D closed square with no-slip walls.} This is a two-dimensional square cavity.  All four sides are closed, impermeable no-slip walls; it is not periodic and has no inflow or outflow.  \latentexteriornote\par
\end{figure}

\clearpage
\begin{figure}[H]
\centering
\includegraphics[width=\linewidth,height=0.83\textheight,keepaspectratio]{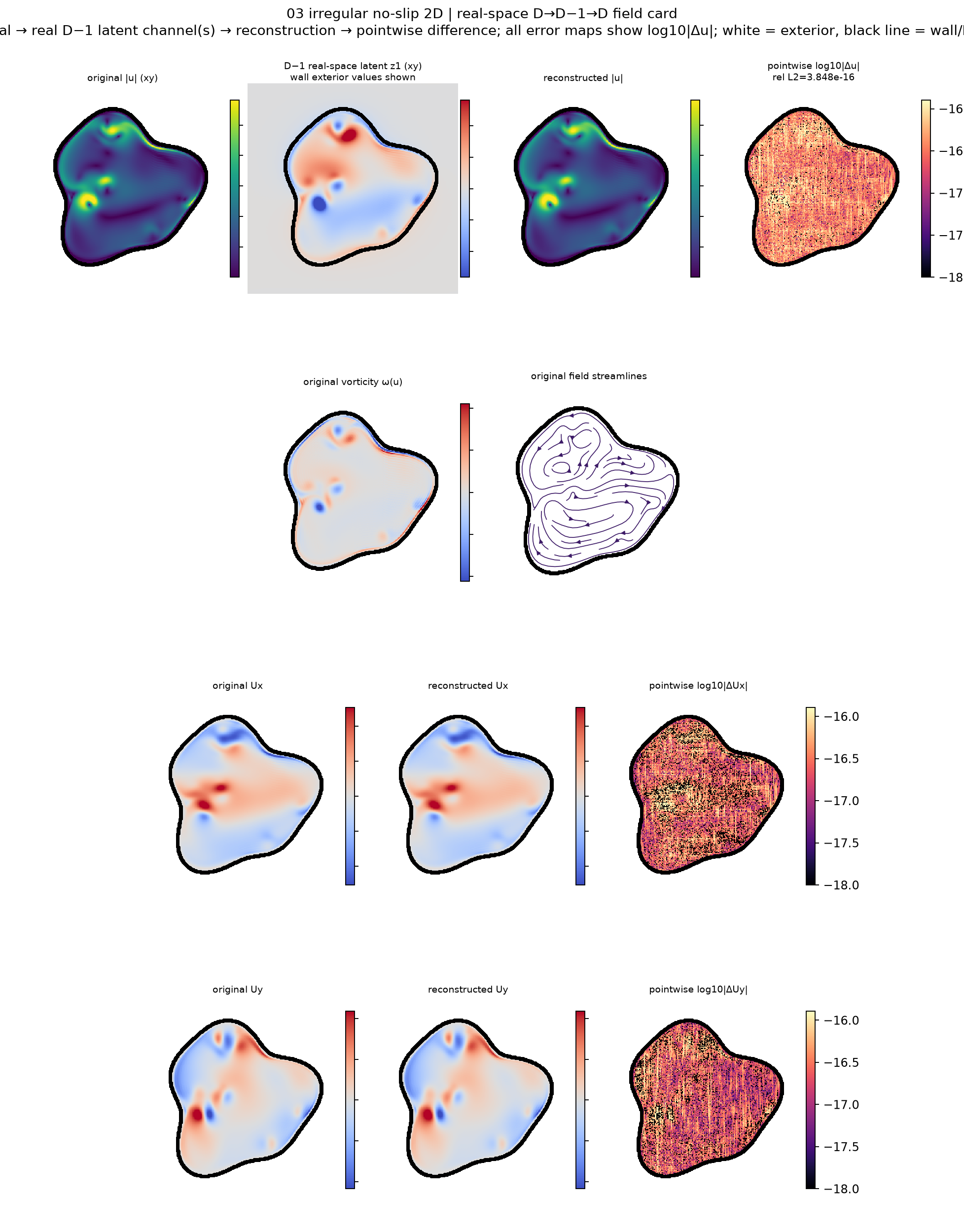}
\par\smallskip
\noindent\textbf{2D irregular closed no-slip domain.} This two-dimensional simply connected irregular domain is bounded by a solid no-slip wall.  The exterior is masked white, and there is no periodic matching or open-boundary extension.  \latentexteriornote\par
\end{figure}

\clearpage
\begin{figure}[H]
\centering
\includegraphics[width=\linewidth,height=0.83\textheight,keepaspectratio]{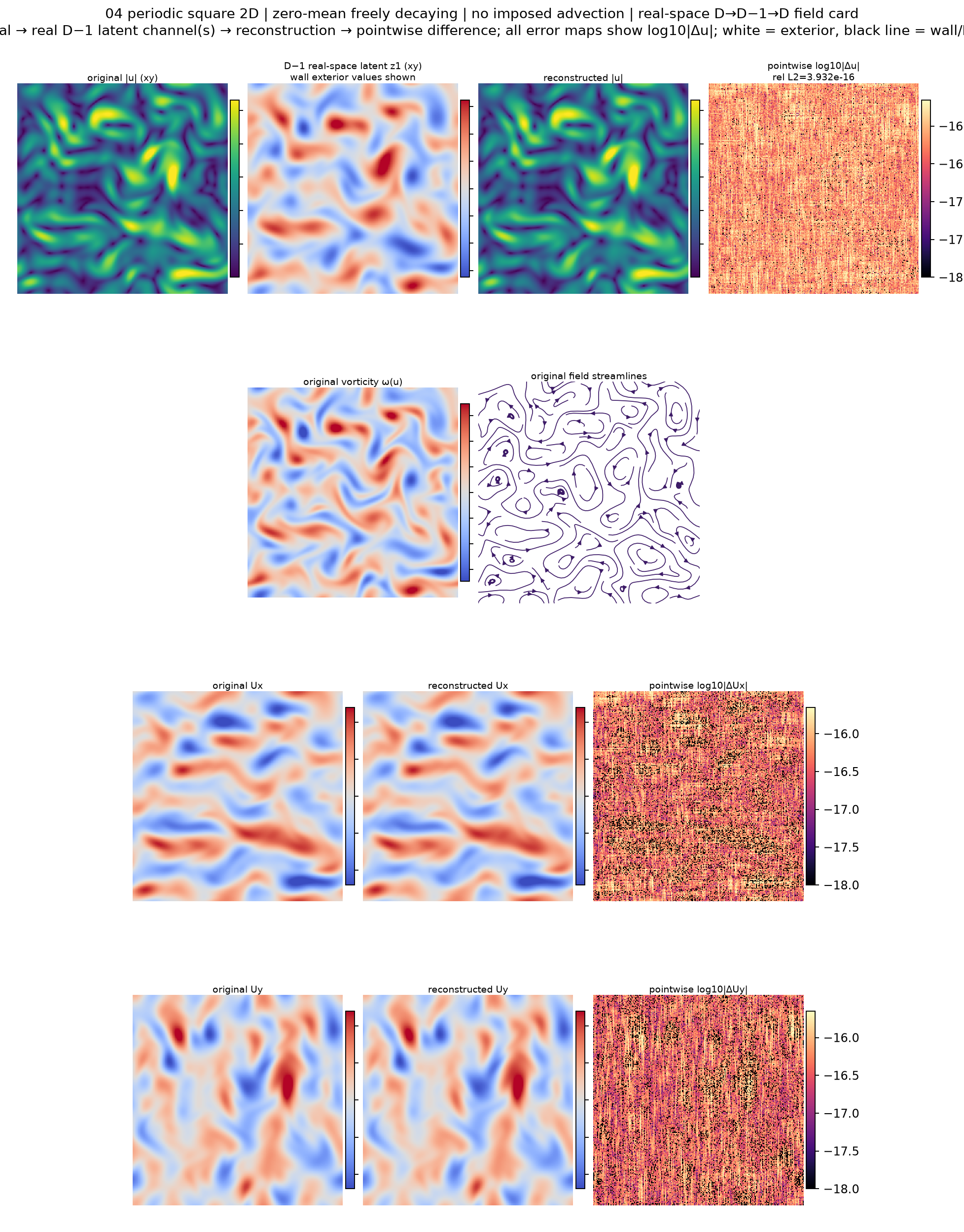}
\par\smallskip
\noindent\textbf{2D periodic square turbulence.} This is a two-dimensional square with opposite sides identified, i.e., a periodic torus.  It has no physical wall or inlet/outlet boundary, so no Fourier continuation is required.\par
\end{figure}

\clearpage
\begin{figure}[H]
\centering
\includegraphics[width=\linewidth,height=0.83\textheight,keepaspectratio]{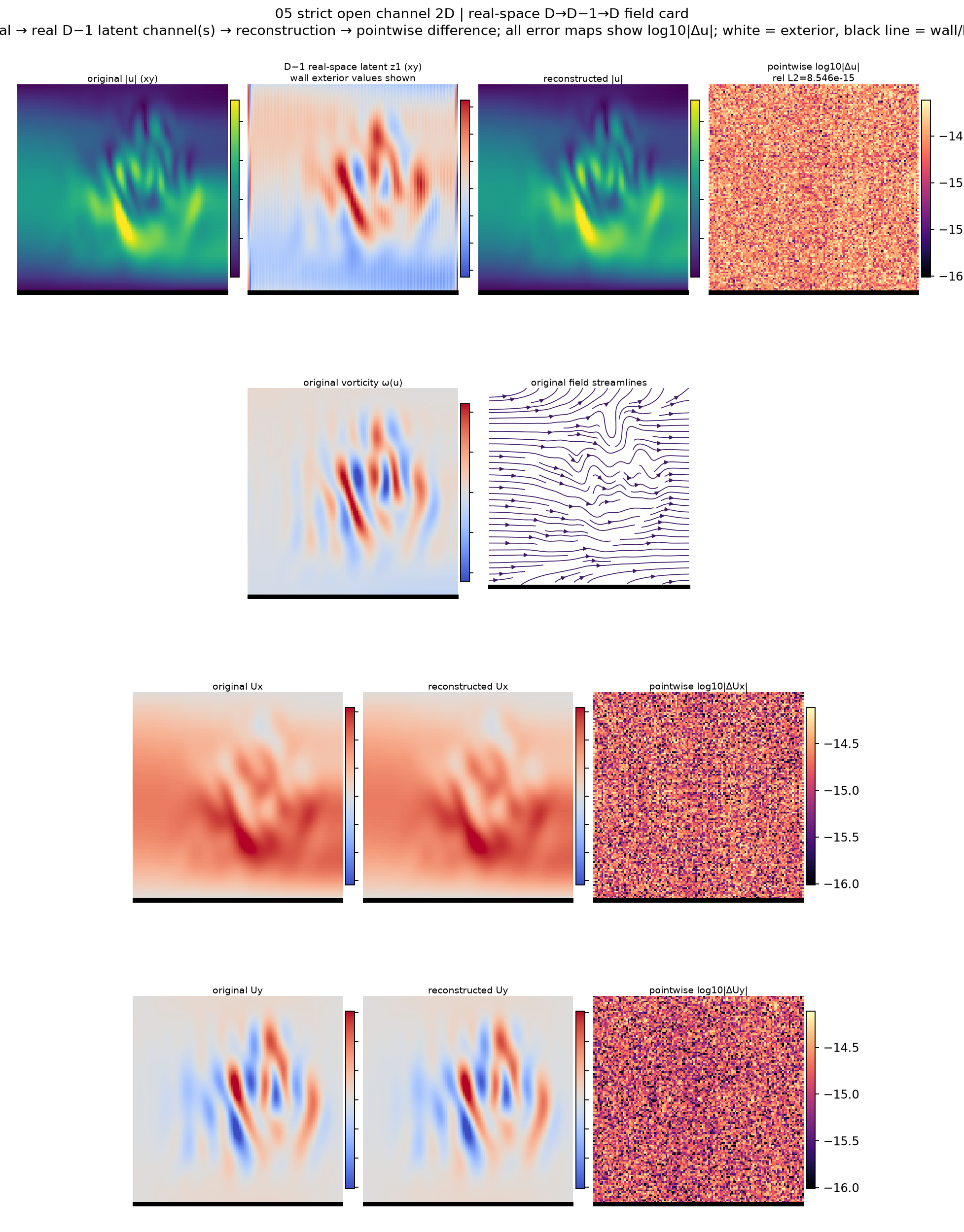}
\par\smallskip
\noindent\textbf{2D strict open channel.} This is a two-dimensional channel with nonperiodic inlet/outlet ends.  Its Fourier representation therefore uses an extended periodic box rather than imposing direct periodicity on the physical open channel.  The virtual portion is an intentional Fourier continuation, not an exterior latent tail of a closed-wall field.\par
\end{figure}

\clearpage
\begin{figure}[H]
\centering
\includegraphics[width=\linewidth,height=0.83\textheight,keepaspectratio]{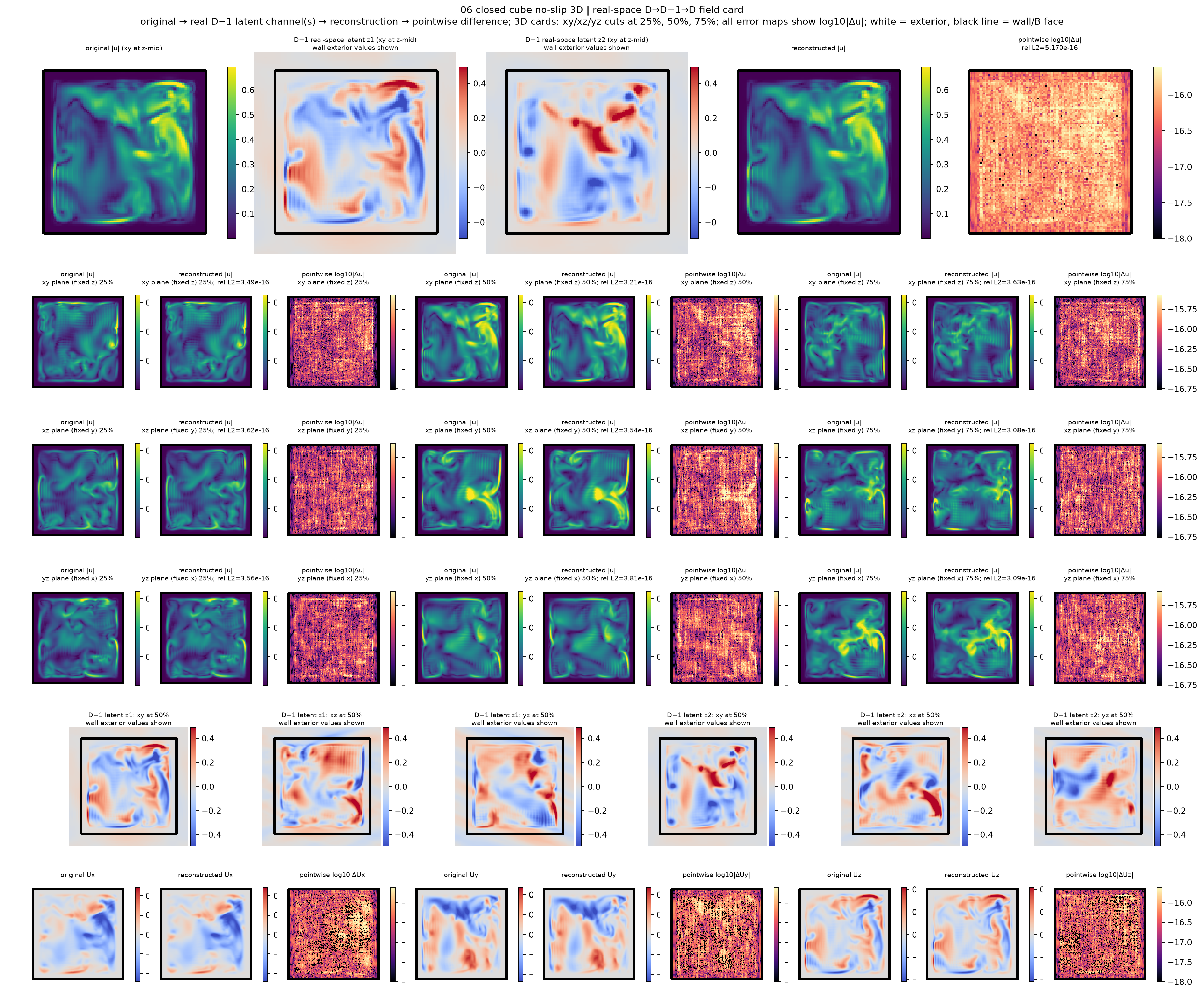}
\par\smallskip
\noindent\textbf{3D closed cube with no-slip walls.} This is a three-dimensional cubical cavity.  Its faces are closed no-slip and impermeable walls, and the $xy$, $xz$, and $yz$ slices expose the enclosed geometry.  \latentexteriornote\par
\end{figure}

\clearpage
\begin{figure}[H]
\centering
\includegraphics[width=\linewidth,height=0.83\textheight,keepaspectratio]{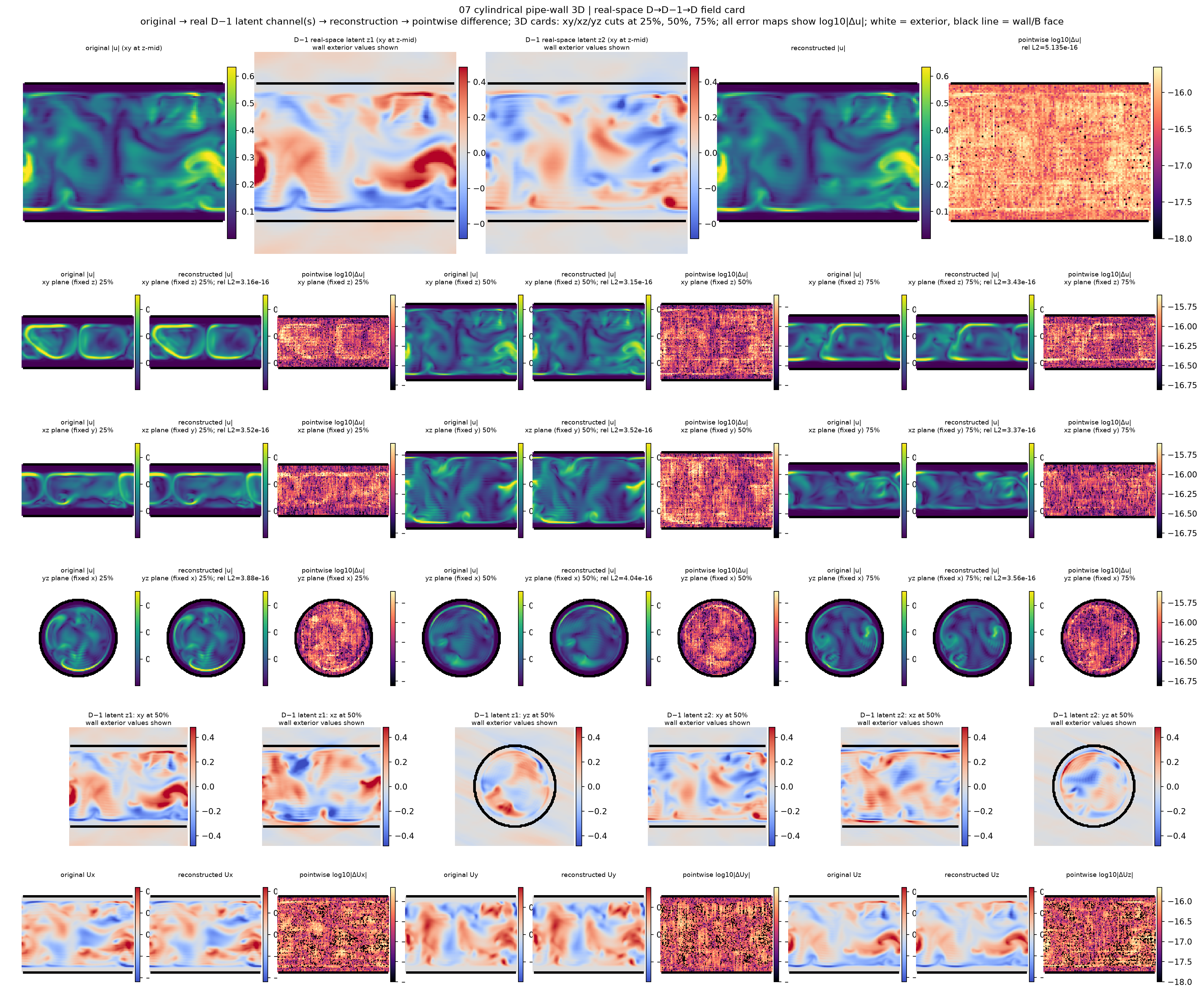}
\par\smallskip
\noindent\textbf{3D cylindrical pipe with side wall.} This is a three-dimensional circular pipe.  The axial end faces are inlet/outlet openings while the cylindrical side wall is solid; a nonperiodic axial treatment requires Fourier extension.  Outside the solid side wall, the $D$-component physical field and displayed reconstruction are deliberately masked, while the $D-1$ latent is left visible because the Fourier/Householder encoder is the nonlocal spatial convolution in Appendix~\ref{app:exterior_latent}.\par
\end{figure}

\clearpage
\begin{figure}[H]
\centering
\includegraphics[width=\linewidth,height=0.83\textheight,keepaspectratio]{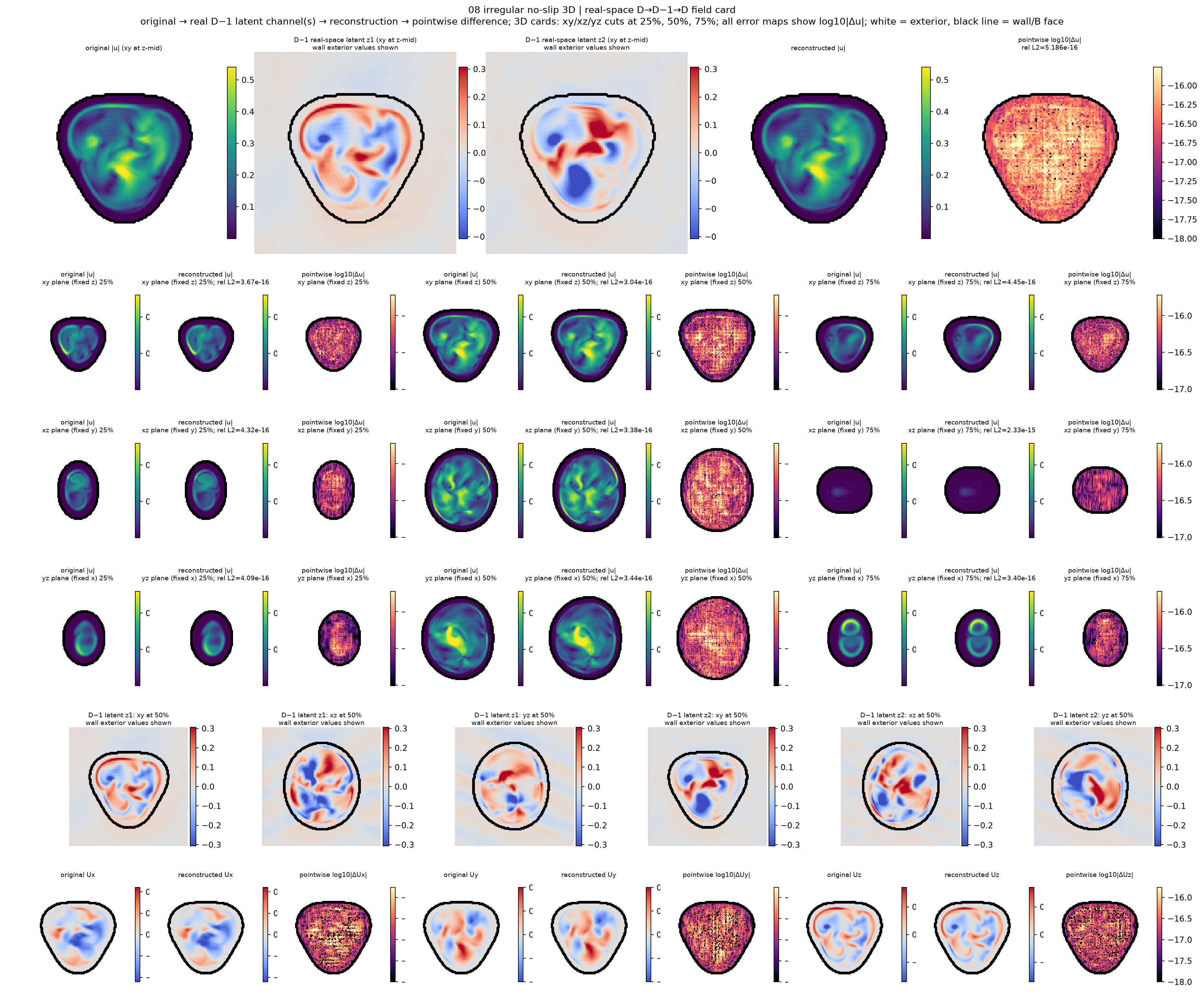}
\par\smallskip
\noindent\textbf{3D irregular closed no-slip domain.} This is a three-dimensional irregular solid-bounded volume.  The black contours mark the closed, impermeable no-slip boundary, while white regions lie outside the physical domain.  \latentexteriornote\par
\end{figure}

\clearpage
\begin{figure}[H]
\centering
\includegraphics[width=\linewidth,height=0.83\textheight,keepaspectratio]{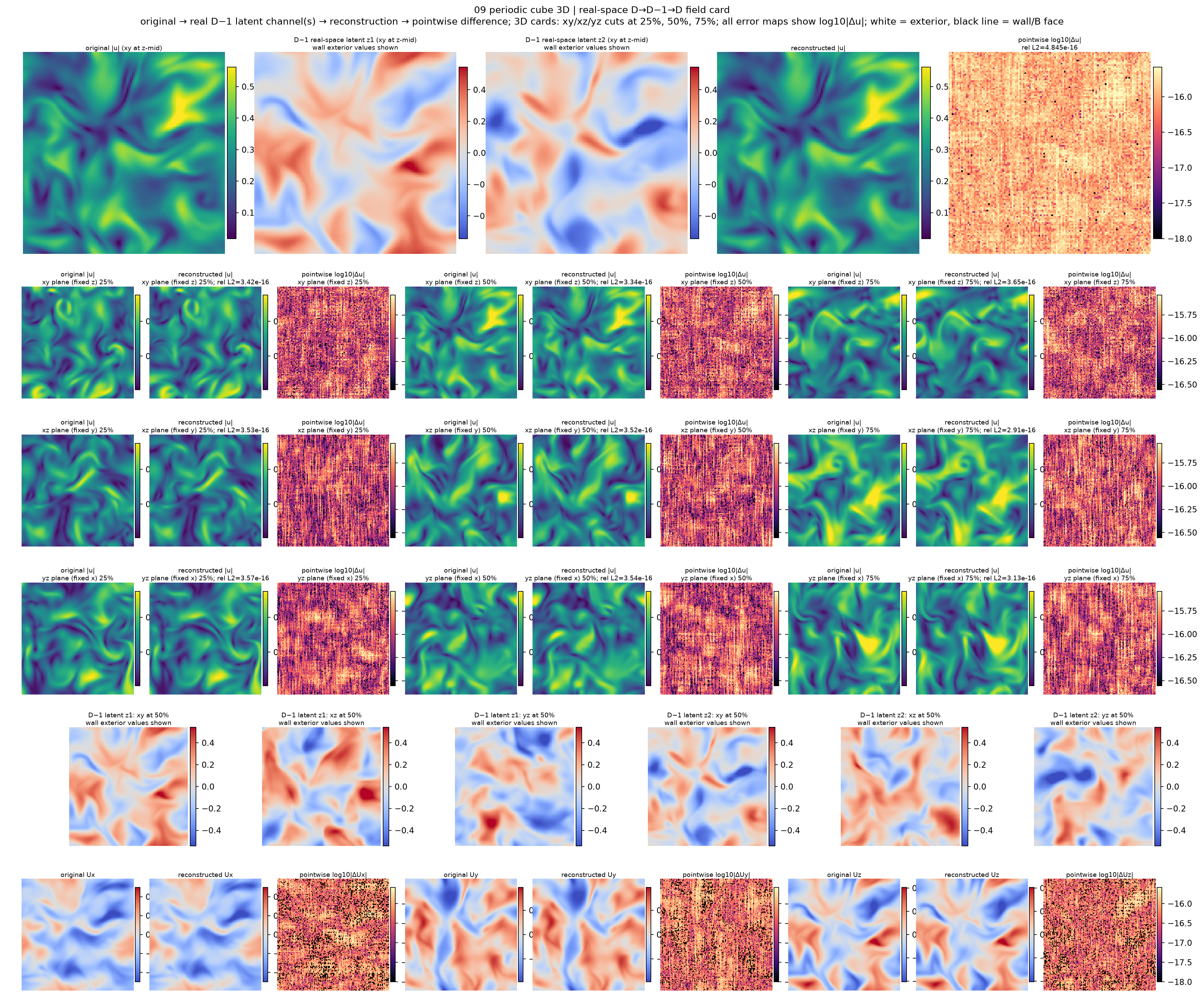}
\par\smallskip
\noindent\textbf{3D periodic cube.} This is a three-dimensional periodic box with opposite faces identified.  It has no physical wall or inlet/outlet boundary and is represented directly in Fourier space.\par
\end{figure}

\clearpage
\begin{figure}[H]
\centering
\includegraphics[width=\linewidth,height=0.83\textheight,keepaspectratio]{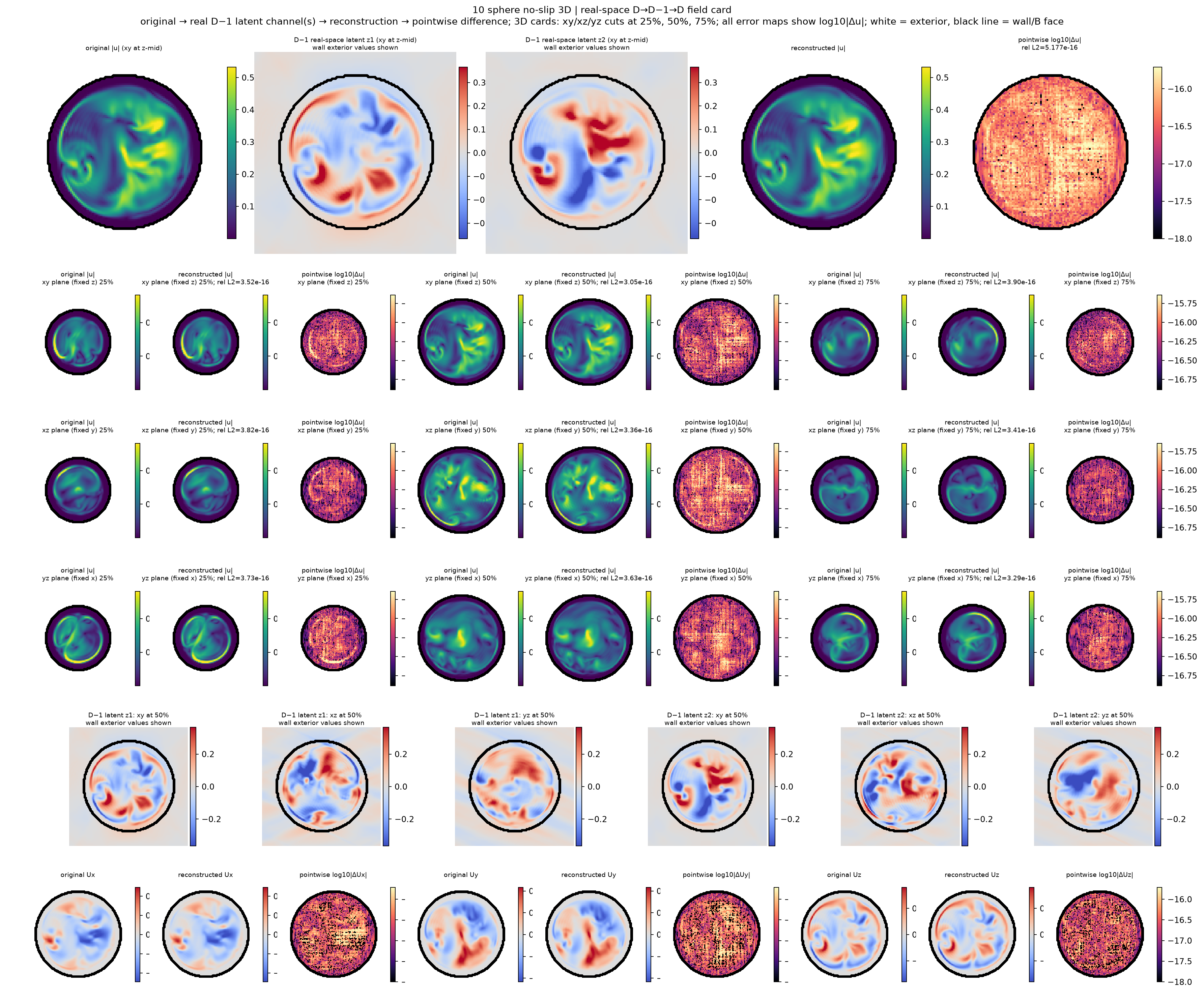}
\par\smallskip
\noindent\textbf{3D spherical no-slip cavity.} This is a three-dimensional spherical enclosed domain.  Its spherical surface is a closed no-slip, impermeable wall rather than a periodic or open boundary.  \latentexteriornote\par
\end{figure}

\clearpage
\begin{figure}[H]
\centering
\includegraphics[width=\linewidth,height=0.83\textheight,keepaspectratio]{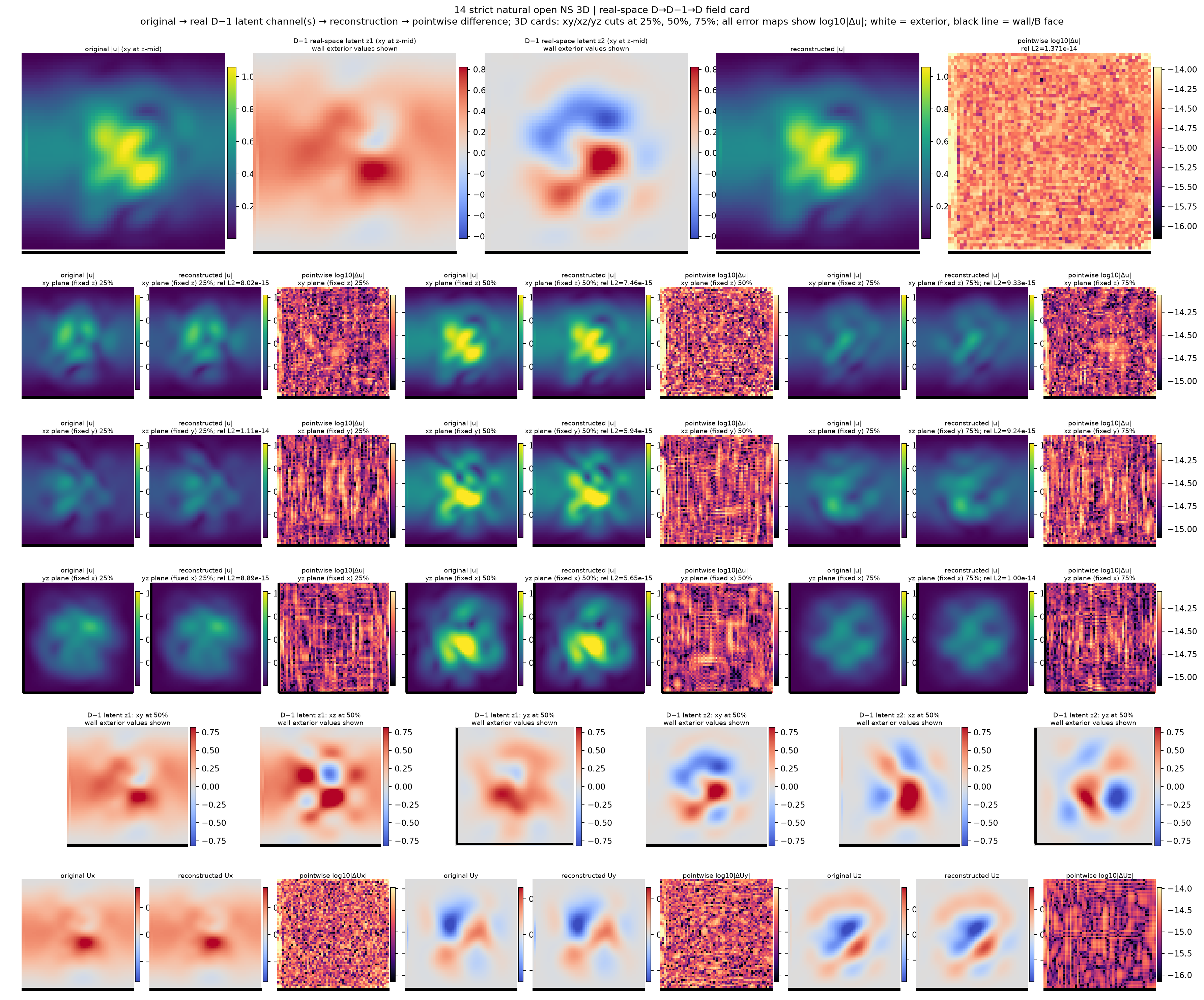}
\par\smallskip
\noindent\textbf{3D natural-open Navier--Stokes field.} This is a three-dimensional nonperiodic open-flow state.  The open axial direction is encoded through Fourier extension to an enlarged periodic box rather than by treating the physical domain as periodic.  Its virtual box region is part of that intentional continuation, not a closed-boundary exterior latent.\par
\end{figure}

\clearpage
\begin{figure}[H]
\centering
\includegraphics[width=\linewidth,height=0.83\textheight,keepaspectratio]{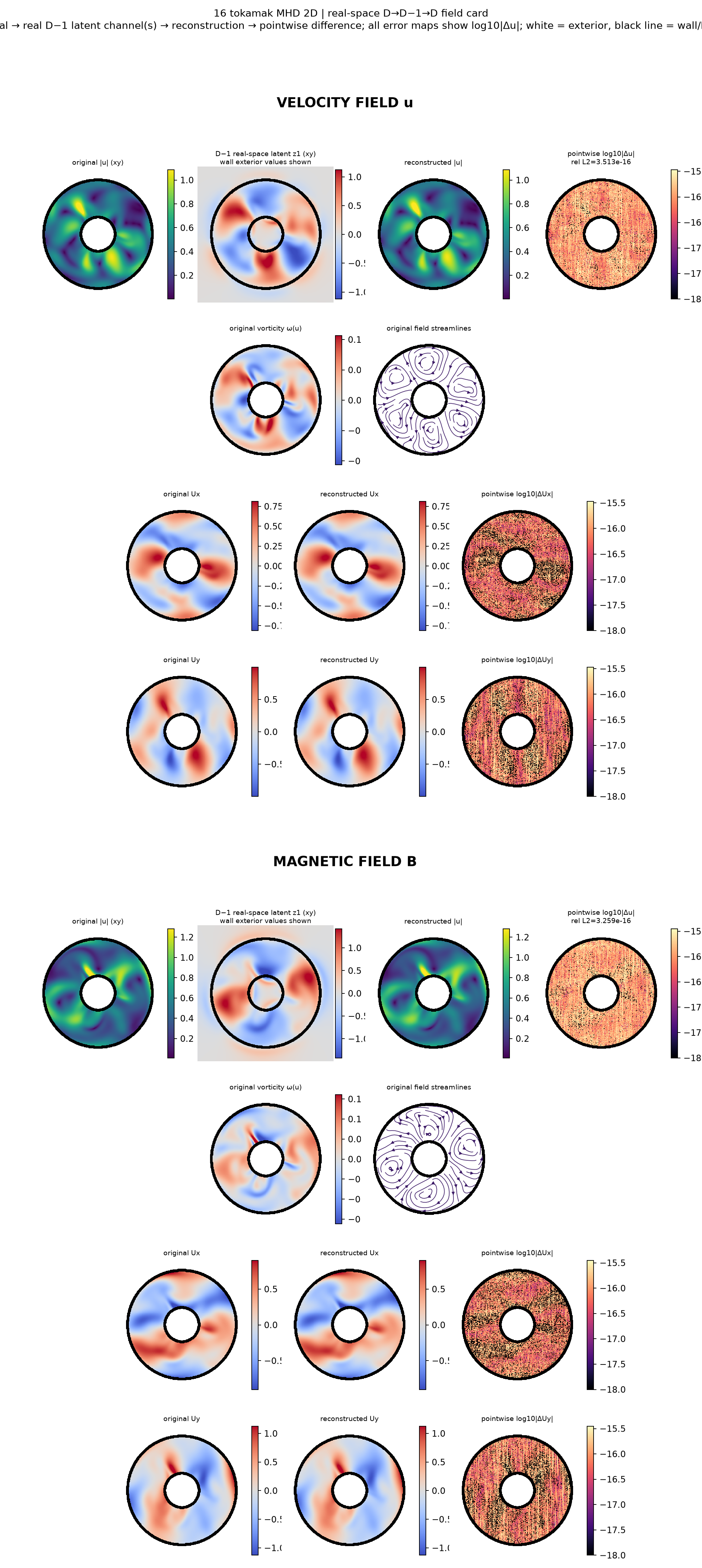}
\par\smallskip
\noindent\textbf{2D tokamak MHD cross-section.} This figure contains both velocity $u$ and magnetic field $B$ on a bounded annular tokamak cross-section.  The black curves mark the closed physical boundary in the displayed plane; it is not an inlet/outlet extension setting.  \latentexteriornote\par
\end{figure}

\clearpage
\begin{figure}[H]
\centering
\includegraphics[width=\linewidth,height=0.83\textheight,keepaspectratio]{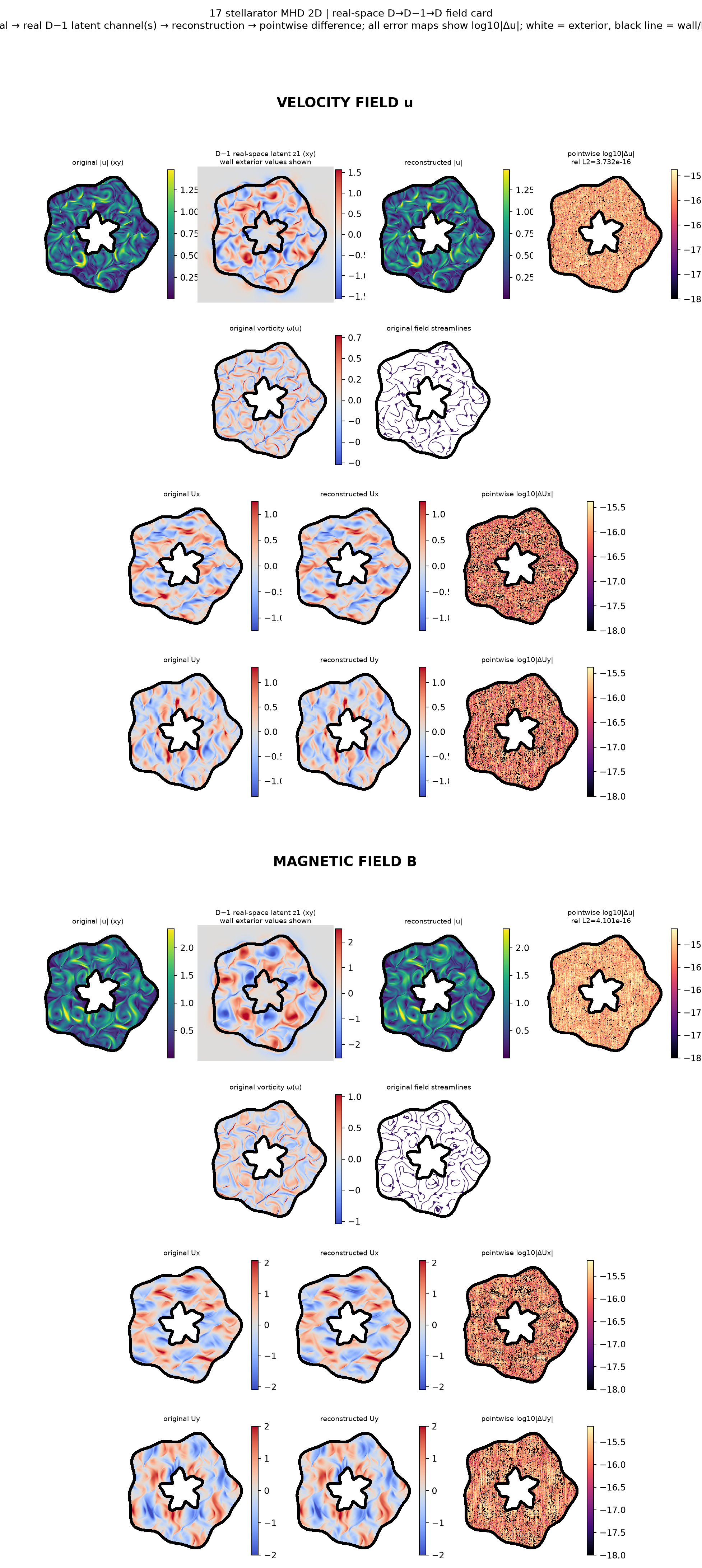}
\par\smallskip
\noindent\textbf{2D stellarator MHD cross-section.} This figure contains both $u$ and $B$ on an irregular annular stellarator cross-section.  The physical cross-section is wall bounded and closed, not a directly periodic or open-channel geometry.  \latentexteriornote\par
\end{figure}

\clearpage
\begin{figure}[H]
\centering
\includegraphics[width=\linewidth,height=0.83\textheight,keepaspectratio]{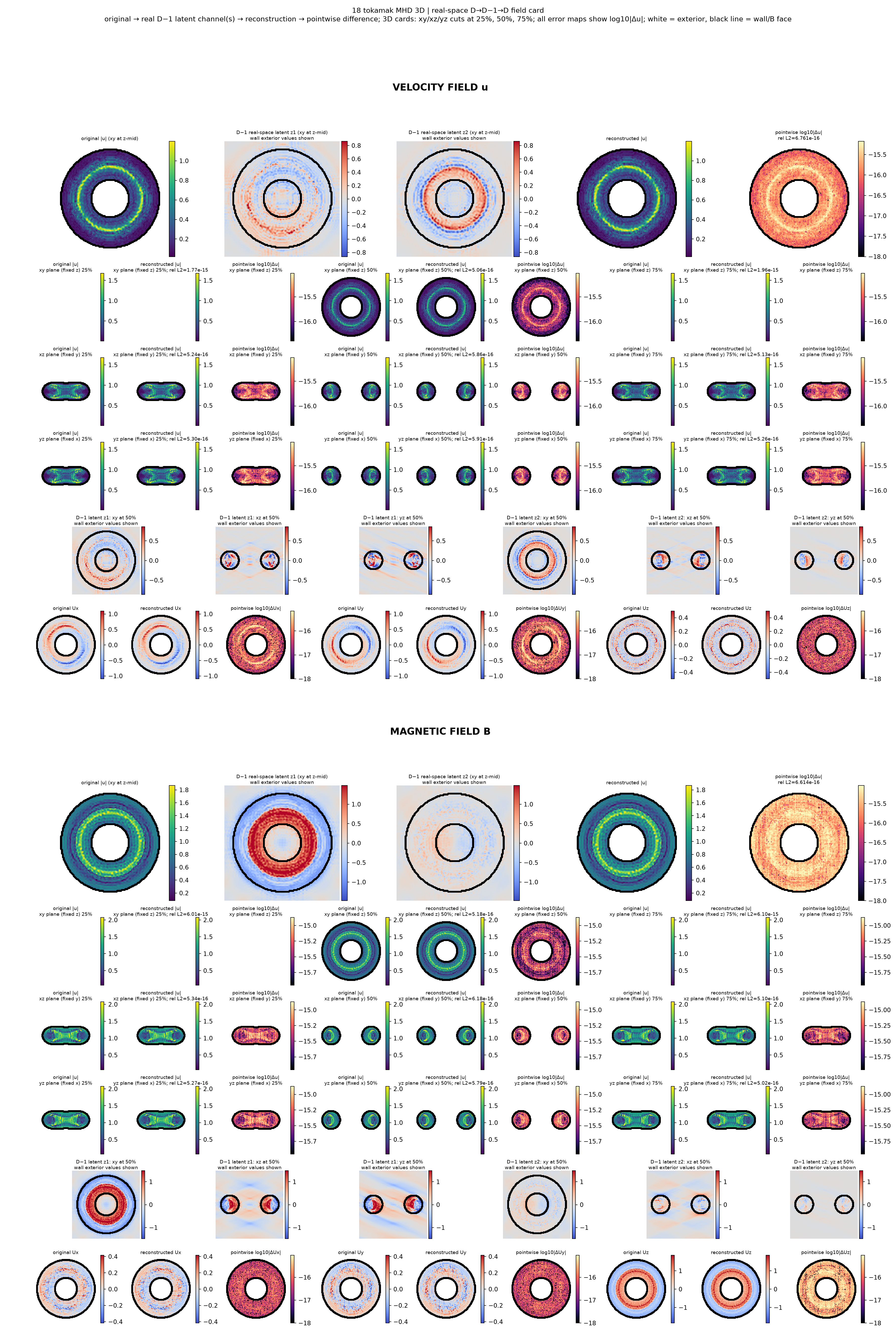}
\par\smallskip
\noindent\textbf{3D tokamak MHD configuration.} This is a three-dimensional toroidal MHD configuration with velocity and magnetic fields.  The toroidal coordinate is periodic, whereas the cross-sectional plasma/wall boundary is closed; no inlet/outlet extension is used.  \latentexteriornote\par
\end{figure}

\clearpage
\begin{figure}[H]
\centering
\includegraphics[width=\linewidth,height=0.83\textheight,keepaspectratio]{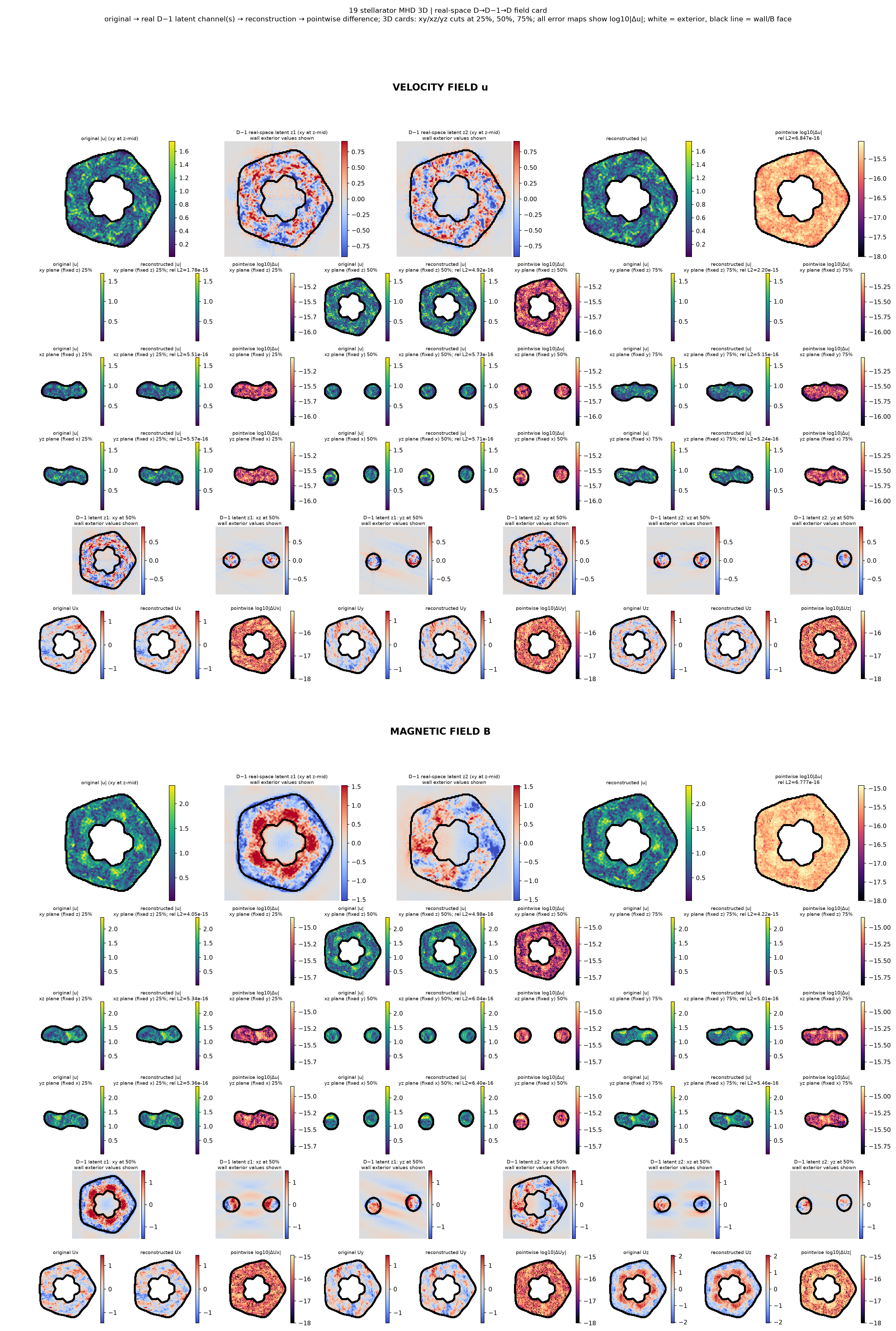}
\par\smallskip
\noindent\textbf{3D stellarator MHD configuration.} This is a three-dimensional twisted toroidal stellarator configuration with both $u$ and $B$.  The toroidal direction is periodic while the irregular cross-sectional boundary is closed, not an open inflow/outflow boundary.  \latentexteriornote\par
\end{figure}

\clearpage
\begin{figure}[H]
\centering
\includegraphics[width=\linewidth,height=0.83\textheight,keepaspectratio]{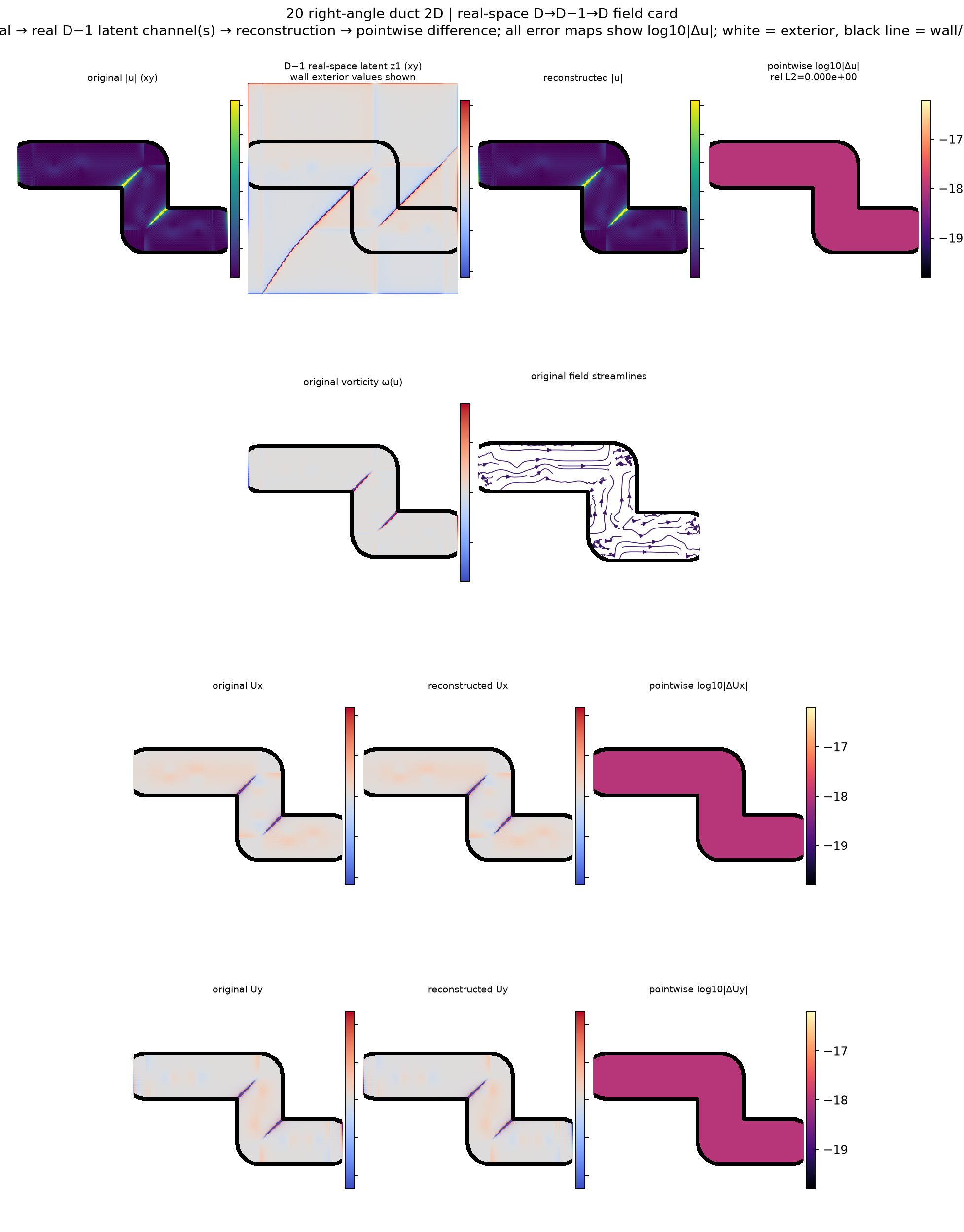}
\par\smallskip
\noindent\textbf{2D right-angle duct.} This is a two-dimensional right-angle duct enclosed by the solid wall shown in black.  It is a wall-bounded, nonperiodic geometry rather than a periodic square or an open-extension benchmark.  \latentexteriornote\par
\end{figure}

\clearpage
\begin{figure}[H]
\centering
\includegraphics[width=\linewidth,height=0.83\textheight,keepaspectratio]{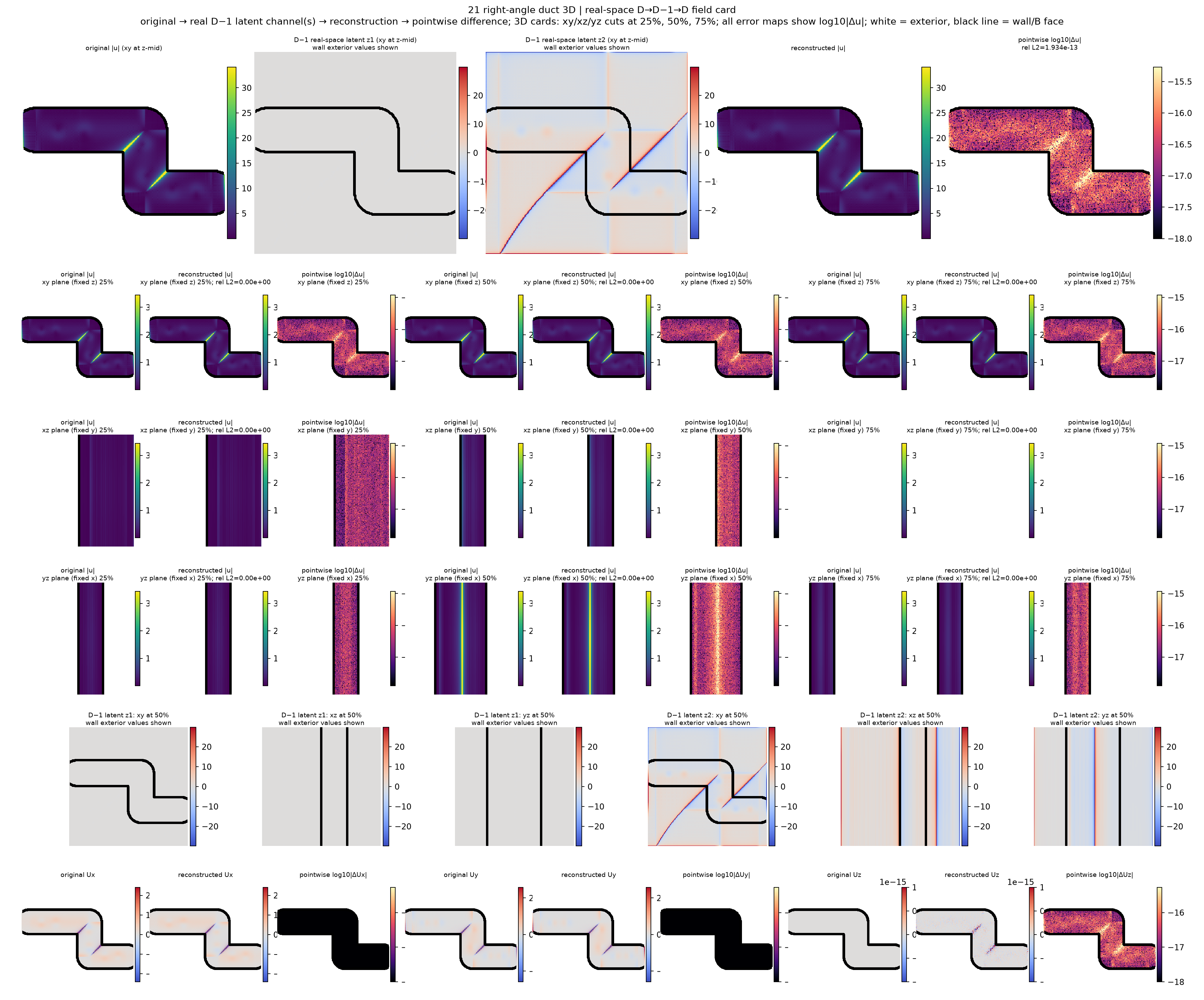}
\par\smallskip
\noindent\textbf{3D right-angle duct.} This is the three-dimensional counterpart of the right-angle duct.  Its wall-bounded geometry is identified by the black contours across all slice orientations, with no periodic identification imposed.  \latentexteriornote\par
\end{figure}

\clearpage
\begin{figure}[H]
\centering
\includegraphics[width=\linewidth,height=0.83\textheight,keepaspectratio]{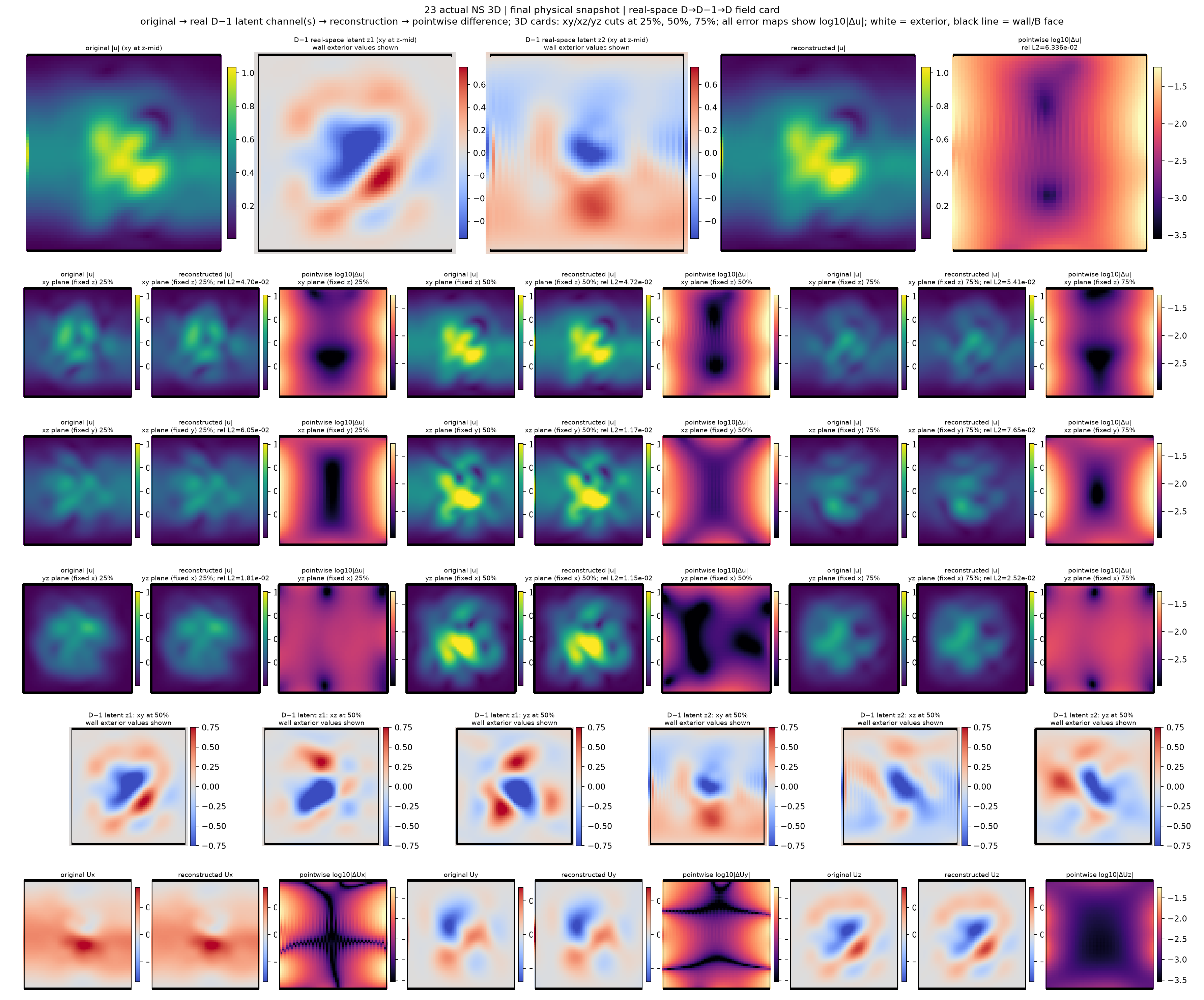}
\par\smallskip
\noindent\textbf{3D Navier--Stokes final snapshot.} This is an actual three-dimensional Navier--Stokes physical snapshot.\par
\end{figure}

\clearpage
\begin{figure}[H]
\centering
\includegraphics[width=\linewidth,height=0.83\textheight,keepaspectratio]{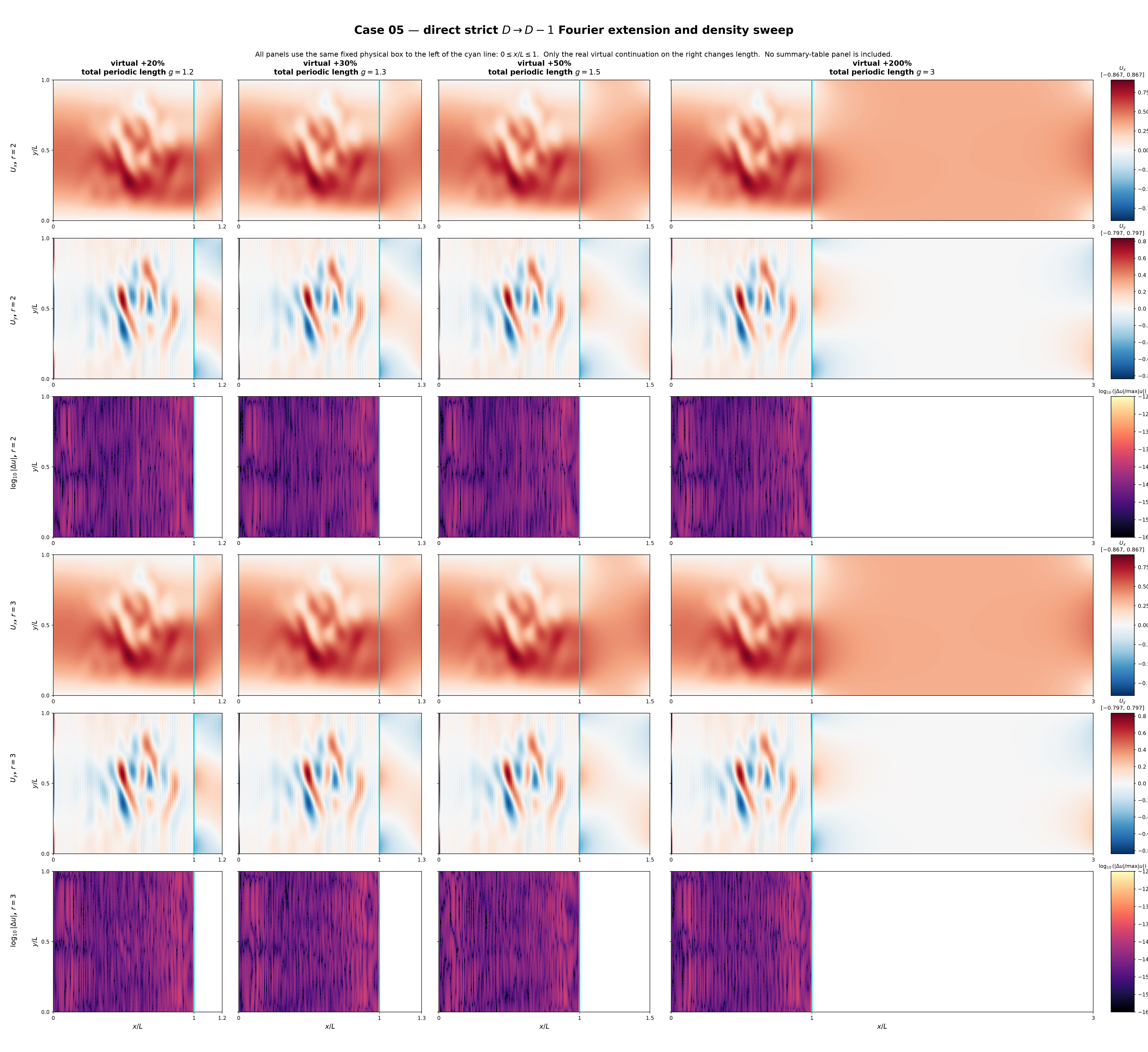}
\par\smallskip
\noindent\textbf{Fourier-extension length and density sweep.} This composite uses the 2D open channel and compares 20\%, 30\%, 50\%, and 200\% virtual extensions ($g=1.2,1.3,1.5,3.0$) at axial Fourier densities $r=2$ and $r=3$.  The cyan divider separates the physical region from the intentional Fourier-extended virtual region.\par
\end{figure}

\clearpage
\section{Redundancy, Optimization, and Identifiability Diagnostics}
\label{app:redundancy_optimization}

This section records representation-learning diagnostics separately from the evolution experiments.  The aim is to distinguish two facts that are often conflated: a redundant parameterization can contain directions that do not change the physical divergence-free field, and a particular optimization run can nevertheless be easier or harder for reasons that include architecture, conditioning, and optimization dynamics.  The results below document both the observed optimization behavior and the exact null-space mechanism.

\subsection{Three-dimensional ball evolving in time: a $D=4$ space--time field}

The first study treats a three-dimensional ball over time as a four-coordinate space--time field $(x,y,z,t)$ and learns its conserved current.  Five parameterizations with approximately $26$k trainable parameters were trained for $600$ seconds each.  Every parameterization used the same fixed stochastic collocation stream; only the ten random neural-network initializations were changed.  The reported loss is the best fixed-validation loss attained by a run, while the final two columns make the optimization speed explicit.

\begin{table}[H]
\centering
\small
\setlength{\tabcolsep}{3pt}
\renewcommand{\arraystretch}{1.15}
\begin{tabularx}{\linewidth}{@{}>{\raggedright\arraybackslash}p{0.31\linewidth}>{\centering\arraybackslash}p{0.15\linewidth}>{\centering\arraybackslash}p{0.18\linewidth}>{\centering\arraybackslash}p{0.17\linewidth}>{\centering\arraybackslash}X@{}}
\toprule
Representation & rank / nullity & best validation loss & optimizer steps & loss decades / s \\
\midrule
Minimal Fourier $a\in\mathbb R^3$ (ours) & $3/0$ & $2.734\!\times\!10^{-2}\pm2.619\!\times\!10^{-3}$ & $16{,}106\pm80$ & $5.286\!\times\!10^{-3}$ \\
Direct complex Fourier coefficients & $3/0$ & $7.735\pm1.404\!\times\!10^{-2}$ & $16{,}205\pm71$ & $1.199\!\times\!10^{-3}$ \\
Quaternion three-potential & $3/0$ & $1.373\!\times\!10^{-3}\pm1.977\!\times\!10^{-4}$ & $28{,}306\pm909$ & $7.445\!\times\!10^{-3}$ \\
NCL vector auxiliary field $B\in\mathbb R^4$ & $3/1$ & $2.516\!\times\!10^{-2}\pm3.241\!\times\!10^{-3}$ & $15{,}034\pm54$ & $5.400\!\times\!10^{-3}$ \\
NCL antisymmetric potential $A\in\mathbb R^6$ & $3/3$ & $1.039\!\times\!10^{-3}\pm6.298\!\times\!10^{-5}$ & $28{,}351\pm233$ & $7.661\!\times\!10^{-3}$ \\
\bottomrule
\end{tabularx}
\caption{Ten-initialization representation-learning record for the three-dimensional ball over time.  Mean $\pm$ standard deviation is reported for loss and optimizer steps.  The rank and nullity are per nonzero $D=4$ Fourier mode.}
\label{tab:ball_redundancy_record}
\end{table}

In this capacity-matched, fixed-budget record, the antisymmetric NCL potential reached the lowest validation loss and the fastest loss decrease among the listed alternatives.  The paired runs also gave lower final validation loss for the NCL potential in all ten comparisons and a broader, flatter measured low-loss neighborhood.  The following separate controlled sweep is the missing experiment that deliberately increases the redundancy of the potential representation.

\clearpage
\subsection{Controlled sweep with deliberately overcomplete potential coordinates}

Here every model solves the same $D=4$ Ball evolution problem for exactly $10{,}000$ updates, with the same sampler, optimizer, hidden architecture, initialization seed, and stochastic collocation stream.  Only the learned potential coordinates and the fixed map from those coordinates to the physical current are changed.  Thus the final validation loss after the common budget is the optimization-progress comparison, while the wall-clock column reports its computational cost.  The direct antisymmetric baseline predicts the six independent entries of $A$.  In the dense rows, an unconstrained vector $z\in\mathbb R^m$ is first mapped by a fixed dense co-isometry $P$ to those six entries, $A=\operatorname{skew}(Pz)$.  In the multi-potential rows, several independently predicted antisymmetric potentials are summed.  In the outer-product rows, no network directly predicts an antisymmetric matrix: each branch predicts unrestricted vectors $p_r,q_r\in\mathbb R^4$ and the decoder constructs
\[
 A=\frac{1}{\sqrt{r/2}}\sum_{s=1}^{r}\bigl(p_s q_s^{\mathsf T}-q_s p_s^{\mathsf T}\bigr).
\]
This is the unrestricted vector--outer-product construction in the original experiment record.

\begin{table}[H]
\centering
\scriptsize
\setlength{\tabcolsep}{3pt}
\renewcommand{\arraystretch}{1.08}
\begin{tabularx}{\linewidth}{@{}>{\raggedright\arraybackslash}p{0.40\linewidth}>{\centering\arraybackslash}p{0.13\linewidth}>{\centering\arraybackslash}p{0.10\linewidth}>{\centering\arraybackslash}p{0.18\linewidth}>{\centering\arraybackslash}X@{}}
\toprule
Potential coordinates & output coordinates & nullity & validation loss at $10$k $\downarrow$ & training s \\
\midrule
Quaternion three-potential construction & $3$ & $0$ & $0.1164051$ & $111.21$ \\
Vector auxiliary field $B$ & $4$ & $1$ & $0.1784628$ & $245.97$ \\
Direct antisymmetric potential $A$ & $6$ & $3$ & $0.1038496$ & $117.72$ \\
Canonical skew potential & $8$ & $5$ & $0.1123825$ & $128.40$ \\
\addlinespace[1pt]
\multicolumn{5}{@{}l}{\emph{Unconstrained dense coordinates followed by a fixed map to $A$}} \\
Dense $z\mapsto A$, $m=12$ & $12$ & $9$ & $0.0974009$ & $121.50$ \\
Dense $z\mapsto A$, $m=18$ & $18$ & $15$ & $0.0988612$ & $117.91$ \\
Dense $z\mapsto A$, $m=24$ & $24$ & $21$ & $0.1001312$ & $127.38$ \\
\addlinespace[1pt]
\multicolumn{5}{@{}l}{\emph{Sum of independently predicted antisymmetric potentials}} \\
$\sum_{s=1}^{2} A_s/\sqrt{2}$ & $12$ & $9$ & $0.1046972$ & $126.65$ \\
$\sum_{s=1}^{3} A_s/\sqrt{3}$ & $18$ & $15$ & $0.1029831$ & $115.75$ \\
$\sum_{s=1}^{4} A_s/2$ & $24$ & $21$ & $0.0986707$ & $112.69$ \\
$\sum_{s=1}^{5} A_s/\sqrt{5}$ & $30$ & $27$ & $0.0986672$ & $116.56$ \\
$\sum_{s=1}^{6} A_s/\sqrt{6}$ & $36$ & $33$ & $0.1047907$ & $120.82$ \\
\addlinespace[1pt]
\multicolumn{5}{@{}l}{\emph{Unrestricted vector outer products, then antisymmetrized by the decoder}} \\
$r=2$ vector pairs & $16$ & $13$ & $0.0951862$ & $128.83$ \\
$r=3$ vector pairs & $24$ & $21$ & $0.0862193$ & $101.95$ \\
$r=4$ vector pairs & $32$ & $29$ & $\mathbf{0.0837907}$ & $116.23$ \\
$r=6$ vector pairs & $48$ & $45$ & $0.0938667$ & $115.80$ \\
\bottomrule
\end{tabularx}
\caption{Full controlled redundancy sweep from the original $10$k-update Ball experiment.  ``Nullity'' is the generic output-coordinate nullity of the map to the three physical current components, i.e., output coordinates minus rank $3$.  The deliberately overcomplete dense, multi-potential, and outer-product constructions mostly reach a lower loss than direct six-entry $A$ at comparable training time.}
\label{tab:controlled_redundancy_sweep}
\end{table}

\subsection{All ten initializations: three reduced components versus four $B$ and six $A$ components}

The following three displays show every learned native component at the same slice $z=0$, $t=0.25$, with one row per random initialization.  They are the representation fields themselves, not the decoded physical current.  The three reduced components learned by our method collapse to the same visible fields across the ten runs.  In contrast, the four components of the NCL vector potential $B$ and the six independent components of the antisymmetric NCL matrix potential $A$ vary visibly from initialization to initialization.  This is the direct visual distinction needed here: our $3$-component coordinate has no gauge/null direction, whereas the $4$-component $B$ and $6$-component $A$ representations contain redundant degrees of freedom.  Once decoded, potential differences that lie in those null directions do not alter the physical divergence-free field.

\begin{figure}[H]
\centering
\includegraphics[height=0.70\textheight,keepaspectratio]{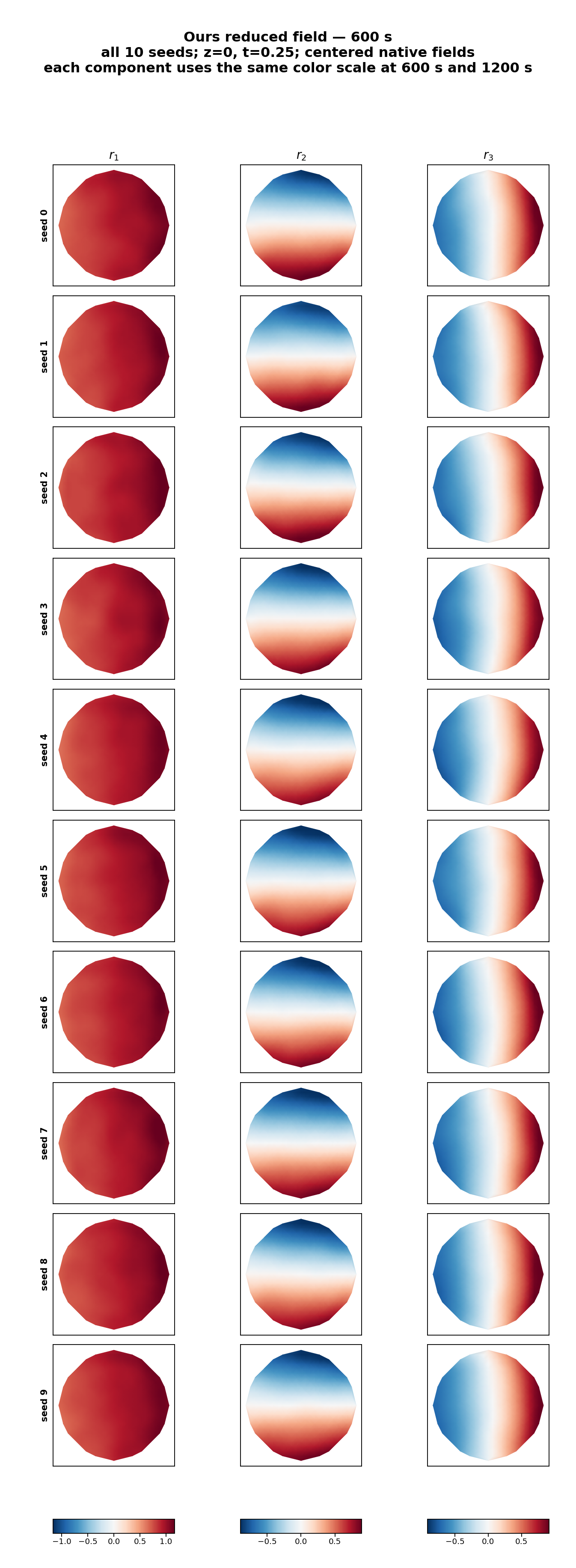}
\par\smallskip
\noindent\textbf{All ten learned three-component reduced fields.} Each row is one independent initialization and the columns are the three intrinsic reduced components.  At this common slice, the ten learned fields agree to visual resolution.\par
\end{figure}

\clearpage
\begin{figure}[H]
\centering
\includegraphics[width=\linewidth,height=0.83\textheight,keepaspectratio]{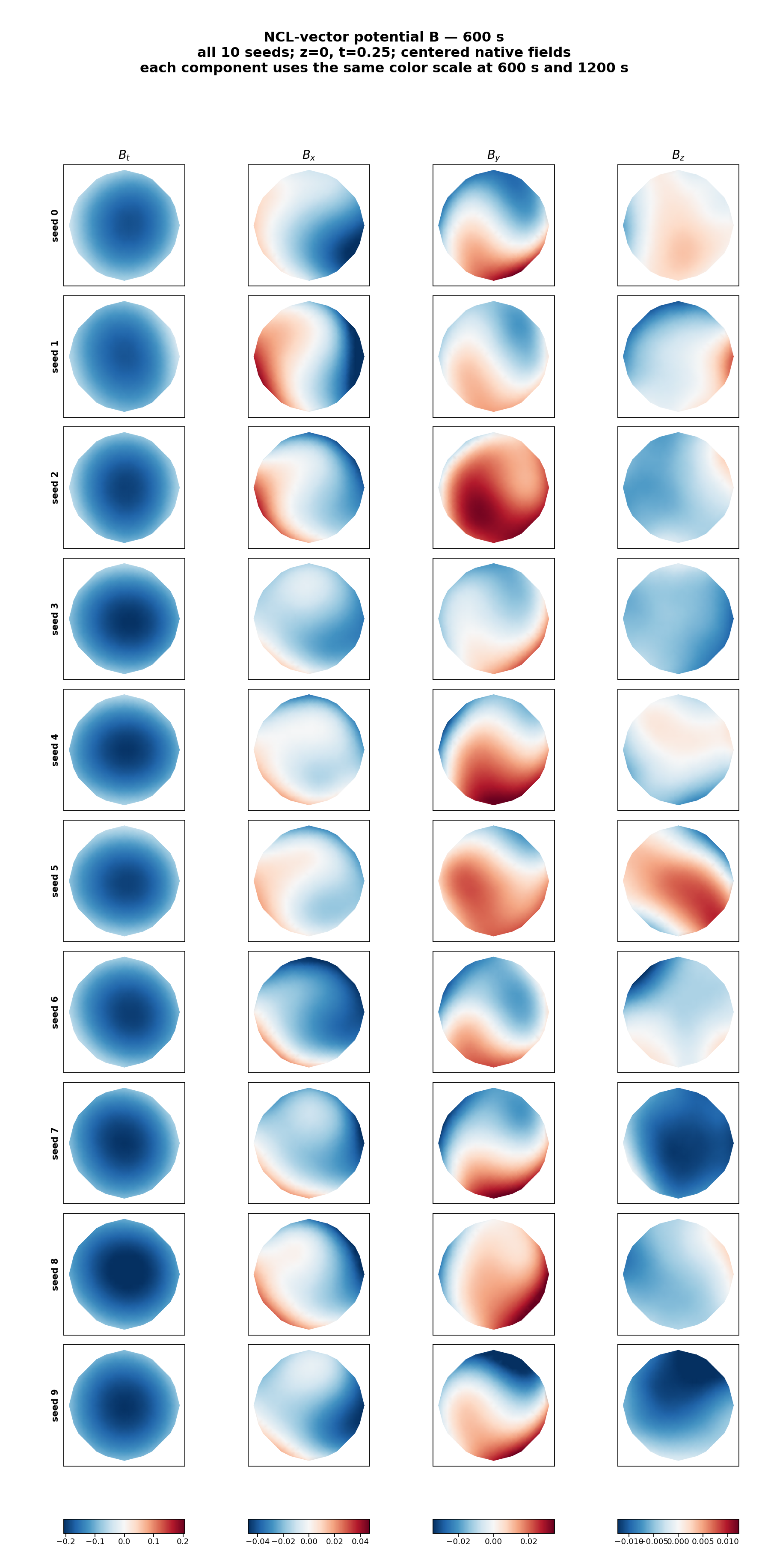}
\par\smallskip
\noindent\textbf{All ten learned four-component vector potentials $B$.} Each row is one independent initialization and the columns are the four potential components.  Unlike the intrinsic three-component coordinates, these native auxiliary fields visibly change between runs.\par
\end{figure}

\clearpage
\begin{figure}[H]
\centering
\includegraphics[width=\linewidth,height=0.83\textheight,keepaspectratio]{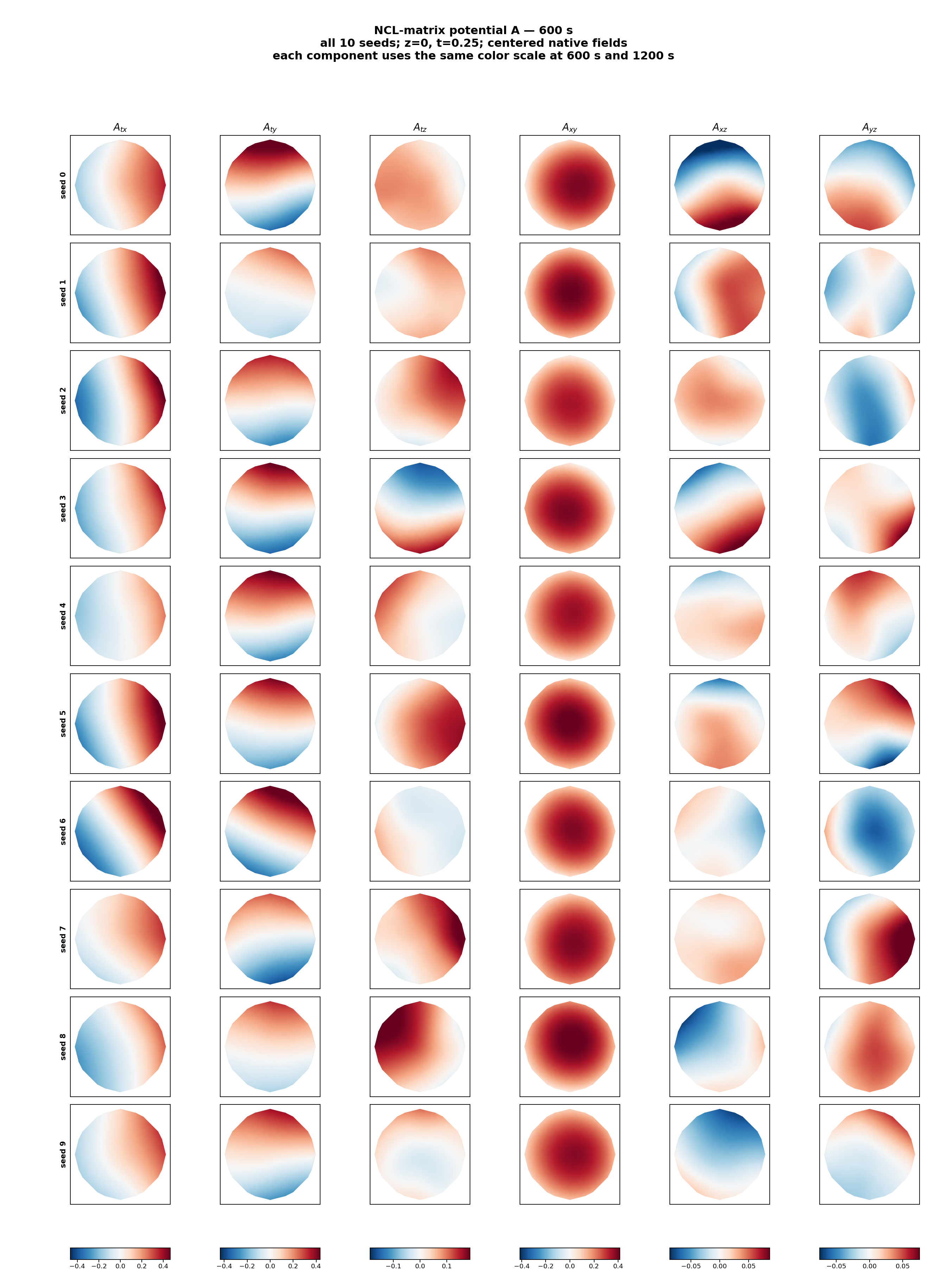}
\par\smallskip
\noindent\textbf{All ten learned six-component antisymmetric potentials $A$.} Each row is one independent initialization and the columns are the six independent entries of $A$.  The substantial run-to-run variation illustrates the redundant potential coordinates available before the fixed decoder maps them to the physical field.\par
\end{figure}

\subsection{Why redundancy can help, and why it is not a monotonic rule}

At a nonzero Fourier mode, write a potential-based decoder abstractly as $\widehat u(k)=\mathcal L_k z(k)$, where $z$ collects the learned auxiliary coordinates.  If $\eta(k)\in\ker\mathcal L_k$, then
\begin{equation}
 \mathcal L_k\bigl(z(k)+\eta(k)\bigr)=\mathcal L_k z(k)=\widehat u(k).
 \label{eq:redundancy_equivalence_class}
\end{equation}
Consequently every physical optimum is an equivalence class $z^\star+\ker\mathcal L_k$, rather than a single auxiliary coordinate.  In the $D=4$ experiment, the minimal coordinates have three outputs, rank three, and no structural null direction.  The vector-$B$ construction has four outputs and one null direction, while the antisymmetric $A$ construction has six independent components and three null directions.  Optimization of a physical-field loss therefore takes place on a quotient: it can approach any member of the equivalent potential family instead of being required to select one particular $A$ or $B$ coordinate.

This mechanism gives a concrete explanation for a broad low-loss valley: the loss is constant to first order along exact gauge/null directions, and may be less sharply curved in nearby directions.  Table~\ref{tab:controlled_redundancy_sweep} is the direct experimental check.  Relative to direct six-entry $A$ ($0.1038496$), the $12$--$24$ coordinate dense constructions reach $0.0974009$, $0.0988612$, and $0.1001312$; the unrestricted vector--outer-product construction reaches $0.0951862$, $0.0862193$, $0.0837907$, and $0.0938667$ for $2$, $3$, $4$, and $6$ vector pairs.  The best record is therefore the $32$-coordinate, $29$-null-direction outer-product representation, which reaches $0.0837907$ in $116.23$ seconds, compared with $0.1038496$ in $117.72$ seconds for direct $A$.

The evidence nevertheless does not support a blanket monotonic law.  Extra coordinates can also introduce cancellations, poor conditioning, or slow motion in weakly identified directions: the $48$-coordinate outer-product version is still better than direct $A$ but not better than the $32$-coordinate version, and the multi-potential sweep has its best values at intermediate branch counts.  The supported conclusion is that deliberately introduced redundancy creates many physically equivalent solutions and, in this controlled record, materially improves accessible loss and optimization speed over a useful range; the amount and decoder structure still matter.

\subsection{Two-dimensional optimal transport evolving in time}

The stored optimal-transport diagnostic evolves a two-dimensional spatial transport field over time, so its conservation law is represented on the three-coordinate space--time domain $(x,y,t)$.  At $10$k optimization steps, the original positive-density NCL branch obtained a loss of $5.989\!\times\!10^{-1}$ in $374$ seconds with $119{,}010$ parameters.  The generic vector-$B$ and matrix-$A$ alternatives used $28{,}611$ parameters and obtained losses $4.931$ and $8.613$ in $296$ and $287$ seconds, respectively.  This raw comparison is useful as a training diagnostic, but it is not a clean causal measurement of redundancy because the original NCL branch has a different positive-density construction and approximately four times as many parameters.

Within the generic overcomplete family itself, merely increasing coordinate count was again nonmonotonic: the base dense representation attained a $10$k loss of $8.396$, whereas its 18- and 24-coordinate versions attained $64.50$ and $78.88$.  The two experimental settings therefore agree on the appropriate interpretation: redundant coordinates can supply equivalent physical solutions and can improve optimization when the representation is well conditioned, but redundancy must be evaluated together with decoder structure, parameter budget, and numerical conditioning.  The controlled Ball record above is the direct evidence for the landscape explanation; the optimal-transport record is retained as a complementary stress test with its differing density architecture stated explicitly.

\end{document}